\documentclass{article}

\usepackage[nonatbib, final]{nips_2018}

\usepackage[
    backend=biber,
    style=authoryear,
    maxcitenames=2,
    maxbibnames=99,
    giveninits=true,
    uniquename=init,
    natbib=true
]{biblatex}

\AtEveryBibitem{%
    \clearfield{month}
    \clearfield{isbn}
    \clearfield{issn}
    \clearfield{pages}
    
    \ifentrytype{online}{}{
        \clearfield{url}
    }
}
\usepackage[utf8]{inputenc} 
\usepackage[T1]{fontenc}    
\usepackage{hyperref}       
\usepackage{url}            
\usepackage{booktabs}       
\usepackage{amsfonts}       
\usepackage{nicefrac}       
\usepackage{microtype}      
\usepackage{appendix}

\usepackage[acronym,smallcaps,nowarn,section,nogroupskip,nonumberlist]{glossaries}

\usepackage{todonotes}
\usepackage{subcaption}
\usepackage[linesnumbered,ruled,vlined]{algorithm2e}
\usepackage{bm} 
\usepackage{enumitem} 

\usepackage{mlmacros} 
\usepackage[nameinlink,capitalise]{cleveref} 

\newcommand*{\defeq}{\mathrel{\vcenter{\baselineskip0.5ex \lineskiplimit0pt
                     \hbox{\scriptsize.}\hbox{\scriptsize.}}}%
                     =}

\presetkeys{todonotes}{%
  backgroundcolor=blue!10!white,
  linecolor=blue!10!white,
  bordercolor=blue!10!white
}{}

\usepackage[skip=5pt,font=small]{caption}
\usepackage{subcaption}

\usepackage{titlesec}

\titlespacing*{\section}{0pt}{.1\baselineskip}{.1\baselineskip}
\titlespacing*{\subsection}{0pt}{0.1\baselineskip}{0.1\baselineskip}

\variables{l,e,s,x,y,z,b,k,g,y,n,w,D,P,S}
\variables[mean]{\mu}
\variables[std]{\sigma}

\probdists{p,q}
\probdists[pd]{p^D}
\probdists[pp]{p^P}
\probdists[qd]{q^D}
\probdists[qp]{q^P}

\newcommand{\STN}{\operatorname{ST}}
\newcommand{\bern}{\operatorname{Bernoulli}}
\newcommand{\cat}{\operatorname{Categorical}}

\definecolor{pink}{HTML}{ff00ff}
\definecolor{orange}{HTML}{ff9900}
\definecolor{blue}{HTML}{0000ff}

\SetKw{Continue}{continue}

\newcommand{\hT}[2]{\textcolor{blue}{\bm{h}_{#1}^{T, #2}}}
\newcommand{\hR}[2]{\textcolor{orange}{\bm{h}_{#1}^{R, #2}}}
\newcommand{\hD}[2]{\textcolor{pink}{\bm{h}_{#1}^{D, #2}}}

\newcommand{\RT}{\textcolor{blue}{\operatorname{R}_\phi^T}}
\newcommand{\Rr}{\textcolor{orange}{\operatorname{R}_\phi^R}}
\newcommand{\RD}{\textcolor{pink}{\operatorname{R}_\phi^D}}

\newacronym{VAE}{vae}{variational auto-encoder}
\newacronym{IS}{is}{importance sampling}
\newacronym{IWAE}{iwae}{importance-weighted auto-encoder}
\newacronym{FIVO}{fivo}{Filtering Variational Objectives}
\newacronym{SMC}{smc}{sequential Monte Carlo}
\newacronym{SSM}{ssm}{state-space model}
\newacronym{SGA}{sga}{stochastic gradient ascent}
\newacronym{SGD}{sgd}{stochastic gradient descent}
\newacronym{ELBO}{elbo}{evidence lower bound}
\newacronym{KL}{kl}{Kullback-Leibler}
\newacronym{LSTM}{lstm}{long short-term memory}
\newacronym{CNN}{cnn}{convolutional neural network}
\newacronym{MML}{mml}{maximum marginal likelihood}
\newacronym{NVIL}{nvil}{Neural Variational Inference and Learning}
\newacronym{VIMCO}{vimco}{Variational inference for Monte Carlo objectives}
\newacronym[firstplural=Imagination-Augmented Agent, plural=I2As]{I2A}{i2a}{imagination-augmented agent}

\newacronym{REINFORCE}{reinforce}{REINFORCE}
\glsunset{REINFORCE}
\newacronym{RL}{rl}{reinforcement learning}

\newacronym{ADAM}{adam}{ADAM}
\glsunset{ADAM}
\newacronym{RMSprop}{rmsprop}{RMSprop}
\glsunset{RMSprop}

\newacronym{GAN}{gan}{generative adversarial network}
\newacronym{GRU}{gru}{gated recurrent unit}
\newacronym{MLP}{mlp}{multilayer perceptron}
\newacronym{MC}{mc}{Monte Carlo}
\newacronym{bnp}{BNP}{Bayesian nonparametric}
\newacronym{mibp}{MIBP}{Markov Indian buffet process}
\newacronym{VRNN}{vrnn}{Variational Recurrent Neural Network}
\newacronym{LGSSM}{lgssm}{linear Gaussian state space model}
\newacronym[firstplural=recurrent neural networks, plural=RNNs]{RNN}{rnn}{recurrent neural network}

\newacronym{AIR}{air}{Attend, Infer, Repeat}
\newacronym{SQAIR}{sqair}{Sequential Attend, Infer, Repeat}
\newacronym{prop}{prop}{Propagation}
\newacronym{disc}{disc}{Discovery}
\newacronym{ELU}{elu}{exponential linear unit}

\newacronym{MNIST}{mnist}{mnist}
\glsunset{MNIST}

\title{Sequential Attend, Infer, Repeat: Generative Modelling of Moving Objects}

\author{
  Adam R.~Kosiorek\thanks{Corresponding author: \href{mailto:adamk@robots.ox.ac.uk}{adamk@robots.ox.ac.uk}}$~~^{\mathcal{x}}$ $^{\mathcal{y}}$
   \And
   Hyunjik Kim$^{\mathcal{y}}$ 
   \And
   Ingmar Posner$^{\mathcal{x}}$ 
   \And
   Yee Whye Teh$^{\mathcal{y}}$ 
   \AND
   $^{\mathcal{x}}$ Applied Artificial Intelligence Lab\\
   Oxford Robotics Institute \\
   University of Oxford
   \And
   $^{\mathcal{y}}$ Department of Statistics \\
   University of Oxford
}

\begin{document}

\maketitle

\vspace{-20pt}
\begin{abstract}
\vspace{-5pt}
We present Sequential Attend, Infer, Repeat (\textsc{sqair}), an interpretable deep generative model for videos of moving objects.
It can reliably discover and track objects throughout the sequence of frames, and can also generate future frames conditioning on the current frame, thereby simulating expected motion of objects. 
This is achieved by explicitly encoding object presence, locations and appearances in the latent variables of the model.
\textsc{Sqair} retains all strengths of its predecessor, Attend, Infer, Repeat (\textsc{air}, \cite{Eslami2016}), including learning in an unsupervised manner, and addresses its shortcomings.
We use a moving multi-\textsc{mnist} dataset to show limitations of AIR in detecting overlapping or partially occluded objects, and show how \textsc{sqair} overcomes them by leveraging temporal consistency of objects.
Finally, we also apply \textsc{sqair} to real-world pedestrian CCTV data, where it learns to reliably detect, track and generate walking pedestrians with no supervision.
\end{abstract}
\section{Introduction}
\label{sec:intro}

The ability to identify objects in their environments and to understand relations between them is a cornerstone of human intelligence \citep{kemp2008discovery}. Arguably, in doing so we rely on a notion of spatial and temporal consistency which gives rise to an expectation that objects do not appear out of thin air, nor do they spontaneously vanish, and that they can be described by properties such as location, appearance and some dynamic behaviour that explains their evolution over time. We argue that this notion of consistency can be seen as an \textit{inductive bias} that improves the efficiency of our learning. Equally, we posit that introducing such a bias towards spatio-temporal consistency into our models should greatly reduce the amount of supervision required for learning.

One way of achieving such inductive biases is through model structure. 
While recent successes in deep learning demonstrate that progress is possible without explicitly imbuing models with interpretable structure \citep{lecun2015deep}, recent works show that introducing such structure into deep models can indeed lead to favourable inductive biases improving performance  e.g.\ in convolutional networks \citep{lecun1989backpropagation} or in tasks requiring relational reasoning \citep{Santoro2017}.
Structure can also make neural networks useful in new contexts by significantly improving generalization, data efficiency \citep{jacobsen2016struc} or extending their capabilities to unstructured inputs \citep{Graves2016}.

\gls{AIR}, introduced by \cite{Eslami2016}, is a notable example of such a structured probabilistic model that relies on deep learning and admits efficient amortized inference.
Trained without any supervision, \gls{AIR} is able to decompose a visual scene into its constituent components and to generate a (learned) number of latent variables that explicitly encode the location and appearance of each object. While this approach is inspiring, its focus on modelling individual (and thereby inherently static) scenes leads to a number of limitations. For example, it often merges two objects that are close together into one since no temporal context is available to distinguish between them. Similarly, we demonstrate that \gls{AIR} struggles to identify partially occluded objects, e.g.\ when they extend beyond the boundaries of the  scene frame (see \Cref{fig:partial_glimpse} in \Cref{sec:expr_mnist}). 

Our contribution is to mitigate the shortcomings of  \gls{AIR} by introducing a sequential version that models sequences of frames, enabling it to discover and track objects over time as well as to generate convincing extrapolations of frames into the future. We achieve this by leveraging temporal information to learn a richer, more capable generative model. Specifically, we extend \gls{AIR} into a spatio-temporal state-space model and train it on unlabelled image sequences of dynamic objects. 
We show that the resulting model, which we name Sequential \gls{AIR} (\acrshort{SQAIR}), retains the strengths of the original AIR formulation while outperforming it on moving \gls{MNIST} digits.

The rest of this work is organised as follows.
In \Cref{sec:air}, we describe the generative model and inference of \gls{AIR}.
In \Cref{sec:sqair}, we discuss its limitations and how it can be improved, thereby introducing \gls{SQAIR}, our extension of \gls{AIR} to image sequences.
In \Cref{sec:experiments}, we demonstrate the model on a dataset of multiple moving MNIST digits (\Cref{sec:expr_mnist}) and compare it against \gls{AIR} trained on each frame and \gls{VRNN} of \cite{Chung2015} with convolutional architectures, and show the superior performance of \gls{SQAIR} in terms of log marginal likelihood and interpretability of latent variables.
We also investigate the utility of inferred latent variables of \gls{SQAIR} in downstream tasks.
In \Cref{sec:expr_duke} we apply \gls{SQAIR} on real-world pedestrian CCTV data, where \gls{SQAIR} learns to reliably detect, track and generate walking pedestrians without any supervision.
Code for the implementation on the \textsc{mnist} dataset\footnote{code: \href{https://github.com/akosiorek/sqair}{github.com/akosiorek/sqair}} and the results video\footnote{video: \href{https://youtu.be/-IUNQgSLE0c}{youtu.be/-IUNQgSLE0c}} are available online.
\section{Attend, Infer, Repeat (\textsc{AIR})}
\label{sec:air}

\gls{AIR}, introduced by \cite{Eslami2016}, is a structured \gls{VAE} capable of decomposing a static scene $\bx$ into its constituent objects, where each object is represented as a separate triplet of continuous latent variables $\bz = \{\bz^{\mathrm{what}, i}, \bz^{\mathrm{where}, i}, z^{\mathrm{pres},i}\}_{i=1}^n$, $n \in \mathbb{N}$ being the (random) number of objects in the scene.
Each triplet of latent variables explicitly encodes position, appearance and presence of the respective object, and the model is able to infer the number of objects present in the scene. Hence it is able to count, locate and describe objects in the scene, all learnt in an unsupervised manner, made possible by the inductive bias introduced by the model structure.

\begin{figure}
    \centering
    \includegraphics[width=0.7\linewidth]{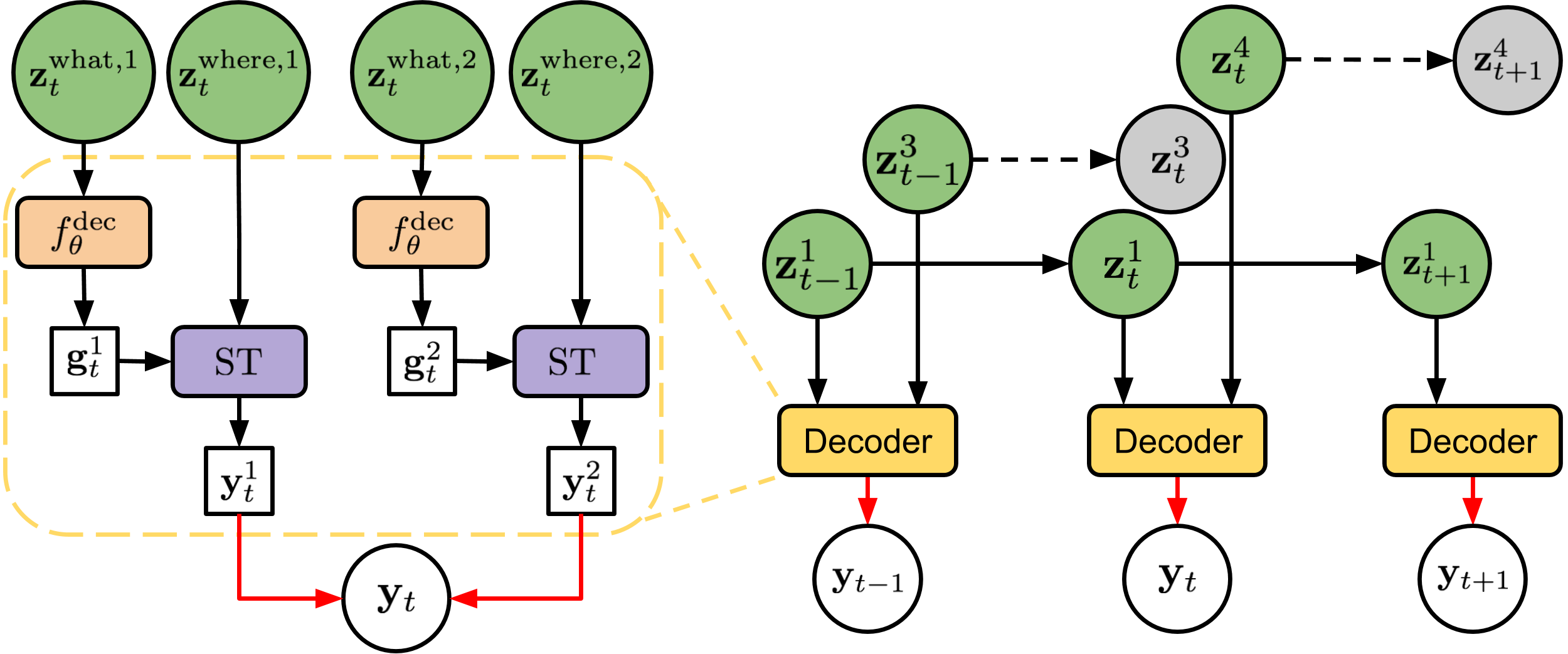}
    \caption{
        \textit{Left}:
            Generation in \gls{AIR}.
            The image mean $\byt$ is generated by first using the \textit{glimpse decoder} $f_\theta^\mathrm{dec}$ to map the \textit{what} variables into glimpses $\bgt$, transforming them with the \textit{spatial transformer} $\STN$ according to the \textit{where} variables and summing up the results.
        \textit{Right}:
            Generation in \gls{SQAIR}.
            When new objects enter the frame, new latent variables (here, $\bzt^4$) are sampled from the \textit{discovery} prior. The temporal evolution of already present objects is governed by the \textit{propagation} prior, which can choose to forget some variables (here, $\bzt^3$ and $\bz_{t+1}^4$) when the object moves out of the frame. The image generation process, which mimics the left-hand side of the figure, is abstracted in the \textit{decoder} block. 
    }
    \label{fig:generation}
\end{figure}

\textbf{Generative Model}
The generative model of \gls{AIR} is defined as follows
\vspace{-7pt}
\begin{align}
\label{eq:air_gen}
    \p{n}{}{\theta} &= \mathrm{Geom} (n \mid \theta), 
    &
    \p{\bz^{\mathrm{w}}}{n}{\theta} &= \prod_{i=1}^n \p{\bz^{w,i}}{}{\theta} = \prod_{i=1}^n \gauss{\bz^{w,i}|\bf{0}, \bf{I}},\nonumber \\
    \p{\bx}{\bz}{\theta} &= \gauss{\bx \mid \byt, \sigma^2_x \bm{I}}, 
    &\text{with}~~\byt &= \sum_{i=1}^n \operatorname{h}^\mathrm{dec}_\theta (
        \bz^{\mathrm{what}, i}, \bz^{\mathrm{where}, i}
    ),
\end{align}
where $\bz^{\mathrm{w},i} \defeq (\bz^{\mathrm{what}, i}, \bz^{\mathrm{where}, i})$, $z^{\mathrm{pres}, i}=1$ for $i=1 \ldots n$ and $h^\mathrm{dec}_\theta$ is the object decoder with parameters $\theta$.
It is composed of a \textit{glimpse decoder} $f_\theta^\mathrm{dec}: \bgt^i \mapsto \byt^i$,
which constructs an image patch and a 
spatial transformer ($\operatorname{ST}$, \cite{Jaderberg2015}), which scales and shifts it according to $\bz^\mathrm{where}$; see \Cref{fig:generation} for details.

\textbf{Inference}
\cite{Eslami2016} use a sequential inference algorithm, where latent variables are inferred one at a time; see \Cref{fig:sqair_inf_flow}.
The number of inference steps $n$ is given by $z^{\mathrm{pres}, 1:n+1}$, a random vector of $n$ ones followed by a zero. The $\bz^{i}$ are sampled sequentially from
\vspace{-7pt}
\begin{equation} \label{eq:air_posterior}
    \q{\bz}{\bx}{\phi} = 
        \q{z^{\mathrm{pres}, n+1} = 0}{\bz^{\mathrm{w}, 1:n}, \bx}{\phi} 
        \prod_{i=1}^n 
        \q{\bz^{\mathrm{w}, i}, z^{\mathrm{pres}, i} = 1}{\bz^{1:i-1}, \bx}{\phi},
\end{equation}
where $q_\phi$ is implemented as a neural network with parameters $\phi$. 
To implement explaining away, e.g.\ to avoid encoding the same object twice, it is vital to capture the dependency of $\bz^{\mathrm{w},i}$ and  $z^{\mathrm{pres}, i}$ on $\bz^{1:i-1}$ and $\bx$. This is done using a \gls{RNN} $R_\phi$ with hidden state $\bm{h}^i$, namely:
$
    \bm{\omega}^i, \bm{h}^i = R_\phi (\bx, \bz^{i-1}, \bm{h}^{i-1}).
$
The outputs $\bm{\omega}^i$, which are computed iteratively and depend on the previous latent variables (\textit{cf}. \Cref{algo:sqair_disc}), parametrise $\q{\bz^{\mathrm{w},i}, \bz^{\mathrm{pres},i}}{\bz^{1:i-1}, \bx}{\phi}$. For simplicity the latter is assumed to factorise such that
$
    \q{\bz^{\mathrm{w}}, \bz^{\mathrm{pres}}}{\bz^{1:i-1}, \bx}{\phi} = \q{z^{\mathrm{pres}, n+1} = 0}{\bm{\omega}^{n+1}}{\phi} \prod_{i=1}^n \q{\bz^{\mathrm{w},i}}{\bm{\omega}^i}{\phi} \q{z^{\mathrm{pres}, i} = 1}{\bm{\omega}^i}{\phi}.
$

\section{Sequential Attend-Infer-Repeat}
\label{sec:sqair}

\begin{figure}
    \centering
    \includegraphics[width=.7\linewidth]{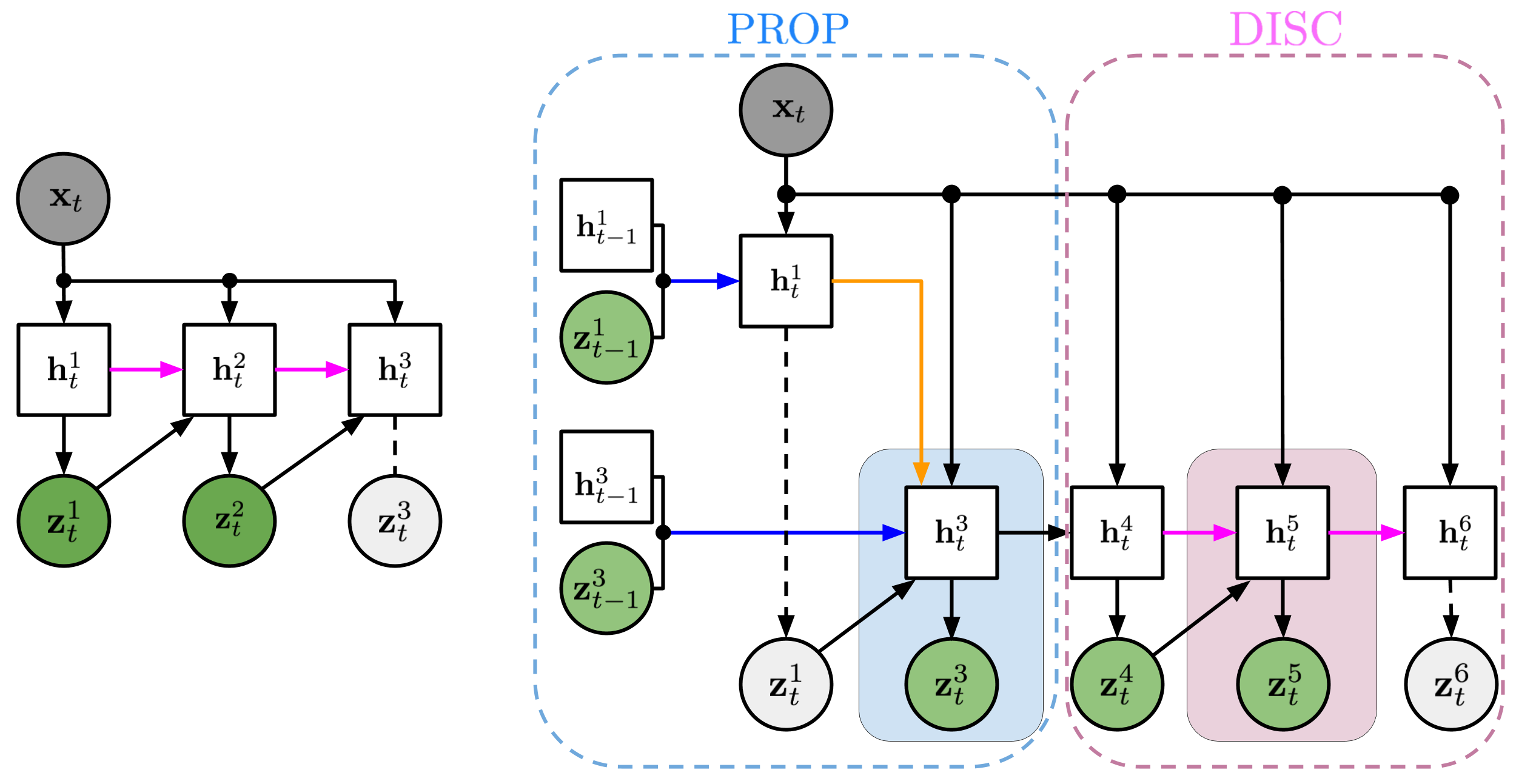}
    \caption{
    \textit{Left}: 
        Inference in \gls{AIR}. 
        The \textcolor{pink}{pink \gls{RNN}} attends to the image sequentially and produces one latent variable $\bzt^i$ at a time. 
        Here, it decides that two latent variables are enough to explain the image and $\bzt^3$ is not generated.
    \textit{Right}:
        Inference in \gls{SQAIR} starts with the \gls{prop} phase.
        \gls{prop} iterates over latent variables from the previous time-step $t-1$ and updates them based on the new observation $\bxt$.
        The \textcolor{blue}{blue \gls{RNN}} runs forward in time to update the hidden state of each object, to model its change in appearance and location throughout time. 
        The \textcolor{orange}{orange \gls{RNN}} runs across all current objects and models the relations between different objects.
        Here, when attending to $\bz^1_{t-1}$, it decides that the corresponding object has disappeared from the frame and \textit{forgets} it.
        Next, the \gls{disc} phase detects new objects as in \gls{AIR}, but in \gls{SQAIR} it is also conditioned on the results of \gls{prop}, to prevent rediscovering objects. See \Cref{fig:sqair_inf_detail} for details of the colored \glspl{RNN}.}
    \label{fig:sqair_inf_flow}
\end{figure}

While capable of decomposing a scene into objects, \gls{AIR} only describes single images. Should we want a similar decomposition of an image sequence, it would be desirable to do so in a temporally consistent manner. For example, we might want to detect objects of the scene as well as infer dynamics and track identities of any persistent objects.
Thus, we introduce  \glsreset{SQAIR} \gls{SQAIR}, whereby \gls{AIR} is augmented with a \gls{SSM} to achieve temporal consistency in the generated images of the sequence.
The resulting probabilistic model is composed of two parts:  \glsreset{disc}\gls{disc}, which is responsible for detecting  (or introducing, in the case of the generation) new objects at every time-step (essentially equivalent to \gls{AIR}), and \glsreset{prop}\gls{prop}, responsible for updating (or forgetting) latent variables from the previous time-step given the new observation (image), effectively implementing the temporal \gls{SSM}.
We now formally introduce \gls{SQAIR} by first describing its generative model and then the inference network.

\textbf{Generative Model}
The model assumes that at every-time step, objects are first propagated from the previous time-step (\gls{prop}). Then, new objects are introduced (\gls{disc}). 
Let $t \in \mathbb{N}$ be the current time-step.
Let $\mathcal{P}_t$ be the set of objects propagated from the previous time-step and let $\mathcal{D}_t$ be the set of objects discovered at the current time-step, and let $\mathcal{O}_t = \mathcal{P}_t \cup \mathcal{D}_t$ be the set of all objects present at time-step~$t$.
Consequently, at every time step, the model retains a set of latent variables $\bzt^{\mathcal{P}_t} = \{ \bzt^i \}_{i \in \mathcal{P}_t}$, and generates a set of new latent variables $\bzt^{\mathcal{D}_t} = \{ \bzt^i \}_{i \in \mathcal{D}_t}$. Together they form $\bzt \defeq [\bzt^{\mathcal{P}_t}, \bzt^{\mathcal{D}_t}]$, where the representation of the $i^\mathrm{th}$ object $\bzt^i \defeq [ \bzt^{\mathrm{what}, i}, \bzt^{\mathrm{where}, i}, \zt^{\mathrm{pres}, i}]$ is composed of three components (as in \gls{AIR}): 
$\bzt^{\mathrm{what},i}$ and $\bzt^{\mathrm{where},i}$ are real vector-valued variables representing appearance and location of the object, respectively. $\zt^{\mathrm{pres},i}$ is a binary variable representing whether the object is present at the given time-step or not.

At the first time-step ($t = 1$) there are no objects to propagate, so we sample $D_1$, the number of objects at $t=1$, from the discovery prior $\pd{D_1}{}{}$. 
Then for each object $i \in \mathcal{D}_t$, we sample latent variables $\bzt^{\mathrm{what}, i}, \bzt^{\mathrm{where}, i}$ from $\pd{z_1^i}{D_1}{}$. 
At time $t=2$, the \textit{propagation} step models which objects from $t=1$ are propagated to $t=2$, and which objects disappear from the frame, using the binary random variable $(\zt^{\mathrm{pres},i})_{i \in \mathcal{P}_t}$. 
The \textit{discovery} step at $t=2$ models new objects that enter the frame, with a similar procedure to $t=1$: sample $D_2$ (which depends on $\bz_2^{\mathcal{P}_2}$) then sample $(\bz_2^{\mathrm{what}, i}, \bz_2^{\mathrm{where}, i})_{i\in \mathcal{D}_2}$. 
This procedure of propagation and discovery recurs for $t = 2, \ldots T$.
Once the $\bzt$ have been formed, we may generate images $\bxt$ using the exact same generative distribution $\p{\bxt}{\bzt}{\theta}$ as in \gls{AIR} (\textit{cf}. \Cref{eq:air_gen,fig:generation,algo:air_decoding}). 
In full, the generative model is:
\begin{equation} \label{eq:full_joint}
p(\bx_{1:T},\bz_{1:T},D_{1:T}) 
    = \pd(D_1,\bz_1^{\mathcal{D}_1}) \prod_{t=2}^T \pd(D_t,\bzt^{\mathcal{D}_t}|\bzt^{\mathcal{P}_t})\pp(\bzt^{\mathcal{P}_t}|\bz_{t-1}) p_{\theta}(\bxt|\bzt),
\end{equation}
The \textit{discovery prior} $\pd(D_t,\bzt^{\mathcal{D}_t}|\bzt^{\mathcal{P}_t})$ samples latent variables for new objects that enter the frame. The \textit{propagation prior} $\pp(\bzt^{\mathcal{P}_t}|\bz_{t-1})$ samples latent variables for objects that persist in the frame and removes latents of objects that disappear from the frame, thereby modelling dynamics and appearance changes. Both priors are learned during training. The exact forms of the priors are given in \Cref{apd:sqair_generation}.
\begin{figure}
    \centering
    \begin{minipage}[c]{0.49\linewidth}
        \includegraphics[width=\linewidth]{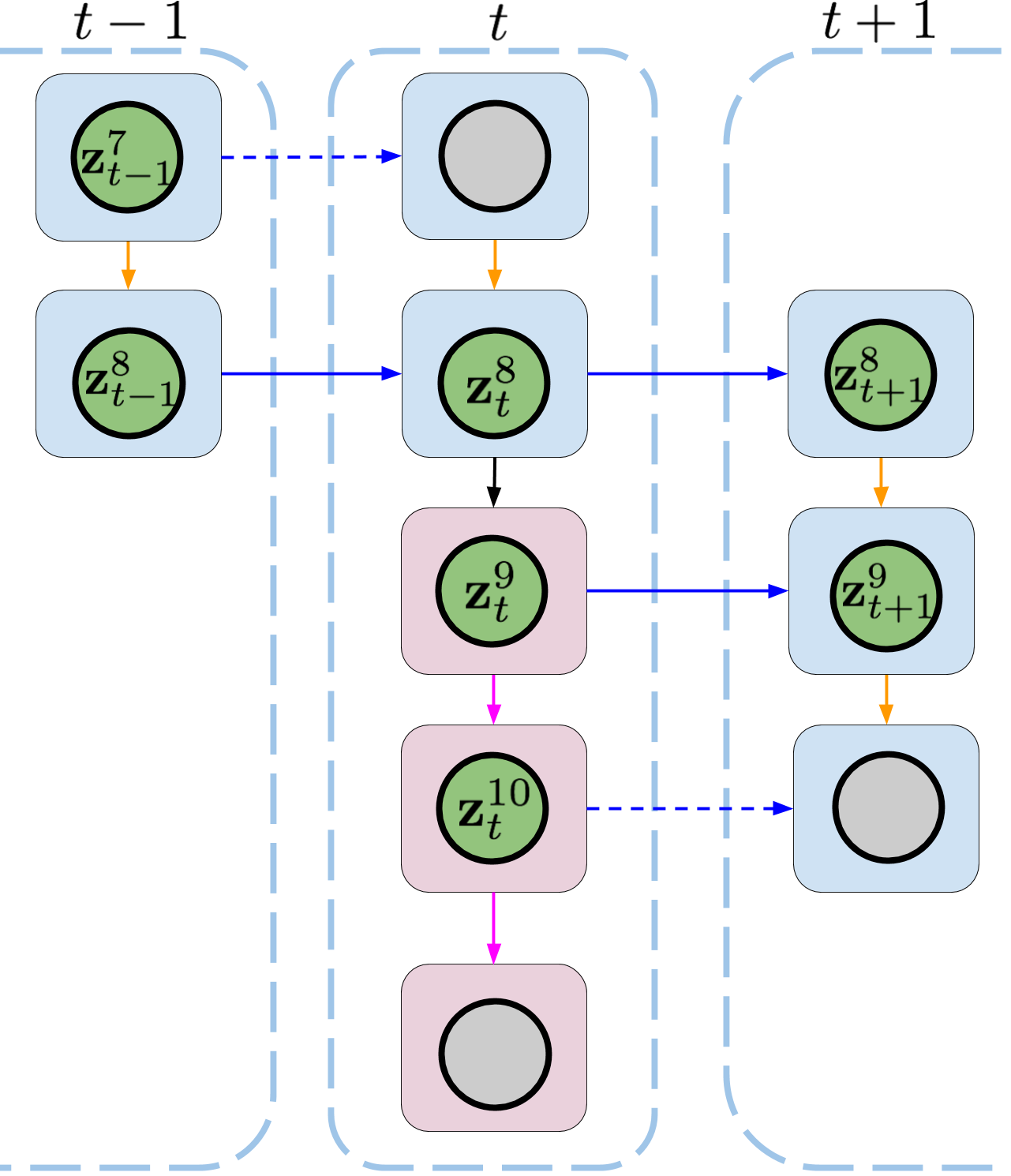}
    \end{minipage}
    \hfill
    \begin{minipage}[c]{0.49\linewidth}
    \caption{
        \textit{Left}:
            Interaction between \gls{prop} and \gls{disc} in \gls{SQAIR}.
            Firstly, objects are propagated to time $t$, and object $i=7$ is dropped.
            Secondly, \gls{disc} tries to discover new objects.
            Here, it manages to find two objects: $i=9$ and $i=10$.
            The process recurs for all remaining time-steps.
            \textcolor{blue}{Blue arrows} update the temporal hidden state, \textcolor{orange}{orange ones} infer relations between objects, \textcolor{pink}{pink ones} correspond to discovery.
        \textit{Bottom}:
            Information flow in a single discovery block (\textit{left}) and propagation block (\textit{right}). 
            In \gls{disc} we first predict \textit{where} and extract a glimpse. We then predict \textit{what} and \textit{presence}.
            \Gls{prop} starts with extracting a glimpse at a candidate location and updating \textit{where}. Then it follows a procedure similar to \gls{disc}, but takes the respective latent variables from the previous time-step into account. 
            It is approximately two times more computationally expensive than \gls{disc}.
            For details, see \Cref{algo:sqair_prop,algo:sqair_disc} in \Cref{app:algo}.
    }
    \label{fig:sqair_inf_detail}
    \end{minipage}
    \begin{minipage}[c]{0.43\linewidth}
        \centering
        \includegraphics[width=\linewidth]{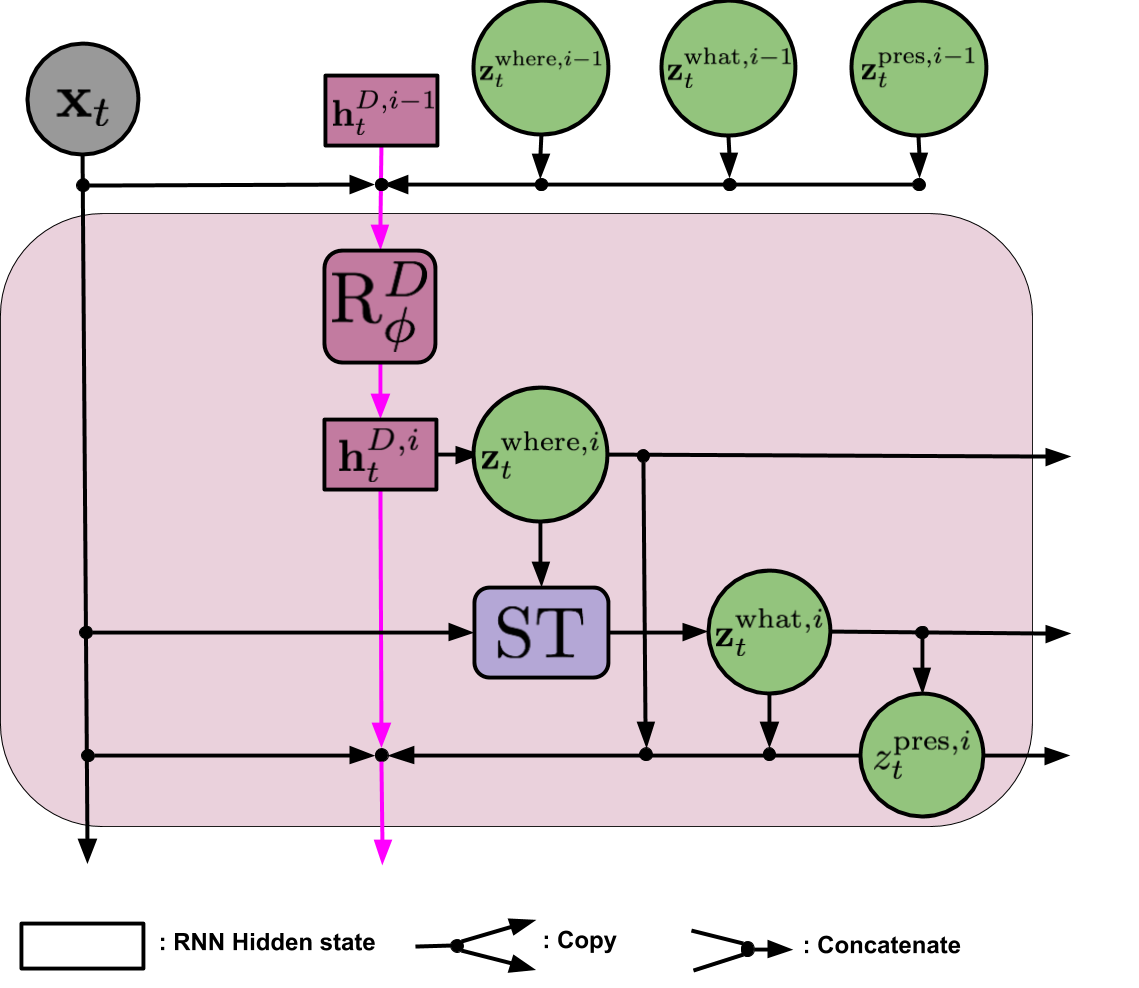}
    \end{minipage}
    \hfill
    \begin{minipage}[c]{0.55\linewidth}
        \centering
        \includegraphics[width=\linewidth]{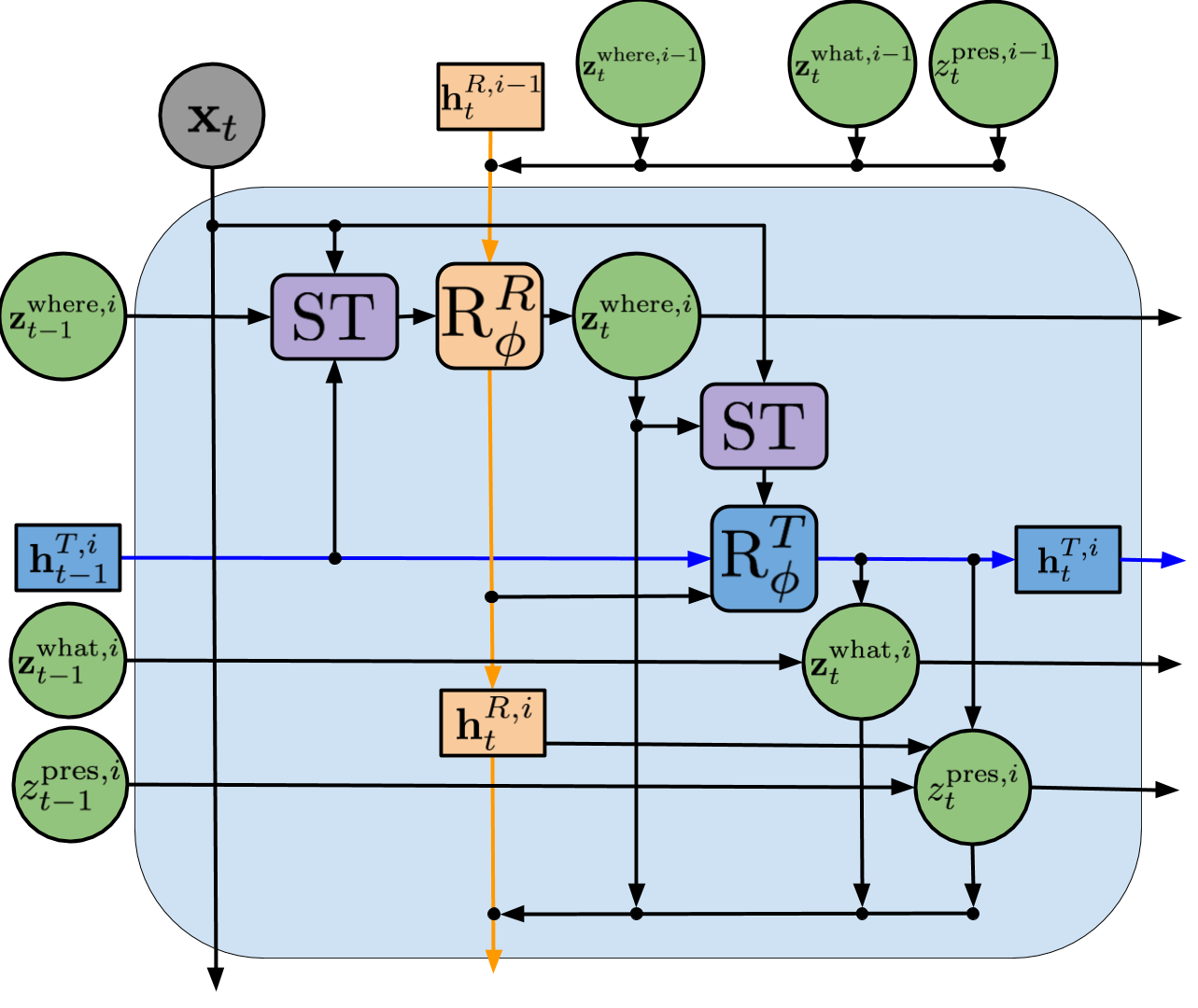}
    \end{minipage}
\end{figure}

\textbf{Inference}
Similarly to \gls{AIR}, inference in \gls{SQAIR} can capture the number of objects and the representation describing the location and appearance of each object that is necessary to explain every image in a sequence.
As with generation, inference is divided into \gls{prop} and \gls{disc}.
During \gls{prop}, the inference network achieves two tasks.
Firstly,
the latent variables from the previous time step are used to infer the current ones, modelling the change in location and appearance of the corresponding objects, thereby attaining temporal consistency. 
This is implemented by the \textit{temporal} \gls{RNN} $\RT$, with hidden states $\textcolor{blue}{\bm{h}_t^T}$ (recurs in $t$). 
Crucially, it does not access the current image directly, but uses the output of the \textit{relation} \gls{RNN} (\textit{cf}. \cite{Santoro2017}).
The relation \gls{RNN} takes relations between objects into account, thereby implementing the \textit{explaining away} phenomenon; 
it is essential for capturing any interactions between objects as well as occlusion (or overlap, if one object is occluded by another). See \Cref{fig:partial_glimpse} for an example.  
These two \gls{RNN}s together decide whether to retain or to forget objects that have been propagated from the previous time step. 
During \gls{disc}, the network infers further latent variables that are needed to describe any new objects that have entered the frame. 
All latent variables remaining after \gls{prop} and \gls{disc} are passed on to the next time step.

See \Cref{fig:sqair_inf_flow,fig:sqair_inf_detail} for the inference network structure .
The full variational posterior is defined as
\vspace{-5pt}
\begin{equation} \label{eq:full_q}
    \q{\Dts, \bzTs}{\bxTs}{\phi} 
        = \prod_{t=1}^T \qd{ \Dt, \bzt^{\mathcal{D}_t} }{ \bxt, \bzt^{\mathcal{P}_t} }{ \phi }
        \prod_{i \in \mathcal{O}_{t-1}} \qp{\bzt^i}{\bz_{t-1}^{i}, \hT{t}{i}, \hR{t}{i}}{\phi}.
\end{equation}
Discovery, described by $\qd_\phi$, is very similar to the full posterior of \gls{AIR}, \textit{cf}. \Cref{eq:air_posterior}.
The only difference is the conditioning on $\bzt^{\mathcal{P}_t}$, which allows for a different number of discovered objects at each time-step and also for objects explained by \gls{prop} not to be explained again.
The second term, or $\qp_\phi$, describes propagation. The detailed structures of $\qd_\phi$ and $\qp_\phi$ are shown in \Cref{fig:sqair_inf_detail}, while all the pertinent algorithms and equations can be found in \Cref{app:algo,apd:sqair_inference}, respectively.

\textbf{Learning}
We train \gls{SQAIR} as an \gls{IWAE} of \cite{Burda2016}. Specifically, we maximise the importance-weighted evidence lower-bound $\loss[\textsc{IWAE}]$, namely

\begin{equation} \label{eq:iwae}
\begin{aligned}
    \loss[\textsc{IWAE}] = \expc[\bxTs \sim \p{\bxTs}{}{\mathrm{data}}]{ 
        \expc[\q]{
            \log \frac{1}{K} \sum_{k=1}^K \frac{ \p{\bxTs, \bzTs}{}{\theta} }{ \q{\bzTs}{\bxTs}{\phi} }   
        }
    }.
\end{aligned}
\end{equation}
To optimise the above, we use \gls{RMSprop}, $K=5$ and batch size of $32$. We use the \textsc{vimco} gradient estimator of \cite{Mnih2016} to backpropagate through the discrete latent variables $z^{\mathrm{pres}}$, and use reparameterisation for the continuous ones \citep{kingma2013auto}.
We also tried to use \textsc{nvil} of \cite{Mnih2014} as in the original work on \gls{AIR}, but found it very sensitive to hyper-parameters, fragile and generally under-performing.
\section{Experiments}
\label{sec:experiments}

We evaluate \gls{SQAIR} on two datasets.
Firstly, we perform an extensive evaluation on moving \gls{MNIST} digits, where we show that it can learn to reliably detect, track and generate moving digits (\Cref{sec:expr_mnist}). Moreover, we show that \gls{SQAIR} can simulate moving objects into the future --- an outcome it has not been trained for. 
We also study the utility of learned representations for a downstream task.
Secondly, we apply \gls{SQAIR} to real-world pedestrian CCTV data from static cameras (\textit{DukeMTMC}, \cite{ristani2016performance}), where we perform background subtraction as pre-processing. In this experiment, we show that \gls{SQAIR} learns to detect, track, predict and generate walking pedestrians without human supervision.

\subsection{Moving multi-\textsc{mnist}}
\label{sec:expr_mnist}

\begin{figure}
    \centering
    \includegraphics[width=\linewidth]{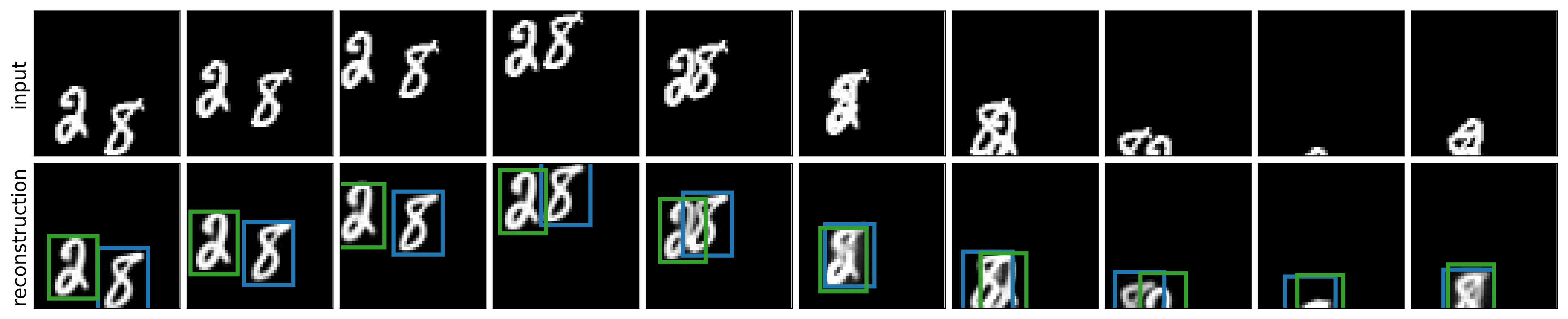}
    \caption{Input images (top) and \gls{SQAIR} reconstructions with marked glimpse locations (bottom). For more examples, see \Cref{fig:mnist_recs_additional} in \Cref{app:mnist_visual}.}
    \label{fig:mnist_recs}
\end{figure}

\begin{figure}
    \centering
    \includegraphics[width=\linewidth]{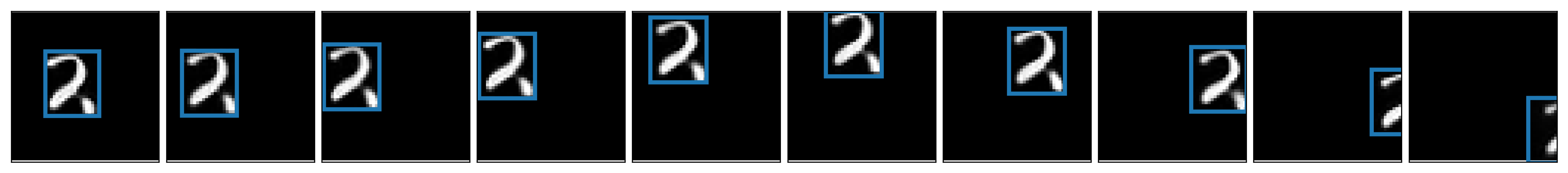}
    \includegraphics[width=\linewidth]{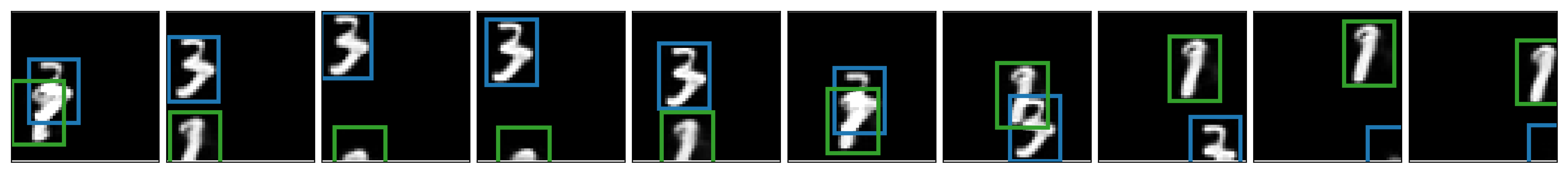}
    \caption{Samples from \gls{SQAIR}. Both motion and appearance are consistent through time, thanks to the propagation part of the model. For more examples, see \Cref{fig:mnist_samples_additional} in \Cref{app:mnist_visual}.}
    \label{fig:mnist_samples}
\end{figure}
\begin{figure}
    \centering
    \includegraphics[width=\linewidth]{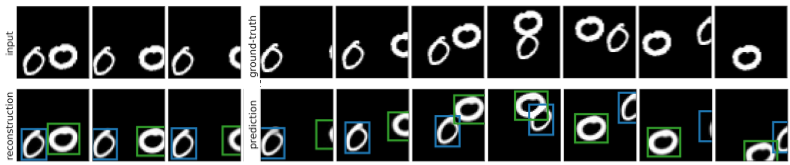}
    \caption{The first three frames are input to \gls{SQAIR}, which generated  the rest conditional on the first frames.}
    \label{fig:mnist_cond_gen}
\end{figure}

\begin{figure}
    \centering
    \begin{minipage}[c]{0.3\linewidth}
        \centering
        \includegraphics[width=\linewidth]{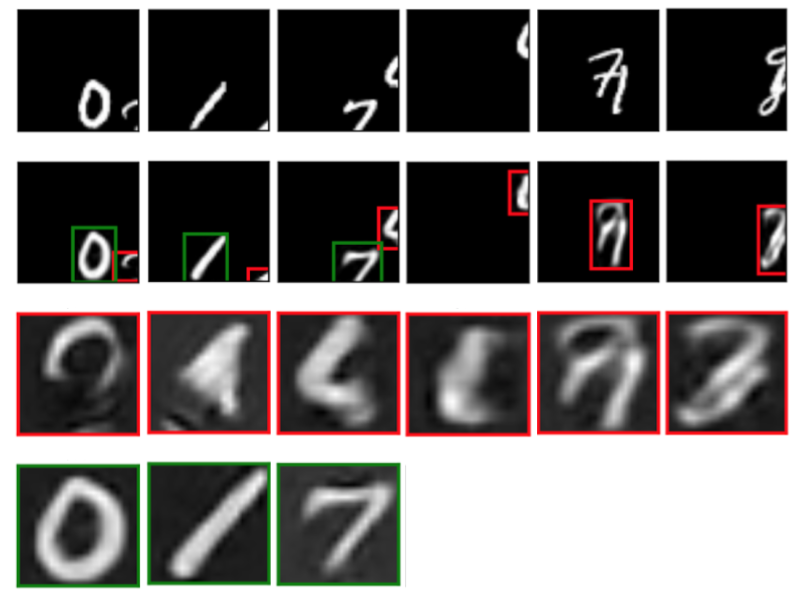}
    \end{minipage}
    \hfill
    \begin{minipage}[c]{0.3\linewidth}
        \centering
        \includegraphics[width=\linewidth]{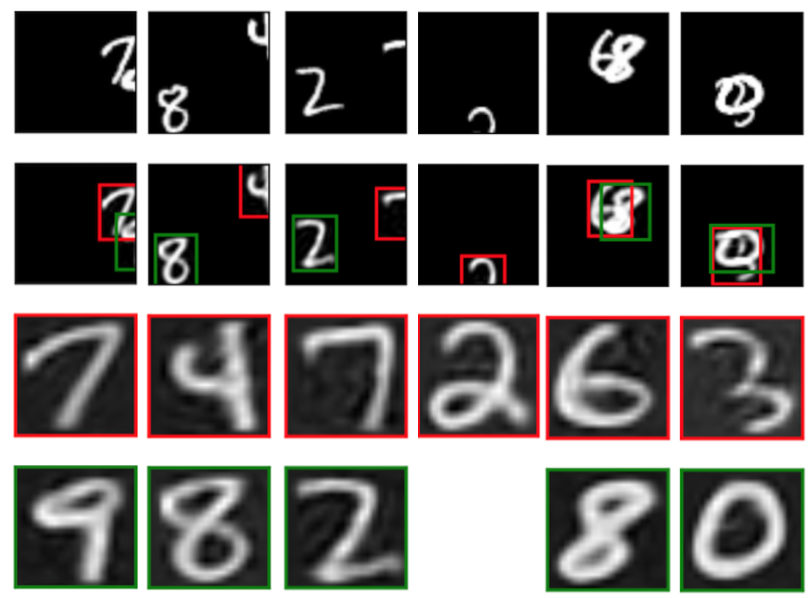}
    \end{minipage}
       \hfill
    \begin{minipage}[c]{0.35\linewidth}
        \centering
        \caption{Inputs, reconstructions with marked glimpse locations and reconstructed glimpses for \gls{AIR} (left) and \gls{SQAIR} (right). \Gls{SQAIR} can model partially visible and heavily overlapping objects by aggregating temporal information.}
        \label{fig:partial_glimpse}
    \end{minipage}
\end{figure}

The dataset consists of sequences of length 10 of multiple moving \gls{MNIST} digits. All images are of size $50 \times 50$ and there are zero, one or two digits in every frame (with equal probability).
Sequences are generated such that no objects overlap in the first frame, and all objects are present through the sequence; the digits can move out of the frame, but always come back.
See \Cref{app:mnist_inout} for an experiment on a harder version of this dataset.
There are 60,000 training and 10,000 testing sequences created from the respective \gls{MNIST} datasets.
We train two variants of \gls{SQAIR}: the \textsc{mlp}-\gls{SQAIR} uses only fully-connected networks, while the \textsc{conv}-\gls{SQAIR} replaces the networks used to encode images and glimpses with convolutional ones; it also uses a subpixel-convolution network as the glimpse decoder \citep{shi2016subpixel}.
See \Cref{app:mnist_details} for details of the model architectures and the training procedure.

We use \gls{AIR} and \gls{VRNN} \citep{Chung2015} as baselines for comparison.  \gls{VRNN} can be thought of as a sequential \gls{VAE} with an \gls{RNN} as its deterministic backbone. Being similar to a \gls{VAE}, its latent variables are not structured, nor easily interpretable. For a fair comparison, we control the latent dimensionality of $\gls{VRNN}$ and the number of learnable parameters. We provide implementation details in \Cref{apd:vrnn}.

\begin{table}
    \centering
    \begin{tabular}{c|c|c|c|c|c}
                         & $\log \p{\bxTs}{}{\theta}$ & $\log \p{\bxTs}{\bzTs}{\theta}$  & $\kl{\q{}{}{\phi}}{\p{}{}{\theta}}$ & Counting & Addition\\
                         \hline
        \textsc{conv}-\gls{SQAIR} & $\bm{6784.8}$ & $\bm{6923.8}$ & $\bm{134.6}$ & $0.9974$ & $0.9990$ \\
        \textsc{mlp}-\gls{SQAIR}  & $6617.6$      & $6786.5$      & $164.5$      & $\mathbf{0.9986}$ & $\mathbf{0.9998}$ \\
        \textsc{mlp}-\gls{AIR}    & $6443.6$      & $6830.6$      & $352.6$      & $0.9058$ & $0.8644$\\
        \textsc{conv}-\gls{VRNN}  & $6561.9$      & $6737.8$      & $270.2$      & n/a & $0.8536$\\
        \textsc{mlp}-\gls{VRNN}   & $5959.3$      & $6108.7$      & $218.3$      & n/a  & 0.8059 \\
    \end{tabular}
    \vspace{5pt}
    \caption{\gls{SQAIR} achieves higher performance than the baselines across a range of metrics. The third column refers to the \gls{KL} divergence between the approximate posterior and the prior. Counting refers to accuracy of the inferred number of objects present in the scene, while addition stands for the accuracy of a supervised digit addition experiment, where a classifier is trained on the learned latent representations of each frame.}
    \label{tab:quant}
\end{table}

The quantitative analysis consists of comparing all models in terms of the marginal log-likelihood $\log \p{\bxTs}{}{\theta}$ evaluated as the $\loss[\textsc{IWAE}]$ bound with $K=1000$ particles, reconstruction quality evaluated as a single-sample approximation of $\expc[\q{}{}{\phi}]{\log \p{\bxTs}{\bzTs}{\theta}}$ and the \gls{KL}-divergence between the approximate posterior and the prior (\Cref{tab:quant}). Additionally, we measure the accuracy of the number of objects modelled by \gls{SQAIR} and \gls{AIR}. \Gls{SQAIR} achieves superior performance across a range of metrics --- its convolutional variant outperforms both \gls{AIR} and the corresponding \gls{VRNN} in terms of model evidence and reconstruction performance. 
The \gls{KL} divergence for \gls{SQAIR} is almost twice as low as for \gls{VRNN} and by a yet larger factor for \gls{AIR}.
We can interpret \gls{KL} values as an indicator of the ability to compress, and we can treat \gls{SQAIR}/\gls{AIR} type of scheme as a version of run-length encoding.
While \gls{VRNN} has to use information to explicitly describe every part of the image, even if some parts are empty, \gls{SQAIR} can explicitly allocate content information ($\bz^\mathrm{what}$) to specific parts of the image (indicated by $\bz^\mathrm{where}$).
\Gls{AIR} exhibits the highest values of \gls{KL}, but this is due to encoding every frame of the sequence independently --- its prior cannot take \textit{what} and \textit{where} at the previous time-step into account, hence higher KL.
The fifth column of \Cref{tab:quant} details the object counting accuracy, that is indicative of the quality of the approximate posterior. It is measured as the sum of $\zt^\mathrm{pres}$ for a given frame against the true number of objects in that frame. As there is no $z^\mathrm{pres}$ for \gls{VRNN} no score is provided. Perhaps surprisingly, this metric is much higher for \gls{SQAIR} than for \gls{AIR}. This is because \gls{AIR} mistakenly infers overlapping objects as a single object. Since \gls{SQAIR} can incorporate temporal information, it does not exhibit this failure mode (\textit{cf}. \Cref{fig:partial_glimpse}).
Next, we gauge the utility of the learnt representations by using them to determine the sum of the digits present in the image (\Cref{tab:quant}, column six). To do so, we train a 19-way classifier (mapping from any combination of up to two digits in the range $[0, 9]$ to their sum) on the extracted representations and use the summed labels of digits present in the frame as the target. \Cref{app:mnist_details} contains details of the experiment. 
\Gls{SQAIR} significantly outperforms \gls{AIR} and both variants of \gls{VRNN} on this tasks.
\Gls{VRNN} under-performs due to the inability of disentangling overlapping objects, while both \gls{VRNN} and \gls{AIR} suffer from low temporal consistency of learned representations, see \Cref{app:mnist_visual}. 
Finally, we evaluate \gls{SQAIR} qualitatively by analyzing reconstructions and samples produced by the model against reconstructions and samples from \gls{VRNN}.
We observe that samples and reconstructions from \gls{SQAIR} are of better quality and, unlike \gls{VRNN}, preserve motion and appearance consistently through time. See \Cref{app:mnist_visual} for direct comparison and additional examples.
Furthermore, we examine conditional generation, where we look at samples from the generative model of \gls{SQAIR} conditioned on three images from a real sequence (see \Cref{fig:mnist_cond_gen}).
We see that the model can preserve appearance over time, and that the simulated objects follow similar trajectories, which hints at good learning of the motion model (see \Cref{app:mnist_visual} for more examples).
\Cref{fig:partial_glimpse} shows reconstructions and corresponding glimpses of \gls{AIR} and \gls{SQAIR}. Unlike \gls{SQAIR}, \gls{AIR} is unable to recognize objects from partial observations, nor can it distinguish strongly overlapping objects (it treats them as a single object; columns five and six in the figure).
We analyze failure cases of \gls{SQAIR} in \Cref{app:fail}.

%
%
%
%
%
%
%

\subsection{Generative Modelling of Walking Pedestrians}
\label{sec:expr_duke}

\begin{figure}
    \centering
    \begin{minipage}[c]{0.49\linewidth}
        \centering
        \includegraphics[width=\linewidth]{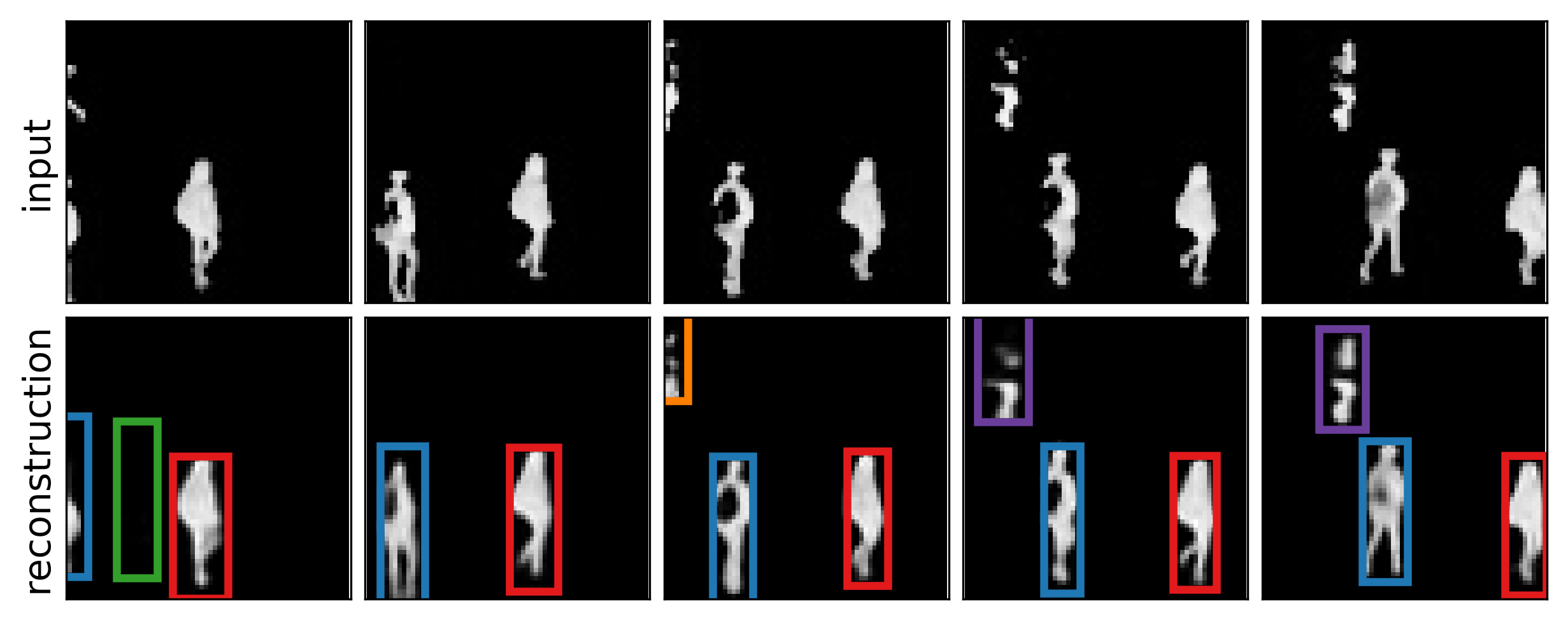}
        \includegraphics[width=\linewidth]{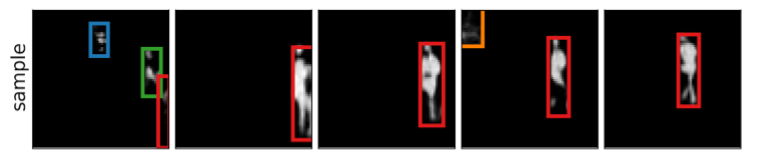}
    \end{minipage}
    \hfill
    \begin{minipage}[c]{0.49\linewidth}
        \centering
        \includegraphics[width=\linewidth]{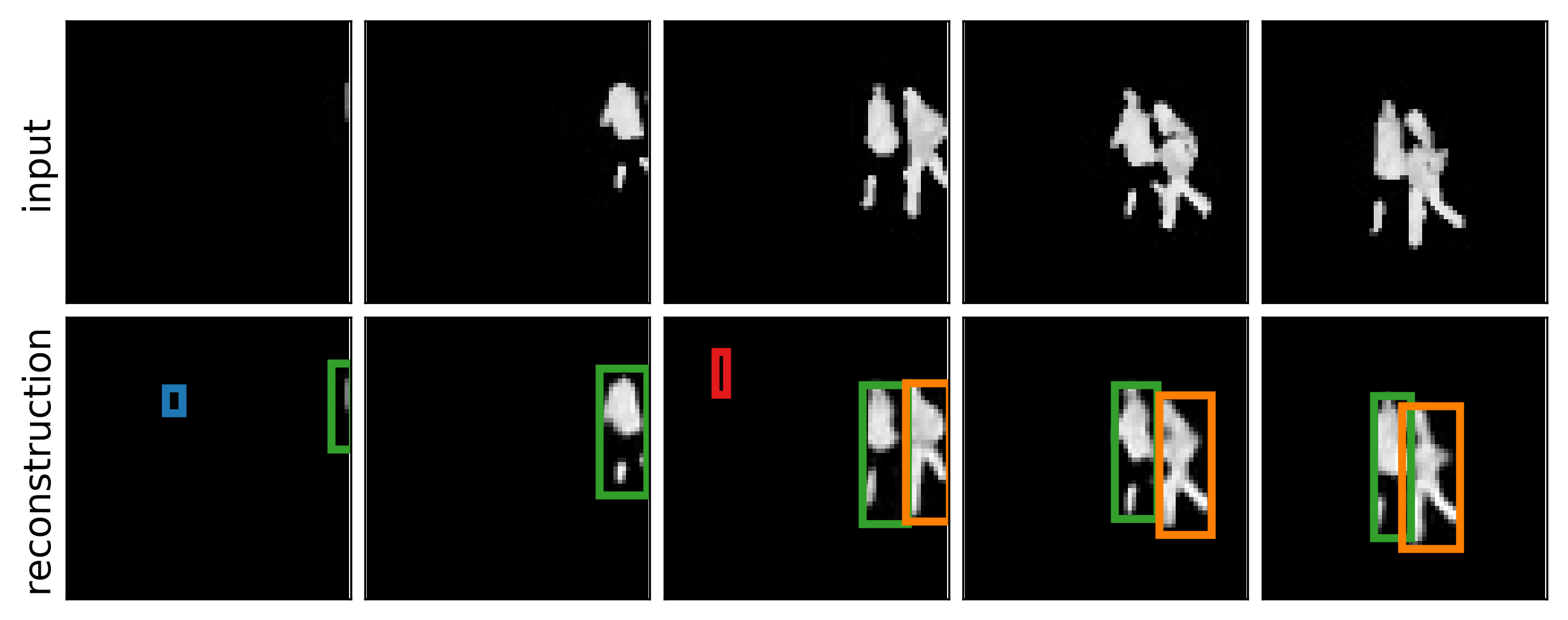}
       \includegraphics[width=\linewidth]{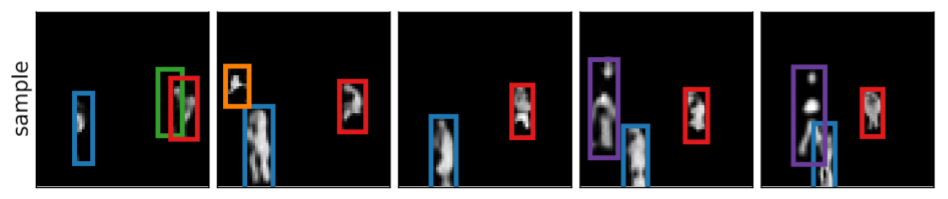}
    \end{minipage}
    \includegraphics[width=.95\linewidth]{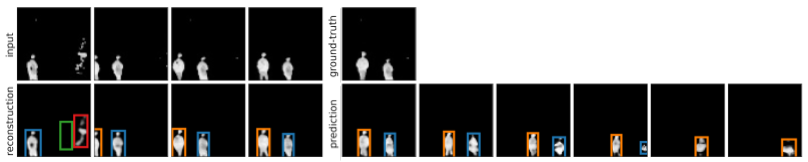}
    \caption{Inputs on the top, reconstructions in the second row, samples in the third row; rows four and five contain inputs and conditional generation: the first four frames in the last row are reconstructions, while the remaining ones are predicted by sampling from the prior. There is no ground-truth, since we used sequences of length five of training and validation.}
    \label{fig:duke_rec}
\end{figure}
To evaluate the model in a more challenging, real-world setting, we turn to data from static CCTV cameras of the \textit{DukeMTMC} dataset \citep{ristani2016performance}. As part of pre-precessing, we use standard background subtraction algorithms \citep{itseez2015opencv}. In this experiment, we use $3150$ training and $350$ validation sequences of length $5$. For details of model architectures, training and data pre-processing, see \Cref{app:duke_details}.
We evaluate the model qualitatively by examining reconstructions, conditional samples (conditioned on the first four frames) and samples from the prior (\Cref{fig:duke_rec} and \Cref{app:duke_visual}).
We see that the model learns to reliably detect and track walking pedestrians, even when they are close to each other.

There are some spurious detections and re-detections of the same objects, which is mostly caused by imperfections of the background subtraction pipeline --- backgrounds are often noisy and there are sudden appearance changes when a part of a person is treated as background in the pre-processing pipeline.
The object counting accuracy in this experiment is $0.5712$ on the validation dataset, and we noticed that it does increase with the size of the training set. We also had to use early stopping to prevent overfitting, and the model was trained for only $315$k iterations ($>1$M for \textsc{mnist} experiments). Hence, we conjecture that accuracy and marginal likelihood can be further improved by using a bigger dataset.
\section{Related Work}

\begin{description}[leftmargin=\parindent]
\item[Object Tracking]
    There have been many approaches to modelling objects in images and videos. 
    Object detection and tracking are typically learned in a supervised manner, where object bounding boxes and often additional labels are part of the training data.
    Single-object tracking commonly use Siamese networks, which can be seen as an \gls{RNN} unrolled over two time-steps \citep{valmadre2017end}. Recently, \cite{kosiorek2017hierch} used an \gls{RNN} with an attention mechanism in the \textsc{hart} model to predict bounding boxes for single objects, while robustly modelling their motion and appearance. Multi-object tracking is typically attained by detecting objects and performing data association on bounding-boxes \citep{bewley2016sort}.
    \cite{schulter2017deepnf} used an end-to-end supervised approach that detects objects and performs data association.
    In the unsupervised setting, where the training data consists of only images or videos, the dominant approach is to distill the inductive bias of spatial consistency into a discriminative model.  \cite{cho2015unsupervised} detect single objects and their parts in images, and \cite{kwak2015unsupervised,xiao2016track} incorporate temporal consistency to better track single objects.
    \Gls{SQAIR} is unsupervised and hence it does not rely on bounding boxes nor additional labels for training, while being able to learn arbitrary motion and appearance models similarly to \textsc{hart} \citep{kosiorek2017hierch}.
    At the same time, is inherently multi-object and performs data association implicitly (\textit{cf}. \Cref{app:algo}).
    Unlike the other unsupervised approaches, temporal consistency is baked into the model structure of \gls{SQAIR} and further enforced by lower \gls{KL} divergence when an object is tracked.

\item[Video Prediction]
     Many works on video prediction learn a deterministic model conditioned on the current frame to predict the future ones \citep{ranzato2014video,srivastava2015unsupervised}.
     Since these models do not model uncertainty in the prediction, they can suffer from the multiple futures problem --- since perfect prediction is impossible, the model produces blurry predictions which are a mean of possible outcomes.
     This is addressed in stochastic latent variable models trained using variational inference to generate multiple plausible videos given a sequence of images \citep{babaeizadeh2017stochastic, denton2018stochastic}.
     Unlike \gls{SQAIR}, these approaches do not model objects or their positions explicitly, thus the representations they learn are of limited interpretability. 
     
\item[Learning Decomposed Representations of Images and Videos]
    Learning decomposed representations of object appearance and position lies at the heart of our model.
    This problem can be also seen as perceptual grouping, which involves modelling pixels as spatial mixtures of entities.
    \cite{Greff2016tagger} and \cite{Greff2017neuralem} learn to decompose images into separate entities by iterative refinement of spatial clusters using either learned updates or the Expectation Maximization algorithm;
    \cite{Ilin2017recurrentln} and \cite{Steenkiste2018relationalnem} extend these approaches to videos, achieving very similar results to \gls{SQAIR}.
    Perhaps the most similar work to ours is the concurrently developed model of \cite{Hsieh2018ddpae}.
    The above approaches rely on iterative inference procedures, but do not exhibit the object-counting behaviour of \gls{SQAIR}.
    For this reason, their computational complexities are proportional to the predefined maximum number of objects, while \gls{SQAIR} can be more computationally efficient by adapting to the number of objects currently present in an image.
    
    Another interesting line of work is the \textsc{gan}-based unsupervised video generation that decomposes motion and content \citep{tulyakov2017mocogan,denton2017unsupervised}. These methods learn interpretable features of content and motion, but deal only with single objects and do not explicitly model their locations. Nonetheless, adversarial approaches to learning structured probabilistic models of objects offer a plausible alternative direction of research.
    
\item[Bayesian Nonparametric Models]
    To the best of our knowledge, \cite{neiswanger2012unsupervised} is the only known approach that models pixels belonging to a variable number of objects in a video together with their locations in the generative sense.
    This work uses a \gls{bnp} model, which relies on mixtures of Dirichlet processes to cluster pixels belonging to an object. 
    However, the choice of the model necessitates complex inference algorithms involving Gibbs sampling and Sequential Monte Carlo, to the extent that any sensible approximation of the marginal likelihood is infeasible.
    It also uses a fixed likelihood function, while ours is learnable.
    
    The object appearance-persistence-disappearance model in \gls{SQAIR} is reminiscent of the \gls{mibp} of \cite{Gael2009}, another \gls{bnp} model. \Gls{mibp} was used as a model for blind source separation, where multiple sources contribute toward an audio signal, and can appear, persist, disappear and reappear independently. 
    The prior in \gls{SQAIR} is similar, but the crucial differences are that \gls{SQAIR} combines the \gls{bnp} prior with flexible neural network models for the dynamics and likelihood, as well as variational learning via amortized inference.
    The interface between deep learning and \gls{bnp}, and graphical models in general, remains a fertile area of research.
\end{description}
\section{Discussion}
\label{sec:discussion}

In this paper we proposed \gls{SQAIR}, a probabilistic model that extends \gls{AIR} to image sequences, and thereby achieves temporally consistent reconstructions and samples. In doing so, we enhanced \gls{AIR}'s capability of disentangling overlapping objects and identifying partially observed objects.

This work continues the thread of
\cite{Greff2017neuralem}, \cite{Steenkiste2018relationalnem} and, together with \cite{Hsieh2018ddpae}, presents unsupervised object detection~\&~tracking with learnable likelihoods by the means of generative modelling of objects.
In particular, our work is the first one to explicitly model object presence, appearance and location through time. 
Being a generative model, \gls{SQAIR} can be used for conditional generation, where it can extrapolate sequences into the future.
As such, it would be interesting to use it in a reinforcement learning setting in conjunction with Imagination-Augmented Agents \citep{weber2017imagination} or more generally as a world model \citep{ha2018worldm}, especially for settings with simple backgrounds,\eg games like Montezuma's Revenge or Pacman.

The framework offers various avenues of further research; 
\Gls{SQAIR} leads to interpretable representations, but the interpretability of \textit{what} variables can be further enhanced by using alternative objectives that disentangle factors of variation in the objects \citep{kim2018disentangling}. 
Moreover, in its current state, \gls{SQAIR} can work only with simple backgrounds and static cameras. In future work, we would like to address this shortcoming, as well as speed up the sequential inference process whose complexity is linear in the number of objects. The generative model, which currently assumes additive image composition, can be further improved by\eg autoregressive modelling \citep{oord2016cond}. It can lead to higher fidelity of the model and improved handling of occluded objects. Finally, the \gls{SQAIR} model is very complex, and it would be useful to perform a series of ablation studies to further investigate the roles of different components.
\section*{Acknowledgements}

We would like to thank Ali Eslami for his help in implementing \gls{AIR}, Alex Bewley and Martin Engelcke for discussions and valuable insights and anonymous reviewers for their constructive feedback. Additionally, we acknowledge that HK and YWT's research leading to these results has received funding from the European Research Council under the European Union's Seventh Framework Programme (FP7/2007-2013) ERC grant agreement no. 617071.

\printbibliography

\newpage
\appendix

\newpage
\section{Algorithms}
\label{app:algo}

Image generation, described by \Cref{algo:air_decoding}, is exactly the same for \gls{SQAIR} and \gls{AIR}.
\Cref{algo:sqair_prop,algo:sqair_disc} describe inference in \gls{SQAIR}. Note that \gls{disc} is equivalent to \gls{AIR} if no latent variables are present in the inputs.

If a function has multiple inputs and if not stated otherwise, all the inputs are concatenated and linearly projected into some fixed-dimensional space,\eg\Cref{l:rel_rnn,l:q_pres} in \Cref{algo:sqair_prop}.
Spatial Transformer ($\STN$,\eg \Cref{l:prop_glimpse} in \Cref{algo:sqair_prop}) has no learnable parameters: it samples a uniform grid of points from an image $\bx$, where the grid is transformed according to parameters $\bz^\mathrm{where}$.
$\operatorname{f_\phi^1}$~is implemented as a perceptron with a single hidden layer.
Statistics of $\qp$ and $\qd$ are a result of applying a two-layer \gls{MLP} to their respective conditioning sets.
Different distributions $\q$ do not share parameters of their \glspl{MLP}.
The \textit{glimpse encoder} $\operatorname{h_\phi^\mathrm{glimpse}}$ (\Cref{l:glimpse_enc_prop,l:glimpse_enc_final} in \Cref{algo:sqair_prop} and \Cref{l:glimpse_enc_disc} in \Cref{algo:sqair_disc}; they share parameters) and the \textit{image encoder} $\operatorname{h_\phi^\mathrm{enc}}$ (\Cref{l:img_encoder} in \Cref{algo:sqair_disc}) are implemented as two-layer \glspl{MLP} or \glspl{CNN}, depending on the experiment (see \Cref{app:mnist_details,app:duke_details} for details).

One of the important details of \gls{prop} is the proposal glimpse extracted in lines \Cref{l:prop_where,l:prop_glimpse} of \Cref{algo:sqair_prop}. 
It has a dual purpose. 
Firstly, it acts as an information bottleneck in \gls{prop}, limiting the flow of information from the current observation $\bxt$ to the updated latent variables $\bzt$.
Secondly, even though the information is limited, it can still provide a high-resolution view of the object corresponding to the currently updated latent variable, \textit{given} that the location of the proposal glimpse correctly predicts motion of this object.
Initially, our implementation used encoding of the raw observation ($\operatorname{h_\phi^\mathrm{enc}} \left( \bxt \right)$, similarly to \Cref{l:img_encoder} in \Cref{algo:sqair_disc}) as an input to the relation-\gls{RNN} (\Cref{l:rel_rnn} in \Cref{algo:sqair_prop}). We have also experimented with other bottlenecks: (1) low resolution image as an input to the image encoder and (2) a low-dimensional projection of the image encoding before the relation-\gls{RNN}. 
Both approaches have led to \textit{ID swaps}, where the order of explaining objects were sometimes swapped for different frames of the sequence (see \Cref{fig:id_swap} in \Cref{app:fail} for an example). 
Using encoded proposal glimpse extracted from a predicted location has solved this issue.

To condition \gls{disc} on propagated latent variables (\Cref{l:latent_enc} in \Cref{algo:sqair_disc}), we encode the latter by using a two-layer \gls{MLP} similarly to \cite{zaheer2017deeps},
\begin{equation}
    \blt = \sum_{i \in \mathcal{P}_t} \operatorname{MLP} \left( \bzt^{\mathrm{what}, i}, \bzt^{\mathrm{where}, i} \right).
\end{equation}
Note that other encoding schemes are possible, though we have experimented only with this one.

\begin{algorithm}
    \caption{Image Generation}
    \label{algo:air_decoding}
    \DontPrintSemicolon
    \SetKwInOut{Input}{Input}
    \SetKwInOut{Output}{Output}
    \SetSideCommentLeft
    \Input{
        $\bzt^\mathrm{what},
         \bzt^\mathrm{where}$ 
         - latent variables from the current time-step.\\
    }
    $\mathcal{O}_t = \operatorname{indices} \left( \bzt^\mathrm{what} \right)$ \tcp*{Indices of all present latent variables.}
    $\byt^0 = \bm{0}$\\
    \For{$i \in \mathcal{O}_t$}{
        $\byt^{\mathrm{att}, i} = \operatorname{f_\theta^\mathrm{dec}} \left( 
            \bzt^{\mathrm{what}, i} 
        \right)$ \tcp*{Decode the glimpse.}
        $\byt^i = \byt^{i-1} + \STN^{-1} \left( 
            \byt^{\mathrm{att}, i}, \bzt^{\mathrm{where}, i} 
        \right)$
    }
    $\hat{\bx}_t \sim \mathcal{N} \left(\bm{x} \mid \bm{y}_n, \sigma^2_x \bm{I} \right)$\\
    \Output{$\hat{\bm{x}}$}
\end{algorithm}

\begin{algorithm}
    \caption{Inference for Propagation}
    \label{algo:sqair_prop}
    \DontPrintSemicolon
    \SetKwInOut{Input}{Input}
    \SetKwInOut{Output}{Output}
    \SetSideCommentLeft
    \Input{$\bm{x_t}$ - image at the current time-step,\\
         $\bz^\mathrm{what}_{t-1},
         \bz^\mathrm{where}_{t-1},
         \bz^\mathrm{pres}_{t-1}$ 
         - latent variables from the previous time-step\\
         $\bm{h}^T_{t-1}$ - hidden states from the previous time-step.
    }
    $\hR{t}{0}, \bm{z}^{\mathrm{what}, 0}_t, \bm{z}^{\mathrm{where}, 0}_t = \operatorname{initialize}()$\\
    $j = 0$ \tcp*{Index of the object processed in the last iteration.}
    \For{$i \in {\mathcal{O}_{t-1}}$}{
        \uIf{$z_{t-1}^{\mathrm{pres}, i} == 0$}{
            \Continue
        }
        $\hat{\bz}_t^{\mathrm{where}, i} = \operatorname{f_\phi^1} \left( \bz_{t-1}^{\mathrm{where}, i},
            \hT{t}{i} \right)$ \label{l:prop_where}\tcp*{Proposal location.}
        $\hat{\bg}^i_t = \STN \left(
            \bxt, \hat{\bz}_t^{\mathrm{where}, i}
        \right)$ \label{l:prop_glimpse}\tcp*{Extract a glimpse from a proposal location.}
    $\hat{\be}_t^i = \operatorname{h_\phi^\mathrm{glimpse}} \left( \hat{\bg}^i_t \right)$ \label{l:glimpse_enc_prop}\tcp*{Encode the proposal glimpse.}
        $\textcolor{orange}{\bwt^{R, i}}, \hR{t}{i} = \Rr \left(
            \hat{\be}_t^i,
            \bz_{t-1}^{\mathrm{what}, i}, \bz_{t-1}^{\mathrm{where}, i}, 
            \hT{t-1}{i}, \hR{t}{j},
            \bzt^{\mathrm{what}, j}, \bzt^{\mathrm{where}, j}
        \right)$  
        \label{l:rel_rnn}\tcp*{Relational state, see \Cref{eq:propagation_state}.}
        $\bzt^{\mathrm{where}, i} \sim \qp{\bm{z}^\mathrm{where}}{\bm{z}^{\mathrm{where}, k}_{t-1}, \textcolor{orange}{\bwt^{R, i}}}{\phi}$\\
        $\bgt^i = \STN \left( 
            \bxt,
            \bzt^{\mathrm{where}, i}
        \right)$ \tcp*{Extract the final glimpse.}
    $\bet^i = \operatorname{h_\phi^\mathrm{glimpse}} \left( \bgt^i \right)$ \label{l:glimpse_enc_final}\tcp*{Encode the final glimpse.}
        $\textcolor{blue}{\bwt^{T, i}}, \hT{t}{i} = \RT \left(
            \bet^i,
            \bzt^{\mathrm{where}, i}, 
            \hT{t-1}{i}, \hR{t}{i}
        \right)$
        \label{l:time_rnn}\tcp*{Temporal state, see \Cref{eq:temporal_state}.}
        $\bzt^{\mathrm{what}, i} \sim \qp {\bz^\mathrm{what}}{\bet^i, \bz_{t-1}^{\mathrm{what}, i}, \textcolor{orange}{\bwt^{R, i}}, \textcolor{blue}{\bwt^{T, i}}}{\phi}$\\
        $\zt^{\mathrm{pres}, i} \sim \qp{z^\mathrm{pres}}{
            z_{t-1}^{\mathrm{pres}, i},
            \bzt^{\mathrm{what}, i},
            \bzt^{\mathrm{where}, i},
            \textcolor{orange}{\bwt^{R, i}},
            \textcolor{blue}{\bwt^{T, i}}}{\phi}$ \label{l:q_pres}\tcp*{\Cref{eq:q_prop_presence}.}
        $j = i$
    }
    \Output{
        $\bzt^{\mathrm{what}, \mathcal{P}_t},
        \bzt^{\mathrm{where}, \mathcal{P}_t},
        \bzt^{\mathrm{pres}, \mathcal{P}_t}$
    }
\end{algorithm}

\begin{algorithm}
    \caption{Inference for Discovery}
    \label{algo:sqair_disc}
    \DontPrintSemicolon
    \SetKwInOut{Input}{Input}
    \SetKwInOut{Output}{Output}
    \SetSideCommentLeft
    \Input{
        $\bm{x_t}$ - image at the current time-step,\\
        $\bzt^{\mathcal{P}_t}$ - propagated latent variables for the current time-step,\\
        $N$ - maximum number of inference steps for discovery.
    }
    $\hD{t}{0}, \bm{z}^{\mathrm{what}, 0}_t, \bm{z}^{\mathrm{where}, 0}_t = \operatorname{initialize}()$\\
    $j = \operatorname{max\_index} \left( \bzt^{\mathcal{P}_t} \right)$ \tcp*{Maximum index among the propagated latent variables.}
    $\bet = \operatorname{h_\phi^\mathrm{enc}} \left( \bxt \right)$ \label{l:img_encoder}\tcp*{Encode the image.}
    $\blt = \operatorname{h_\phi^\mathrm{enc}} \left( 
        \bzt^\mathrm{what},
         \bzt^\mathrm{where},
         \bzt^\mathrm{pres} 
     \right)$ \label{l:latent_enc}\tcp*{Encode latent variables.}
    \For{$i \in [j + 1, \dots, j + N]$}{
        $\textcolor{pink}{\bwt^{D, i}}, \hD{t}{i} = \RD \left(
            \bet,
            \blt,
            \bzt^{\mathrm{what}, i-1}, 
            \bzt^{\mathrm{where}, i-1}, 
            \hD{t}{i-1} 
        \right)$\\
        $\zt^{\mathrm{pres}, i} \sim \qd{z^\mathrm{pres}}{ \textcolor{pink}{\bwt^{D, i}}}{\phi}$\\
        \If{$z^{\mathrm{pres}, i} = 0$}{
            break
        }
        $\bzt^{\mathrm{where}, i} \sim \qd{\bz^\mathrm{where}}{\textcolor{pink}{\bwt^{D, i}}}{\phi}$\\
        $\bgt^i = \STN\left( 
            \bxt,
            \bzt^{\mathrm{where}, i} 
        \right)$\\
        $\bet^i = \operatorname{h_\phi^\mathrm{glimpse}} \left( \bgt^i \right)$ \label{l:glimpse_enc_disc}\tcp*{Encode the glimpse.}
        $\bzt^{\mathrm{what}, i} \sim \qd{\bzt^\mathrm{what}}{\bet^i}{\phi}$
    }
    \Output{
       $\bzt^{\mathrm{what}, \mathcal{D}_t},
        \bzt^{\mathrm{where}, \mathcal{D}_t},
        \bzt^{\mathrm{pres}, \mathcal{D}_t}$
    }
\end{algorithm}
\newpage
\section{Details for the Generative Model of SQAIR} 
\label{apd:sqair_generation}

In implementation, we upper bound the number of objects at any given time by $N$.
In detail, the discovery prior is given by
\begin{equation} \label{eq:disc_prior}
    \pd{D_t,\bzt^{\mathcal{D}_t}}{\bzt^{\mathcal{P}_t}}{} 
     = \pd{D_t}{P_t}{} \prod_{i \in \mathcal{D}_t} \pd(\bzt^{\mathrm{what}, i})\pd(\bzt^{\mathrm{where}, i}) \delta_1(\zt^{\mathrm{pres},i}),
\end{equation}
\begin{equation} \label{eq:disc_step_prior}
    \pd{D_t}{P_t}{}=\cat \left(D_t; N-P_t,p_{\theta}(P_t)\right),
\end{equation}
where $\delta_x(\cdot)$ is the delta function at $x$, $\cat(k;K,p)$ implies $k \in \{0, 1, \ldots ,K\}$ with probabilities $p_0,p_1,\ldots,p_K$ and $\pd(\bzt^{\mathrm{what}, i}),\pd(\bzt^{\mathrm{where}, i})$ are fixed isotropic Gaussians.
The propagation prior is given by 
\begin{equation} \label{eq:prop_prior}
    \pp{\bzt^{\mathcal{P}_t}}{\bz_{t-1}}{}
     = \prod_{i \in \mathcal{P}_t} \pp{\bzt^{\mathrm{pres},i}}{\bz_{t-1}^{\mathrm{pres},i}, \bm{h}_{t-1}}{} \pp{\bzt^{\mathrm{what}, i}}{\bm{h}_{t-1}}{} \pp{\bzt^{\mathrm{where}, i}}{\bm{h}_{t-1}},
\end{equation}
\begin{equation} \label{eq:prop_pres_prior}
    \pp{\bzt^{\mathrm{pres},i}}{\bz_{t-1}^{\mathrm{pres},i}, \bm{h}_{t-1}}{}
     = \bern (\zt^{\mathrm{pres},i}; f_{\theta}(\bm{h}_{t-1})) \delta_1 (z_{t-1}^{\mathrm{pres},i}),
\end{equation}
with $f_{\theta}$ a scalar-valued function with range $[0,1]$ and $\pp(\bzt^{\mathrm{what}, i}|\bm{h}_{t-1}) $, $\pp(\bzt^{\mathrm{where}, i}|\bm{h}_{t-1})$ both factorised Gaussians parameterised by some function of $\bm{h}_{t-1}$.

\section{Details for the Inference of SQAIR} 
\label{apd:sqair_inference}

The propagation inference network $\qp_{\phi}$ is given as below,
\begin{equation} \label{eq:q_prop}
    \qp{\bzt^{\mathcal{P}_t}}{\bxt, \bz_{t-1}, \hT{t}{\mathcal{P}_t}}{\phi}
        = \prod_{i \in {\mathcal{O}_{t-1}}} \qp{\bzt^i}{\bxt, \bz_{t-1}^{i}, \hT{t}{i}, \hR{t}{i}}{\phi},
\end{equation}
with $\hR{t}{i}$ the hidden state of the relation \gls{RNN} (see \Cref{eq:propagation_state}). Its role is to capture information from the observation $\bxt$ as well as to model dependencies between different objects.
The propagation posterior for a single object can be expanded as follows,
\begin{equation} \label{eq:q_prop_obj}
\begin{aligned}
    &\qp{\bzt^i}{\bxt,\bz_{t-1}^{i}, \hT{t}{i}, \hR{t}{i}}{\phi} =\\
        &\quad \qp{\bzt^{\mathrm{where}, i}}{\bz_{t-1}^{\mathrm{what}, i}, \bz_{t-1}^{\mathrm{where}, i}, \hT{t-1}{i}, \hR{t}{i}}{\phi}\\
        &\quad \qp{\bzt^{\mathrm{what}, i}}{\bxt, \bzt^{\mathrm{where}, i}, \bz_{t-1}^{\mathrm{what}, i}, \hT{t}{i}, \hR{t}{i}}{\phi}\\
        &\quad \qp{\zt^{\mathrm{pres}, i}}{\bzt^{\mathrm{what}, i}, \bzt^{\mathrm{where}, i}, z_{t-1}^{\mathrm{pres}, i}, \hT{t}{i}, \hR{t}{i}}{\phi}.
\end{aligned}
\end{equation}
In the second line, we condition the object location $\bzt^{\mathrm{where}, i}$ on its previous appearance and location as well as its dynamics and relation with other objects.
In the third line, current appearance $\bzt^{\mathrm{what}, i}$ is conditioned on the new location.
Both $\bzt^{\mathrm{where}, i}$ and $\bzt^{\mathrm{what}, i}$ are modelled as factorised Gaussians.
Finally, presence depends on the new appearance and location as well as the presence of the same object at the previous time-step.
More specifically, 
\begin{equation} \label{eq:q_prop_presence}
\begin{aligned}
    &\qp{\zt^{\mathrm{pres}, i}}{\bzt^{\mathrm{what}, i}, \bzt^{\mathrm{where}, i}, z_{t-1}^{\mathrm{pres}, i}, \hT{t}{i}, \hR{t}{i}}{\phi}\\
        &\quad= \bern \left(
            \zt^{\mathrm{pres}, i} \mid f_\phi \left(
                \bzt^{\mathrm{what}, i}, \bzt^{\mathrm{where}, i}, \hT{t}{i}, \hR{t}{i}
            \right)
        \right)
        \delta_1( z_{t-1}^{\mathrm{pres}, i}),
\end{aligned}
\end{equation}
where the second term is the delta distribution centered on the presence of this object at the previous time-step.
If it was not there, it cannot be propagated.
Let $j \in \{0, \dots, i-1\}$ be the index of the most recent present object before object $i$. Hidden states are updated as follows,
\begin{equation} \label{eq:propagation_state}
        \hR{t}{i} = \Rr \left(
        \bxt,
        \bz_{t-1}^{\mathrm{what}, i}, \bz_{t-1}^{\mathrm{where}, i}, 
        \hT{t-1}{i}, \hR{t}{i-1},
        \bzt^{\mathrm{what}, j}, \bzt^{\mathrm{where}, j}
    \right),
\end{equation}
\begin{equation} \label{eq:temporal_state}
    \hT{t}{i} = \RT \left(
        \bxt,
        \bzt^{\mathrm{where}, i}, 
        \hT{t-1}{i}, \hR{t}{i}
    \right),
\end{equation}
where $\RT$ and $\Rr$ are temporal and propagation \glspl{RNN}, respectively.
Note that in \cref{eq:propagation_state} the \gls{RNN} does not have direct access to the image $\bxt$, but rather accesses it by extracting an attention glimpse at a proposal location, predicted from $\hT{t-1}{i}$ and $\bz_{t-1}^{\mathrm{where}, i}$.
This might seem like a minor detail, but in practice structuring computation this way prevents ID swaps from occurring, \textit{cf}. \Cref{app:fail}.
For computational details, please see \Cref{algo:sqair_prop,algo:sqair_disc} in \Cref{app:algo}.
\section{Details of the moving-\textsc{mnist} Experiments}
\label{app:mnist_details}

\subsection{\textsc{Sqair} and \textsc{air} Training Details}
    All models are trained by maximising the \gls{ELBO} $\loss[IWAE]$ (\Cref{eq:iwae}) with the \gls{RMSprop} optimizer \citep{tieleman2012rms} with momentum equal to $0.9$. We use the learning rate of $10^{-5}$ and decrease it to $\frac{1}{3} \cdot 10^{-5}$ after 400k and to $10^{-6}$ after 1000k training iterations. Models are trained for the maximum of $2 \cdot 10^6$ training iterations; we apply early stopping in case of overfitting.
    \Gls{SQAIR} models are trained with a curriculum of sequences of increasing length: we start with three time-steps, and increase by one time-step every $10^5$ training steps until reaching the maximum length of 10.
    When training \gls{AIR}, we treated all time-steps of a sequence as independent, and we trained it on all data (sequences of length ten, split into ten independent sequences of length one). 

\subsection{\textsc{Sqair} and \textsc{air} Model Architectures}
    All models use glimpse size of $20 \times 20$ and \gls{ELU} \citep{Clevert2015elu} non-linearities for all layers except \glspl{RNN} and output layers.
    \textsc{mlp}-\gls{SQAIR} uses fully-connected layers for all networks. In both variants of \gls{SQAIR}, the $\RD$ and $\Rr$ \glspl{RNN} are the vanilla \glspl{RNN}. The propagation prior \gls{RNN} and the temporal \gls{RNN} $\RT$ use \gls{GRU}. \gls{AIR} follows the same architecture as \textsc{mlp}-\gls{SQAIR}. All fully-connected layers and \glspl{RNN} in \textsc{mlp}-\gls{SQAIR} and \gls{AIR} have 256 units; they have $2.9$M and $1.7$M trainable parameters, respectively.
    
    \textsc{Conv}-\gls{SQAIR} differs from the  \textsc{mlp} version in that it uses \glspl{CNN} for the glimpse and image encoders and a subpixel-\gls{CNN} \citep{shi2016subpixel} for the glimpse decoder. All fully connected layers and \glspl{RNN} have 128 units. 
    The encoders share the \gls{CNN}, which is followed by a single fully-connected layer (different for each encoder).
    The \gls{CNN} has four convolutional layers with $[16,32,32,64]$ features maps and strides of $[2,2,1,1]$. The glimpse decoder is composed of two fully-connected layers with $[256, 800]$ hidden units, whose outputs are reshaped into $32$ features maps of size $5 \times 5$, followed by a subpixel-\gls{CNN} with three layers of $[32,64,64]$ feature maps and strides of $[1, 2, 2]$.
    All filters are of size $3 \times 3$.
    \textsc{Conv}-\gls{SQAIR} has $2.6$M trainable parameters.
    
    We have experimented with different sizes of fully-connected layers and \glspl{RNN}; we kept the size of all layers the same and altered it in increments of 32 units. Values greater than 256 for \textsc{mlp}-\gls{SQAIR} and 128 for \textsc{conv}-\gls{SQAIR} resulted in overfitting. Models with as few as 32 units per layer ($<0.9$M trainable parameters for \textsc{mlp}-\gls{SQAIR}) displayed the same qualitative behaviour as reported models, but showed lower quantitative performance.
    
    The output likelihood used in both \gls{SQAIR} and \gls{AIR} is Gaussian with a fixed standard deviation set to $0.3$, as used by \cite{Eslami2016}. We tried using a learnable scalar standard deviation, but decided not to report it due to unsable behaviour in the early stages of training. Typically, standard deviation would converge to a low value early in training, which leads to high penalties for reconstruction mistakes. In this regime, it is beneficial for the model to perform no inference steps ($z^\mathrm{pres}$ is always equal to zero), and the model never learns. Fixing standard deviation for the first $10$k iterations and then learning it solves this issue, but it introduces unnecessary complexity into the training procedure.

\subsection{\textsc{Vrnn} Implementation and Training Details} \label{apd:vrnn}

    Our \gls{VRNN} implementation is based on the implementation\footnote{\url{https://github.com/tensorflow/models/tree/master/research/fivo}} of \gls{FIVO} by \cite{maddison2017filtering}.
    We use an \textsc{lstm} with hidden size $J$ for the deterministic backbone of the \gls{VRNN}.
    At time $t$, the \textsc{lstm} receives $\psi^{x}(\bx_{t-1})$ and $\psi^{z}(\bz_{t-1})$ as input and outputs $o_t$, where $\psi^{x}$ is a data feature extractor and $\psi^{z}$ is a latent feature extractor. 
    The output is mapped to the mean and standard deviation of the Gaussian prior $\p{\bzt}{\bx_{t-1}}{\theta}$ by an \gls{MLP}. 
    The likelihood $\p{\bxt}{\bzt,\bx_{t-1}}{\theta}$ is a Gaussian, with mean given by $\psi^{\mathrm{dec}}(\psi^{z}(\bzt), o_t)$ and standard deviation fixed to be $0.3$ as for \gls{SQAIR} and \gls{AIR}. 
    The inference network $\q{\bzt}{\bz_{t-1},\bxt}{\phi}$ is a Gaussian with mean and standard deviation given by the output of separate \gls{MLP}s with inputs $[o_t,\psi^{x}(\bxt)]$.
    
    All aforementioned \glspl{MLP} use the same number of hidden units $H$ and the same number of hidden layers $L$.
    The \textsc{conv}-\gls{VRNN} uses a \gls{CNN} for $\psi^{x}$ and a transposed \gls{CNN} for $\psi^{\mathrm{dec}}$. The \textsc{mlp}-\gls{VRNN} uses an \gls{MLP} with $H'$ hidden units  and $L'$ hidden layers for both. 
    \Gls{ELU} were used throughout as activations. The latent dimensionality was fixed to 165, which is the upper bound of the number of latent dimensions that can be used per time-step in \gls{SQAIR} or \gls{AIR}. Training was done by optimising the \gls{FIVO} bound, which is known to be tighter than the \gls{IWAE} bound for sequential latent variable models \citep{maddison2017filtering}. We also verified that this was the case with our models on the moving-\textsc{mnist} data. We train with the \gls{RMSprop} optimizer with a learning rate of $10^{-5}$, momentum equal to $0.9$, and training until convergence of test \gls{FIVO} bound.
    
    For each of \textsc{mlp}-\gls{VRNN} and \textsc{conv}-\gls{VRNN}, we experimented with three architectures: small/medium/large. 
    We used $H$=$H'$=$J$=128/256/512 and $L$=$L'$=2/3/4 for \textsc{mlp}-\gls{VRNN}, giving number of parameters of $1.2$M/$2.1$M/$9.8$M. 
    For \textsc{conv}-\gls{VRNN}, the number of features maps we used was $[32,32,64,64]$, $[32,32,32,64,64,64]$ and $[32,32,32,64,64,64,64,64,64]$, with strides of $[2,2,2,2]$, $[1,2,1,2,1,2]$ and $[1,2,1,2,1,2,1,1,1]$, all with $3 \times 3$ filters, $H$=$J$=$128$/$256$/$512$ and $L$=1, giving number of parameters of $0.8$M/$2.6$M/$6.1$M. 
    The largest convolutional encoder architecture is very similar to that in \cite{gulrajani2016pixelvae} applied to \gls{MNIST}.
    
    We have chosen the medium-sized models for comparison with \gls{SQAIR} due to overfitting encountered in larger models.

\begin{table}
    \centering
    \caption{Number of trainable parameters for the reported models.}
    \label{tab:num_params}
    \begin{tabular}{c|c|c|c|c|c}
         & \textsc{conv}-\gls{SQAIR} & \textsc{mlp}-\gls{SQAIR} & \textsc{mlp}-\gls{AIR} & \textsc{conv}-\gls{VRNN} & \textsc{mlp}-\gls{VRNN}\\
                         \hline
         number of parameters & $2.6$M & $2.9$M & $1.7$M & $2.6$M  & $2.1$M 
    \end{tabular}
\end{table}

\subsection{Addition Experiment}

We perform the addition experiment by feeding latent representations extracted from the considered models into a 19-way classifier, as there are 19 possible outputs (addition of two digits between 0 and 9). 
The classifier is implemented as an \gls{MLP} with two hidden layers with 256 \gls{ELU} units each and a softmax output. For \gls{AIR} and \gls{SQAIR}, we use concatenated $\bz^\mathrm{what}$ variables multiplied by the corresponding $z^\mathrm{pres}$ variables, while for \gls{VRNN} we use the whole 165-dimensional latent vector.
We train the model over $10^7$ training iterations with the \gls{ADAM} optimizer \citep{kingma2015adam} with default parameters (in tensorflow).
\section{Details of the \textit{DukeMTMC} Experiments}
\label{app:duke_details}

We take videos from cameras one, two, five, six and eight from the \textit{DukeMTMC} dataset \citep{ristani2016performance}. As pre-processing, we invert colors and subtract backgrounds using standard OpenCV tools \citep{itseez2015opencv}, downsample to the resolution of $240 \times 175$, convert to gray-scale and randomly crop fragments of size $64 \times 64$. Finally, we generate $3500$ sequences of length five such that the maximum number of objects present in any single frame is three and we split them into training and validation sets with the ratio of $9:1$.

We use the same training procedure as for the \gls{MNIST} experiments. The only exception is the learning curriculum, which goes from three to five time-steps, since this is the maximum length of the sequences. 

The reported model is similar to \textsc{conv}-\gls{SQAIR}. We set the glimpse size to $28 \times 12$ to account for the expected aspect ratio of pedestrians. Glimpse and image encoders share a \gls{CNN} with $[16,32,64,64]$ feature maps and strides of $[2,2,2,1]$ followed by a fully-connected layer (different for each encoder). The glimpse decoder is implemented as a two-layer fully-connected network with 128 and 1344 units, whose outputs are reshaped into 64 feature maps of size $7 \times 3$, followed by a subpixel-\gls{CNN} with two layers of $[64, 64]$ feature maps and strides of $[2, 2]$. All remaining fully-connected layers in the model have 128 units. The total number of trainable parameters is $3.5$M.
\section{Harder multi-\textsc{mnist} Experiment}
\label{app:mnist_inout}
\begin{figure}
    \centering
    \includegraphics[width=\linewidth]{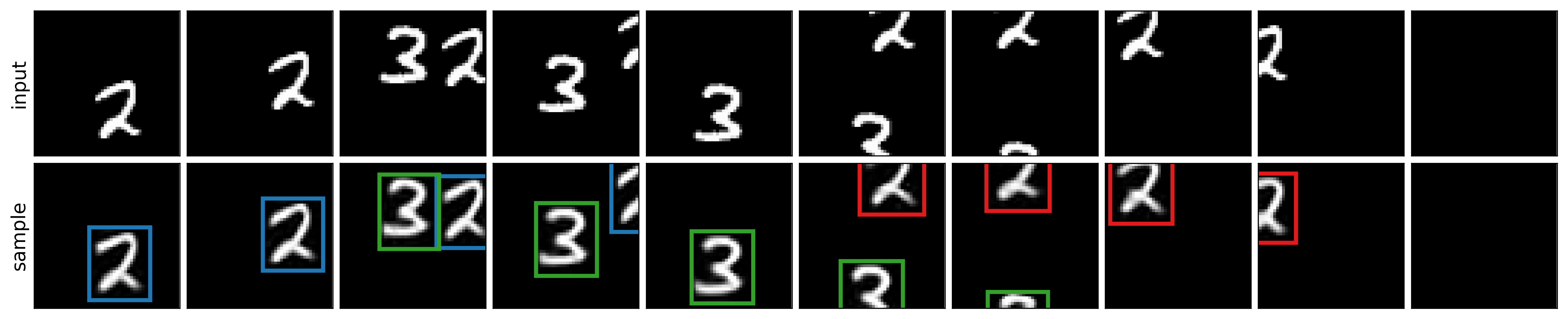}
    \caption{\gls{SQAIR} trained on a harder version of moving-textsc{mnist}. Input images (top) and \gls{SQAIR} reconstructions with marked glimpse locations (bottom)}
    \label{fig:mnist_rec_in_and_out}
\end{figure}

We created a version of the multi-\textsc{mnist} dataset, where objects can appear or disappear at an arbitrary point in time.
It differs from the dataset described in \Cref{sec:expr_mnist}, where all digits are present throughout the sequence.
All other dataset parameters are the same as in \Cref{sec:expr_mnist}.
\Cref{fig:mnist_rec_in_and_out} shows an example sequence and \textsc{mlp}-\gls{SQAIR} reconstructions with marked glimpse locations.
The model has no trouble detecting new digits in the middle of the sequence and rediscovering a digit that was previously present.

\newpage
\section{Failure cases of \textsc{sqair}}
\label{app:fail}

\begin{center}
    \centering
    \begin{minipage}[c]{0.65\linewidth}
        \centering
        \includegraphics[width=\linewidth]{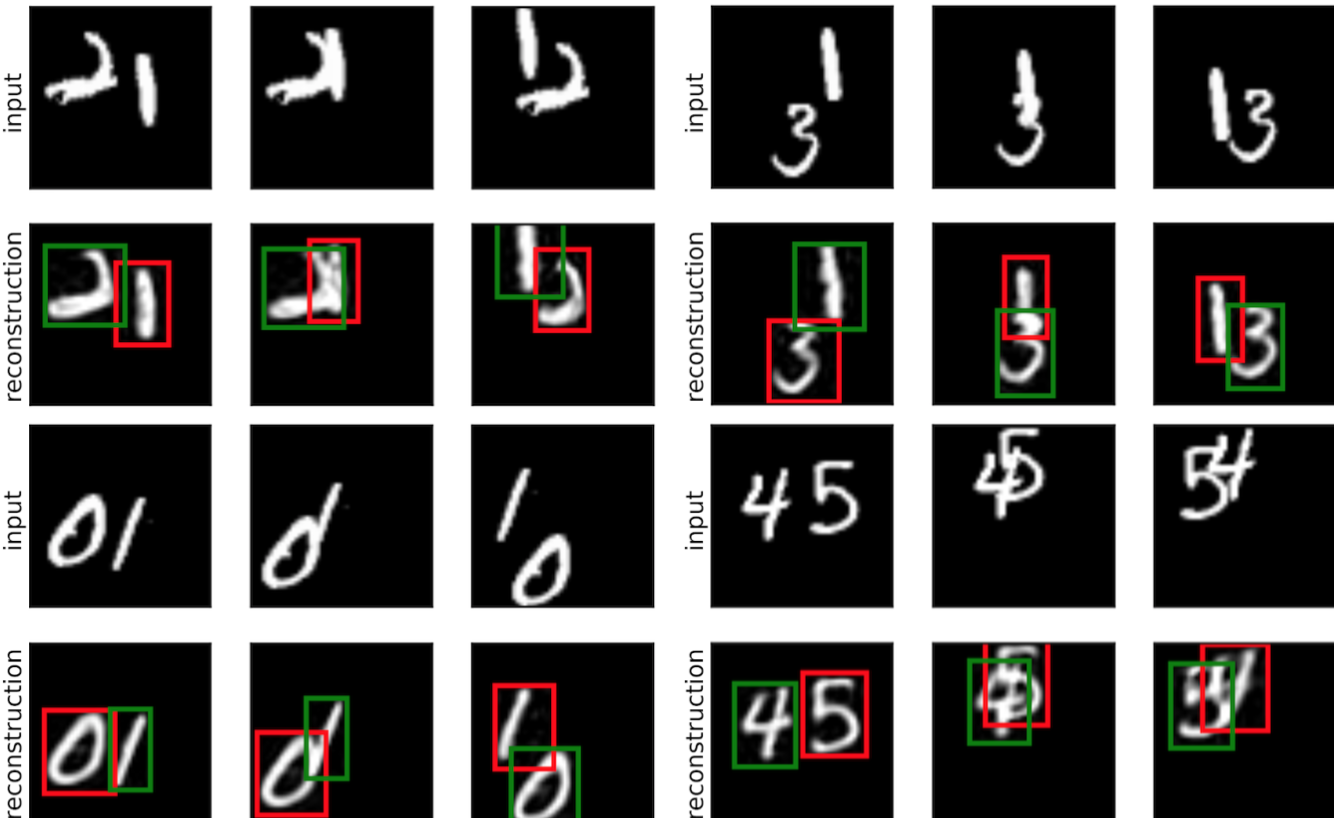}
    \end{minipage}
    \hfill
    \begin{minipage}[c]{0.33\linewidth}
    \centering
        \captionof{figure}{Examples of ID swaps in a version of \gls{SQAIR} \textit{without} proposal glimpse extraction in \gls{prop} (see \Cref{app:algo} for details). Bounding box colours correspond to object index (or its identity). When \gls{prop} is allowed the same access to the image as \gls{disc}, then it often prefers to ignore latent variables, which leads to swapped inference order.}
        \label{fig:id_swap}
    \end{minipage}
\end{center}
\begin{center}
    \centering
    \begin{minipage}[c]{0.65\linewidth}
        \centering
        \includegraphics[width=\linewidth]{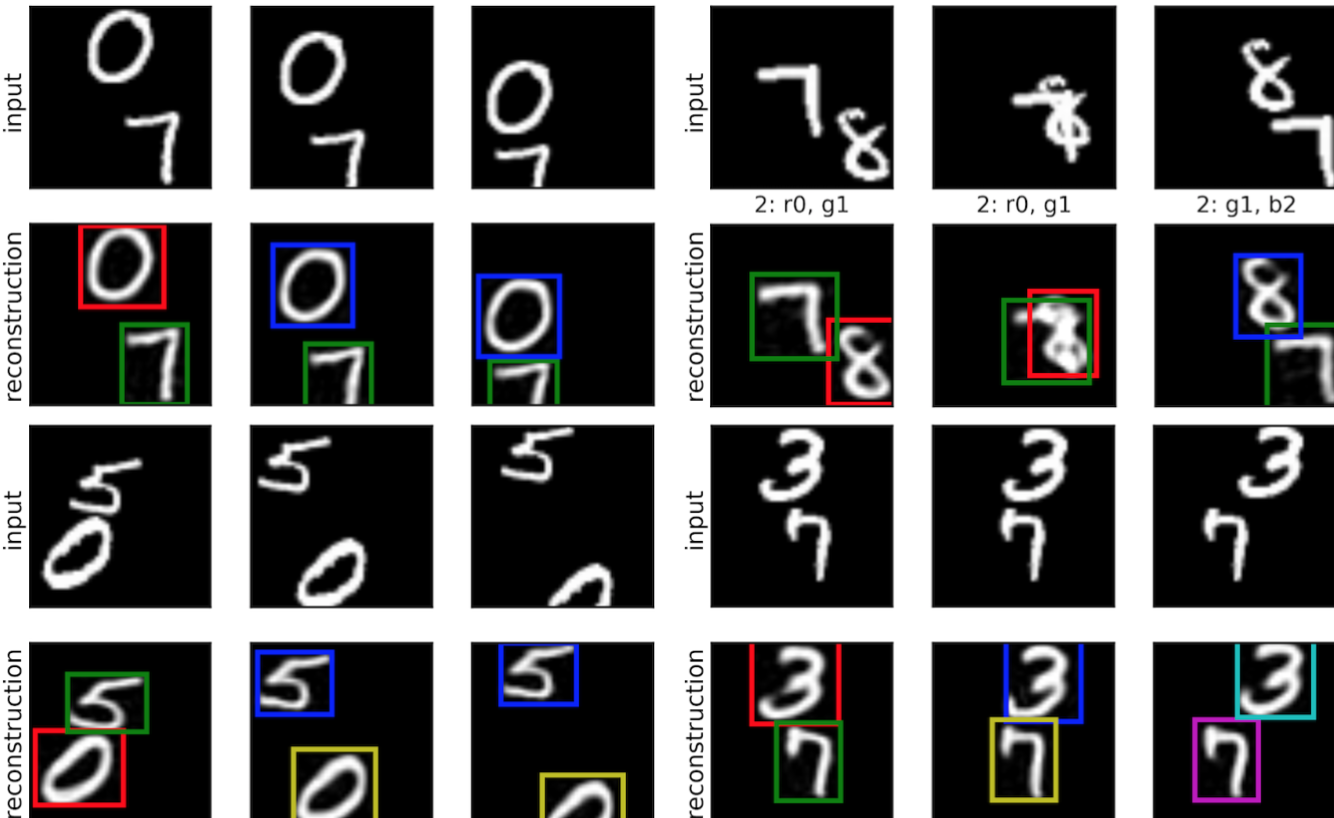}
    \end{minipage}
    \hfill
    \begin{minipage}[c]{0.33\linewidth}
        \centering
        \vspace{8pt}
        \captionof{figure}{Examples of re-detections in \textsc{mlp}-\gls{SQAIR}. Bounding box colours correspond to object identity, assigned to it upon discovery. In some training runs, \gls{SQAIR} converges to a solution, where objects are re-detected in the second frame, and \gls{prop} starts tracking only in the third frame (left). Occasionally, an object can be re-detected after it has severely overlapped with another one (top right). Sometimes the model decides to use only \gls{disc} and repeatedly discovers all objects (bottom right). These failure mode seem to be mutually exclusive -- they come from different training runs.}
        \label{fig:redetect}
    \end{minipage}
\end{center}

\begin{center}
    \includegraphics[width=\linewidth]{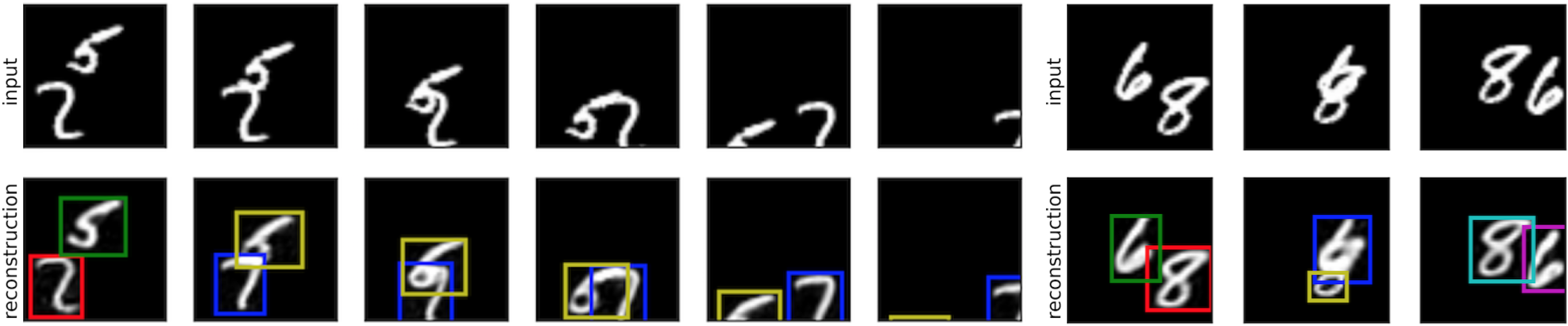}
    \captionof{figure}{Two failed reconstructions of \gls{SQAIR}.
        \textit{Left}:
            \Gls{SQAIR} re-detects objects in the second time-step. Instead of 5 and 2, however, it reconstructs them as 6 and 7. Interestingly, reconstructions are consistent through the rest of the sequence.
        \textit{Right:}
            At the second time-step, overlapping 6 and 8 are explained as 6 and a small 0. The model realizes its mistake in the third time-step, re-detects both digits and reconstructs them properly.
    }
    \label{fig:weird_rec}
\end{center}
\newpage
\section{Reconstruction and Samples from the Moving-MNIST Dataset}
\label{app:mnist_visual}

\subsection{Reconstructions}
\begin{center}

    \includegraphics[width=\linewidth]{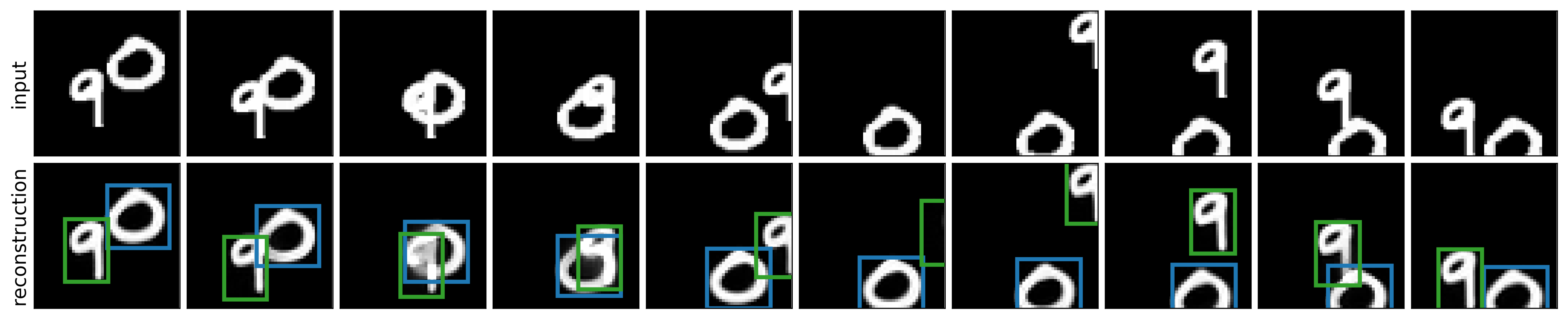}
    \includegraphics[width=\linewidth]{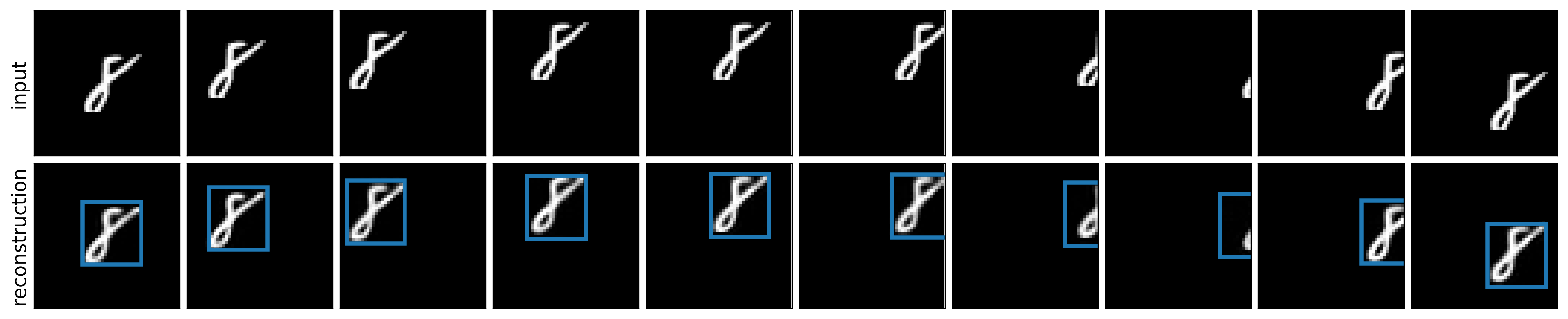}
    \includegraphics[width=\linewidth]{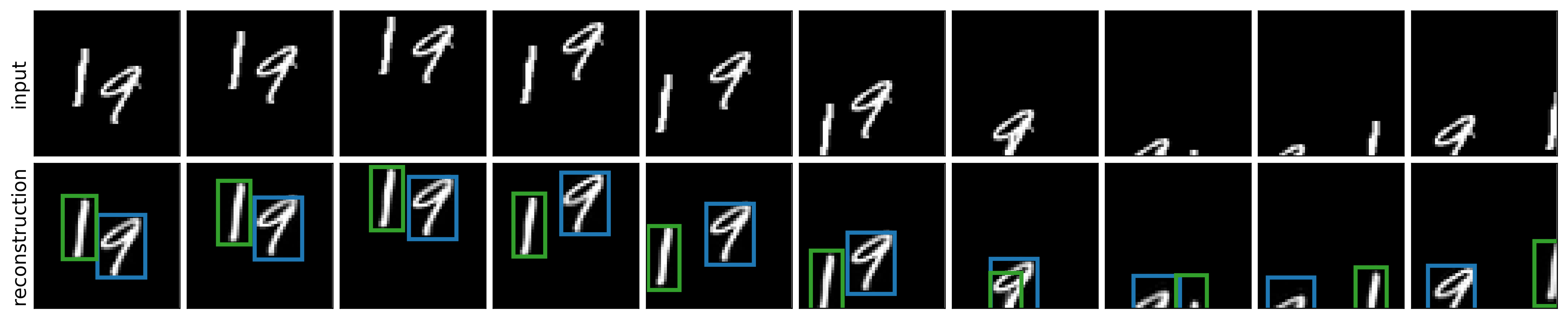}
    \includegraphics[width=\linewidth]{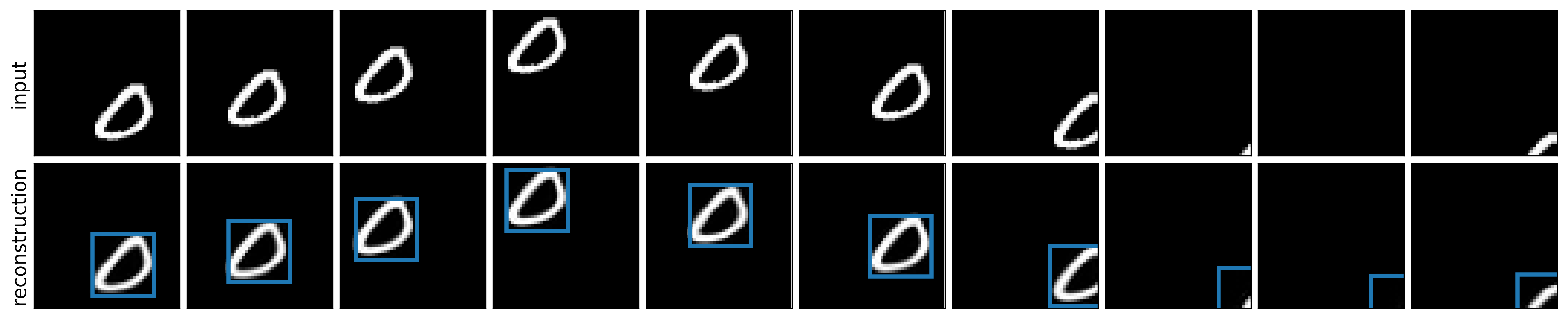}
    \includegraphics[width=\linewidth]{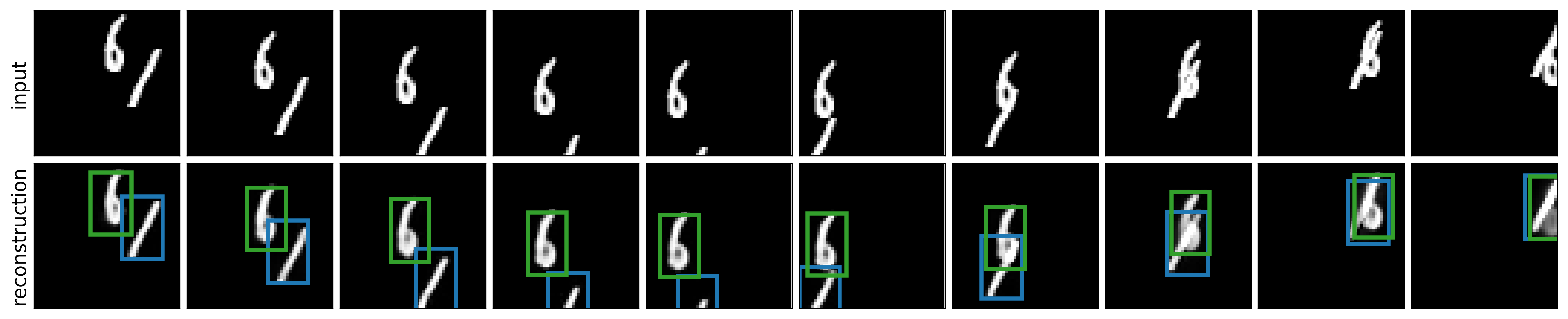}
    \includegraphics[width=\linewidth]{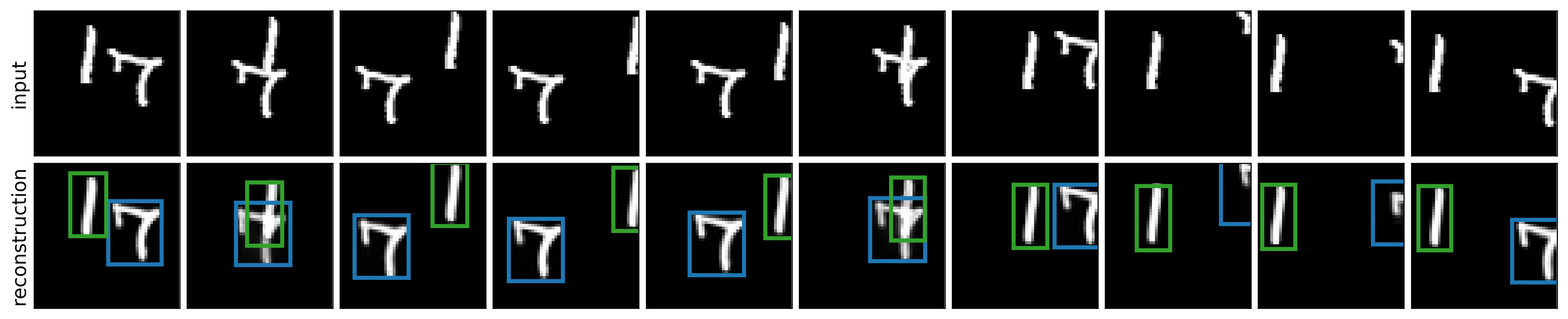}
    \includegraphics[width=\linewidth]{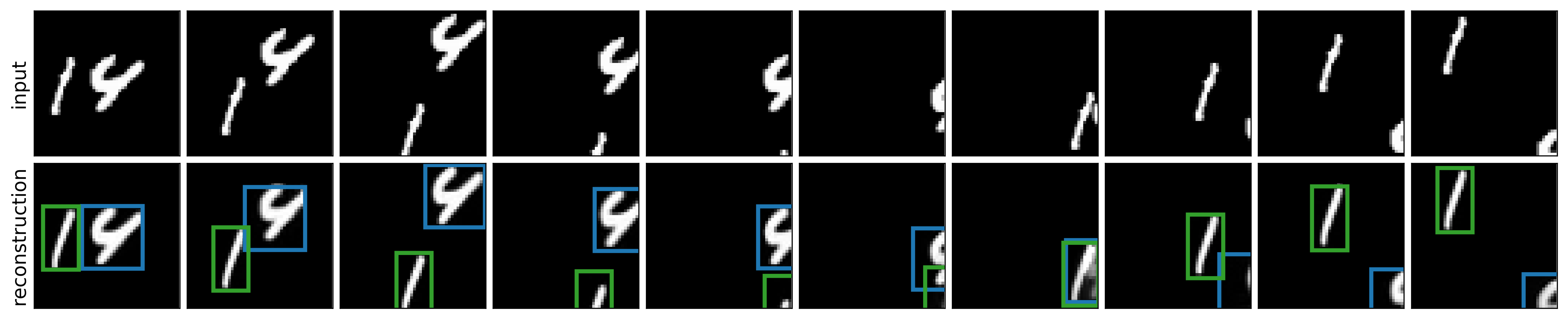} 
    \captionof{figure}{Sequences of input (first row) and \gls{SQAIR} reconstructions with marked glimpse locations. Reconstructions are all temporally consistent.}
    \label{fig:mnist_recs_additional}
\end{center}


\begin{center}
    \includegraphics[width=\linewidth]{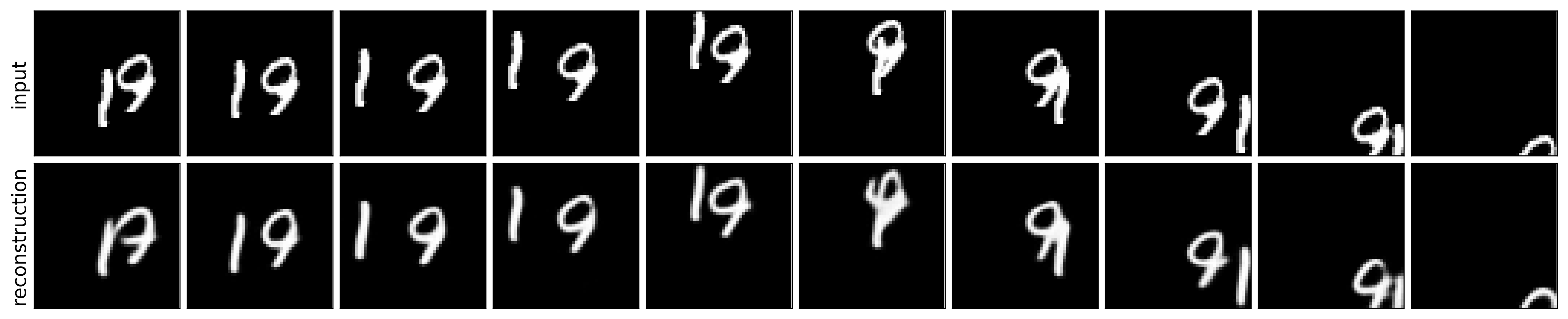}
    \includegraphics[width=\linewidth]{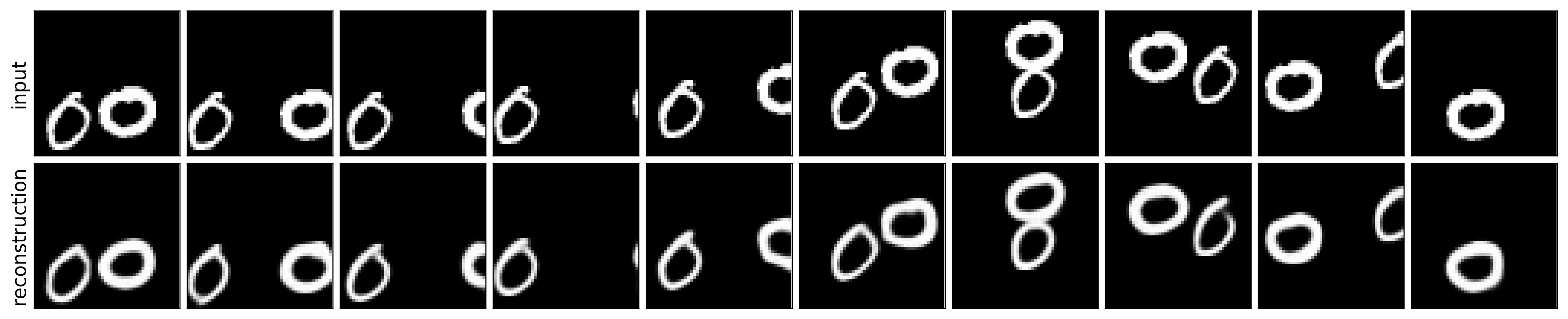}
    \includegraphics[width=\linewidth]{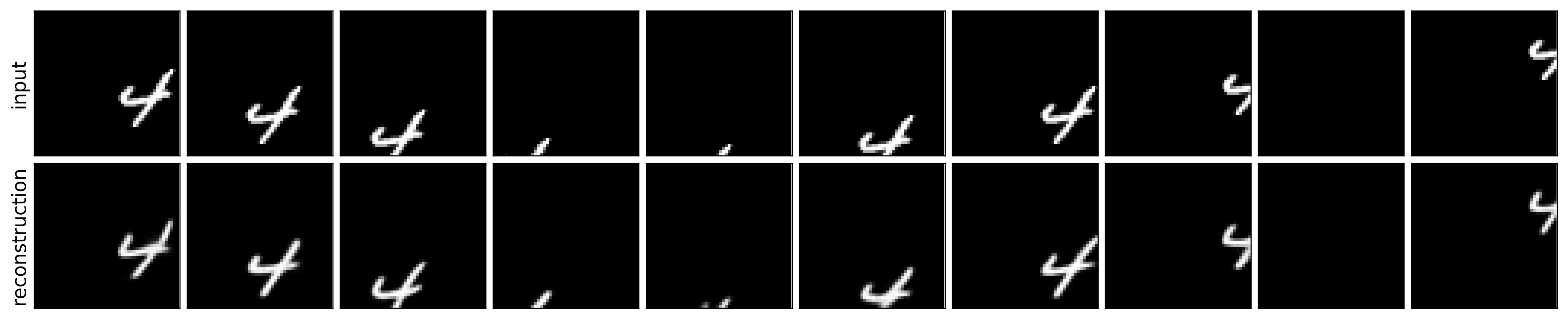}
    \includegraphics[width=\linewidth]{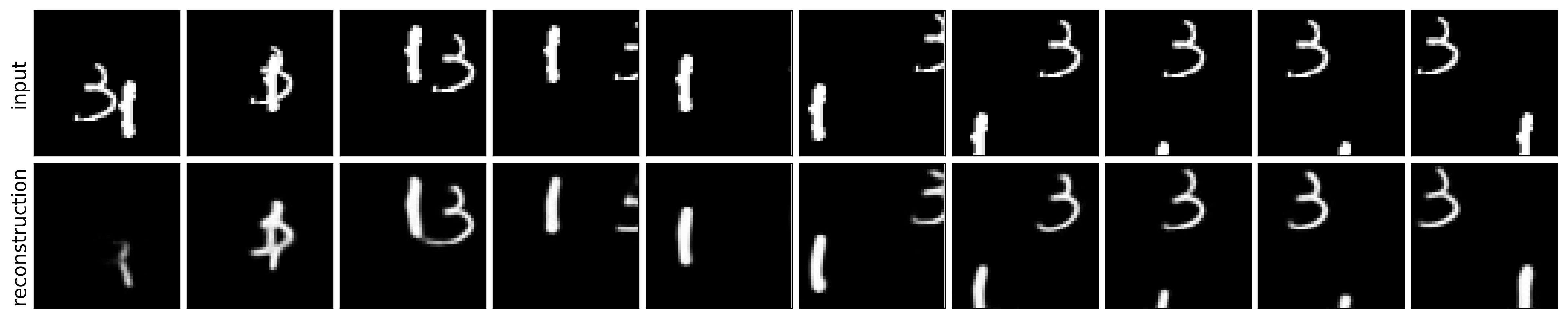}
    \includegraphics[width=\linewidth]{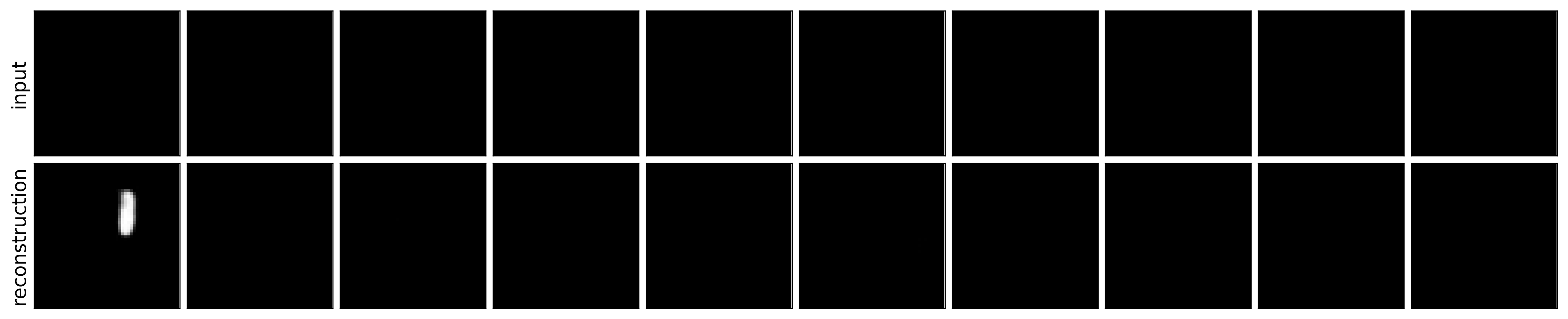}
    \includegraphics[width=\linewidth]{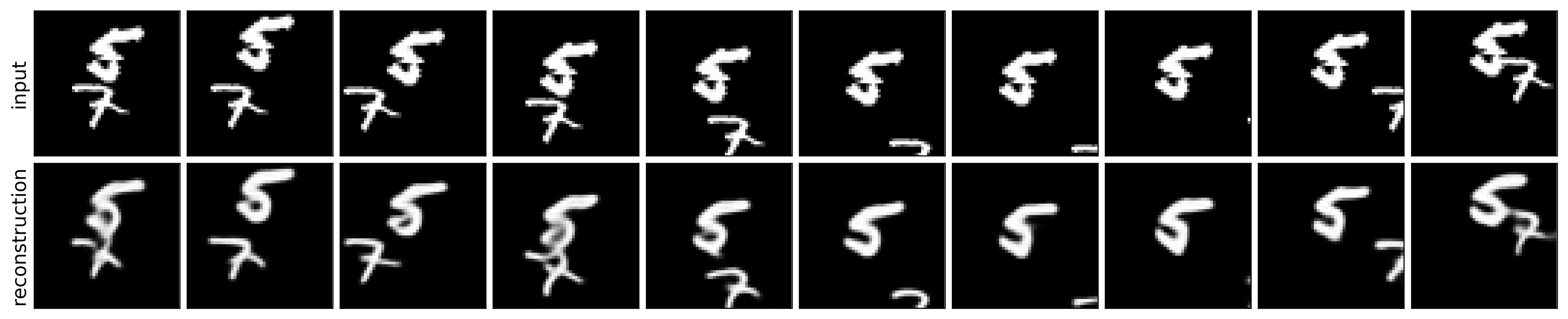}
    \includegraphics[width=\linewidth]{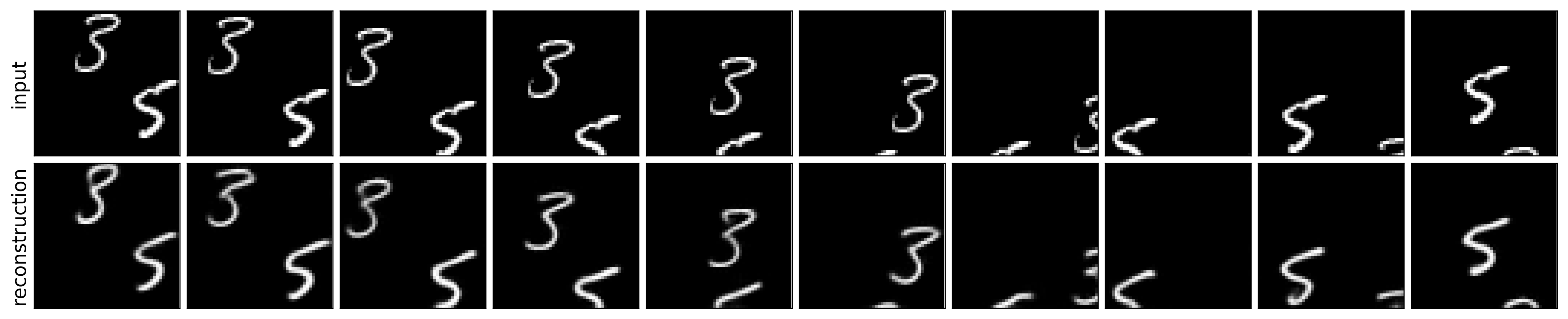}
    \captionof{figure}{Sequences of input (first row) and \textsc{conv}-\gls{VRNN} reconstructions. They are not temporally consistent. The reconstruction at time $t=1$ is typically of lower quality and often different than the rest of the sequence.}
    \label{fig:mnist_recs_vrnn}
\end{center}

\subsection{Samples}
\begin{center}
    \includegraphics[width=\linewidth]{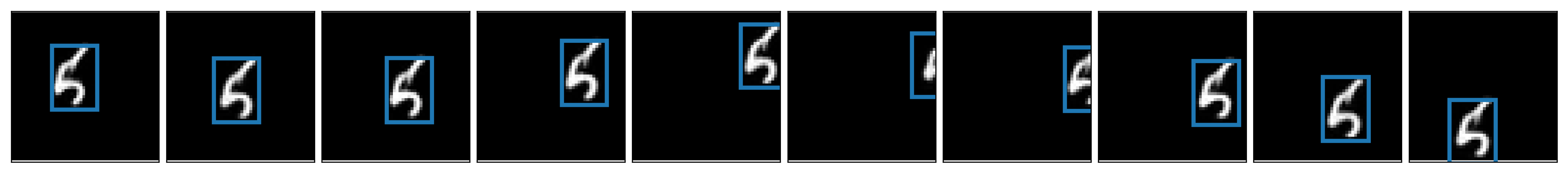}
    \includegraphics[width=\linewidth]{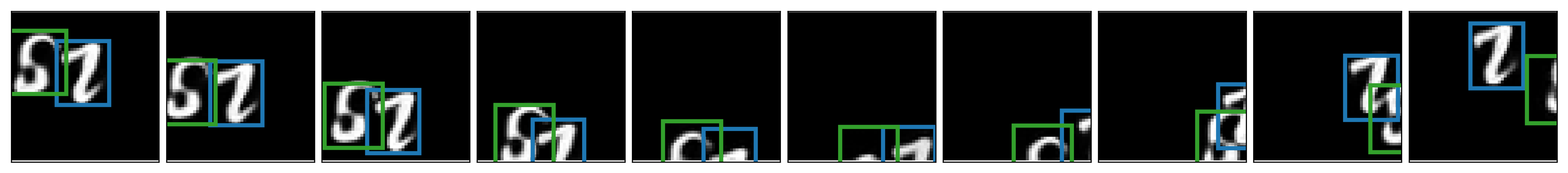}
    \includegraphics[width=\linewidth]{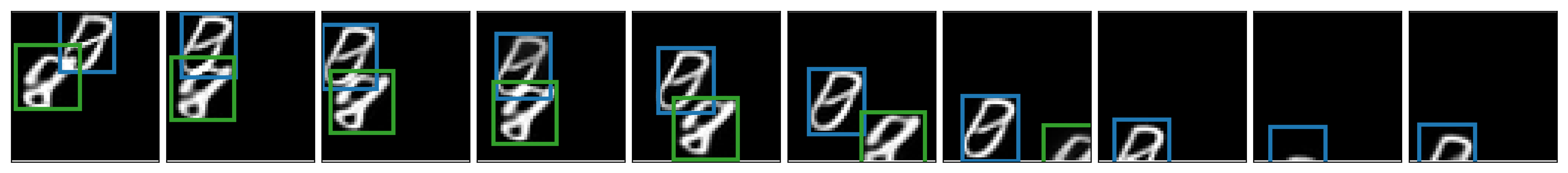}
    \includegraphics[width=\linewidth]{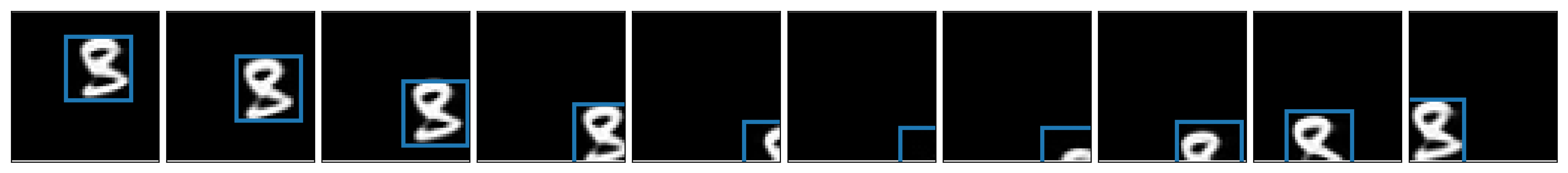}
    \includegraphics[width=\linewidth]{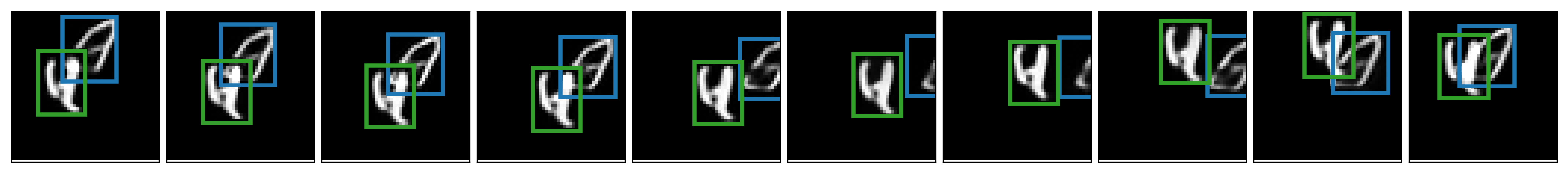}
    \includegraphics[width=\linewidth]{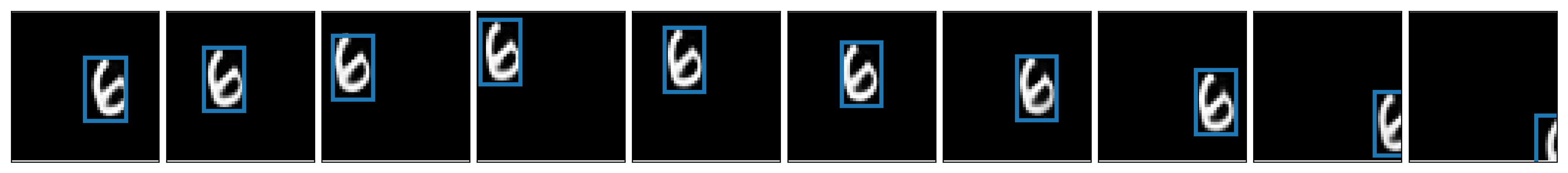}
    \includegraphics[width=\linewidth]{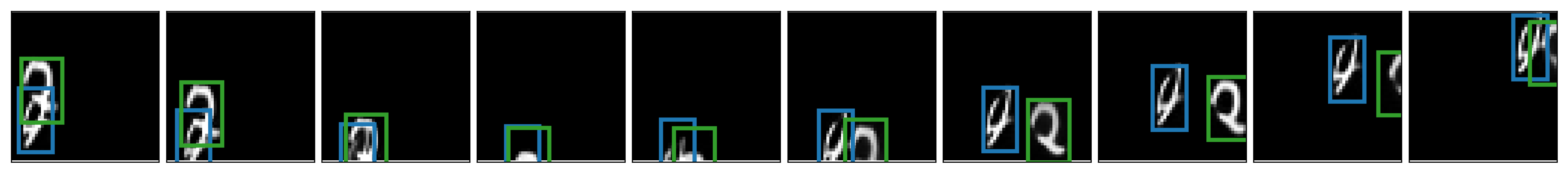}
    \includegraphics[width=\linewidth]{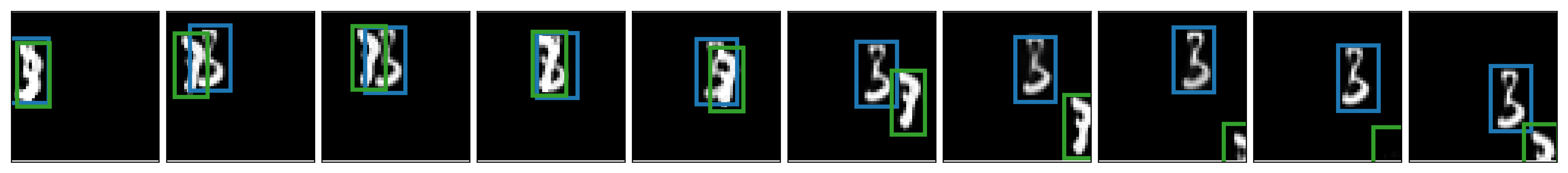}
    \includegraphics[width=\linewidth]{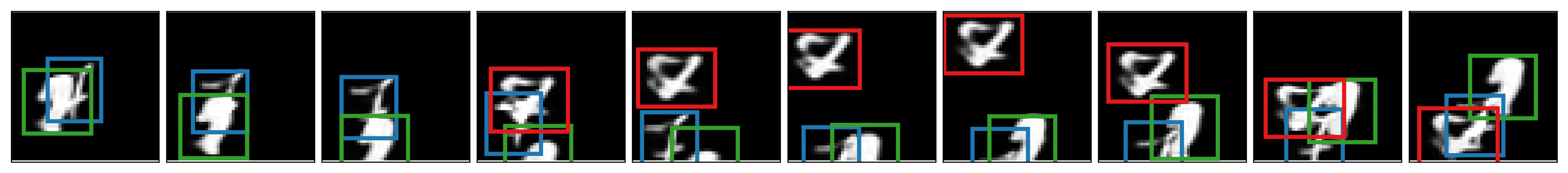}
    \captionof{figure}{Samples from \gls{SQAIR}. Both motion and appearance are temporally consistent. In the last sample, the model introduces the third object despite the fact that it has seen only up to two objects in training.}
    \label{fig:mnist_samples_additional}
\end{center}


\newpage
\begin{center}
    \includegraphics[width=\linewidth]{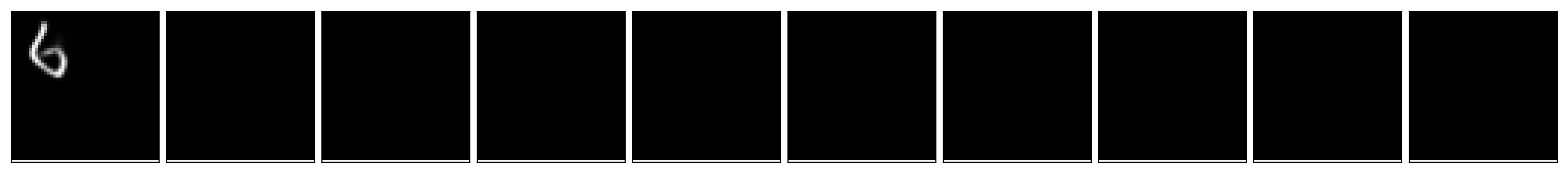}
    \includegraphics[width=\linewidth]{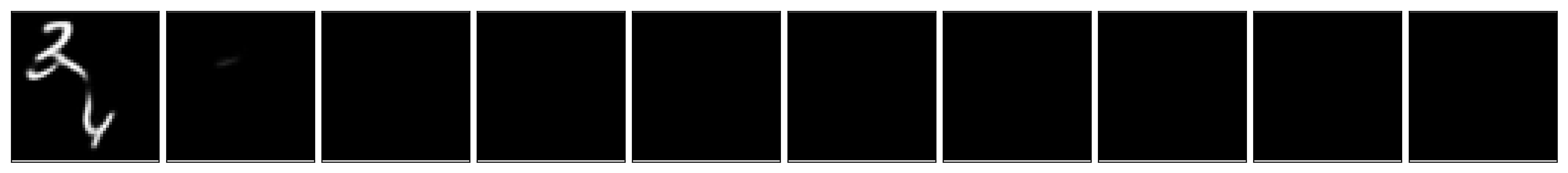}
    \includegraphics[width=\linewidth]{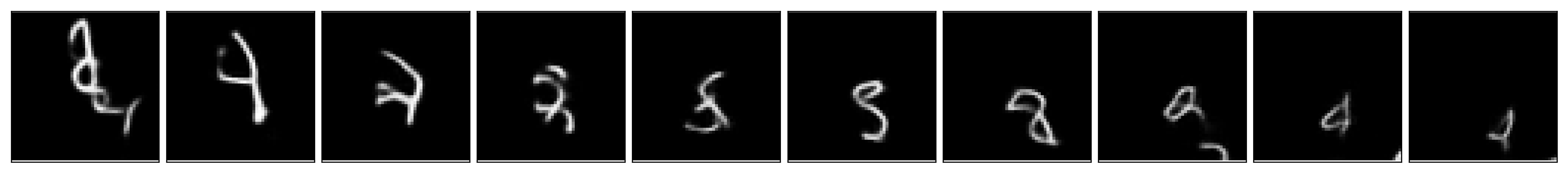}
    \includegraphics[width=\linewidth]{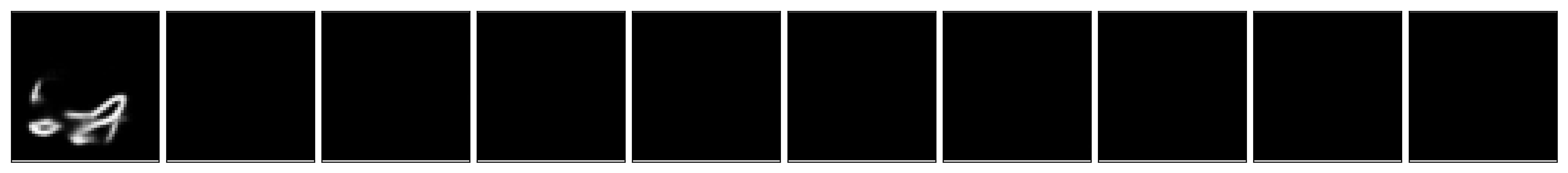}
    \includegraphics[width=\linewidth]{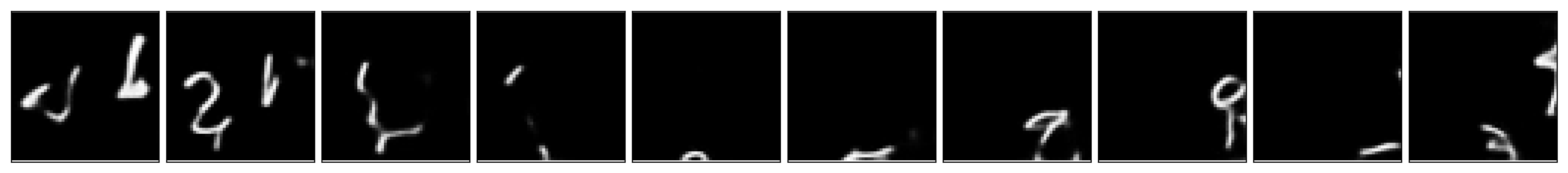}
    \includegraphics[width=\linewidth]{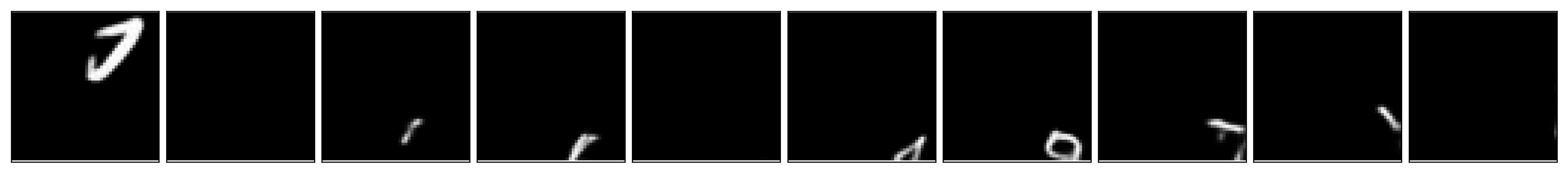}
    \includegraphics[width=\linewidth]{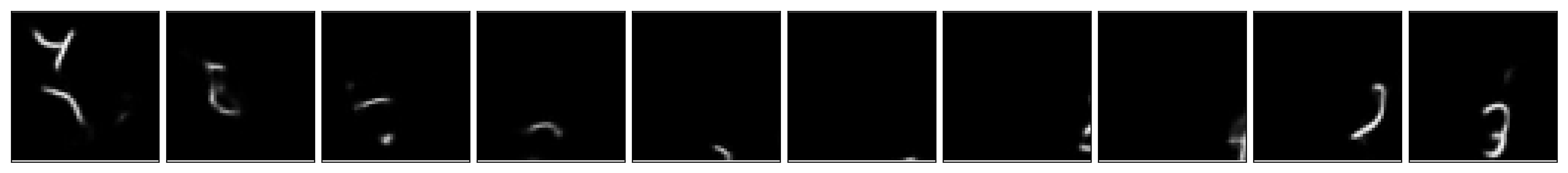}
    \includegraphics[width=\linewidth]{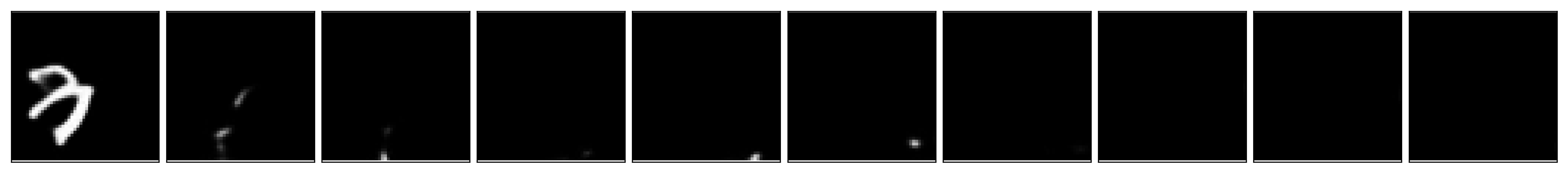}
    \includegraphics[width=\linewidth]{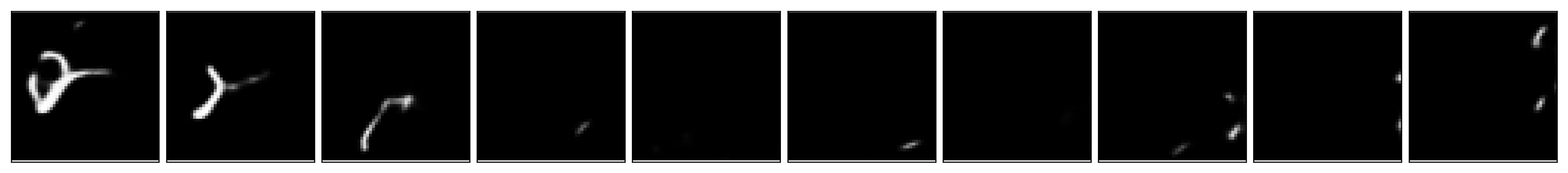}
    \captionof{figure}{Samples from \textsc{conv}-\gls{VRNN}. They show lack of temporal consistency. Objects in the generated frames change between consecutive time-steps and they do not resamble digits from the training set.}
    \label{fig:mnist_samples_vrnn}
\end{center}

\newpage
\subsection{Conditional Generation}

\begin{center}
    \includegraphics[width=\linewidth]{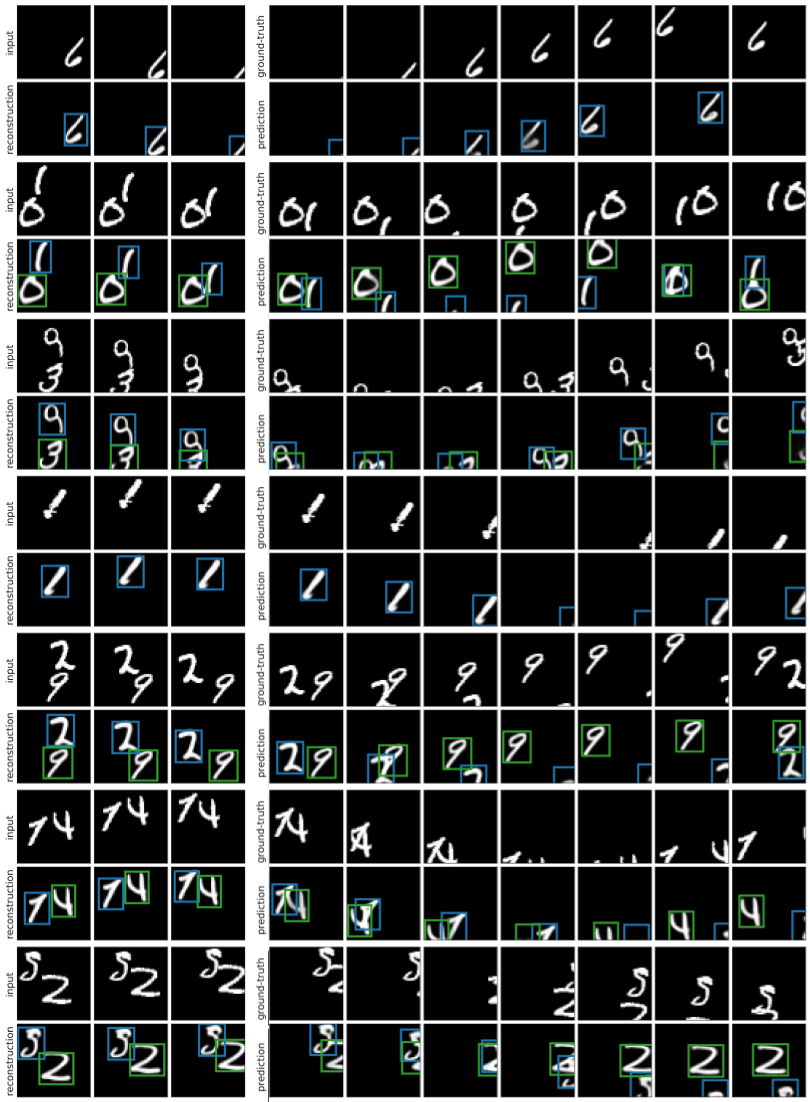}
    \captionof{figure}{Conditional generation from \gls{SQAIR}, which sees only the first three frames in every case. Top is the input sequence (and the remaining ground-truth), while bottom is reconstruction (first three time-steps) and then generation.}
    \label{fig:mnist_cond_gen_sqair}
\end{center}
\section{Reconstruction and Samples from the DukeMTMC Dataset}
\label{app:duke_visual}

\begin{center}
    \begin{minipage}[c]{0.49\linewidth}
        \centering
        \includegraphics[width=\linewidth]{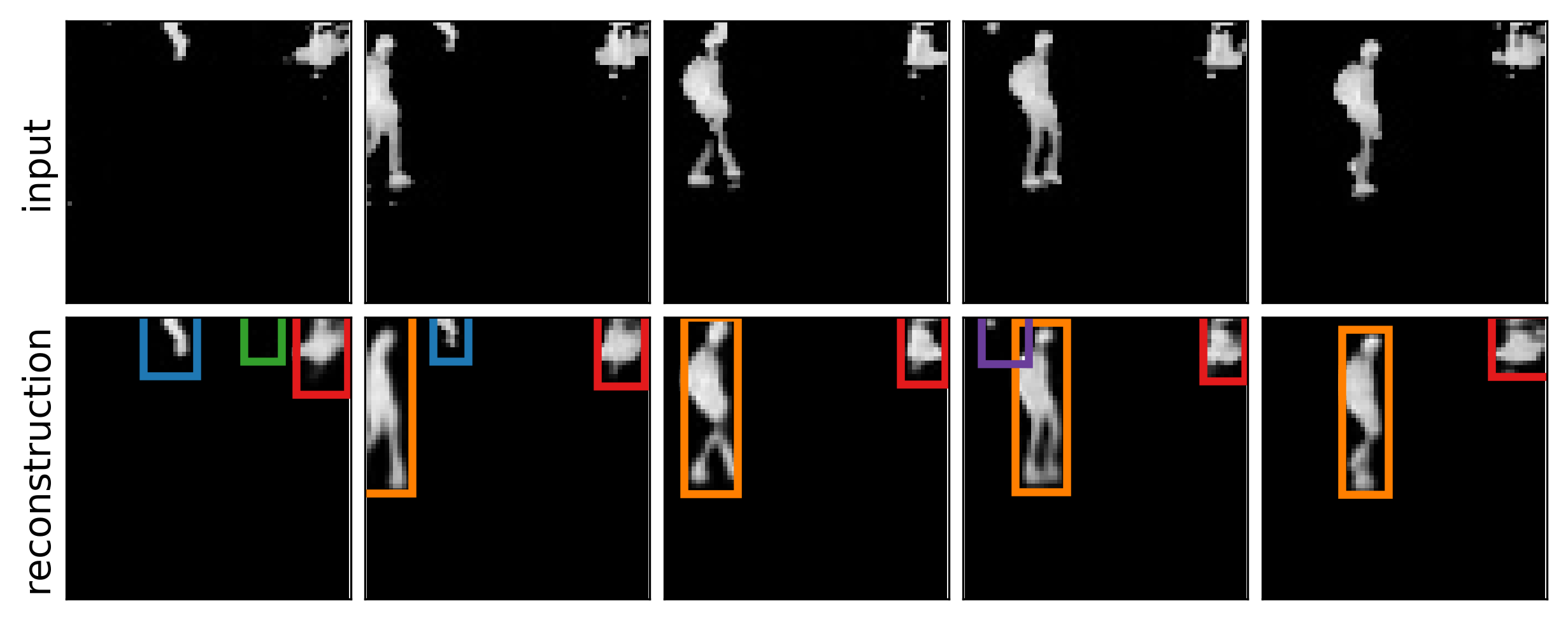}
        \includegraphics[width=\linewidth]{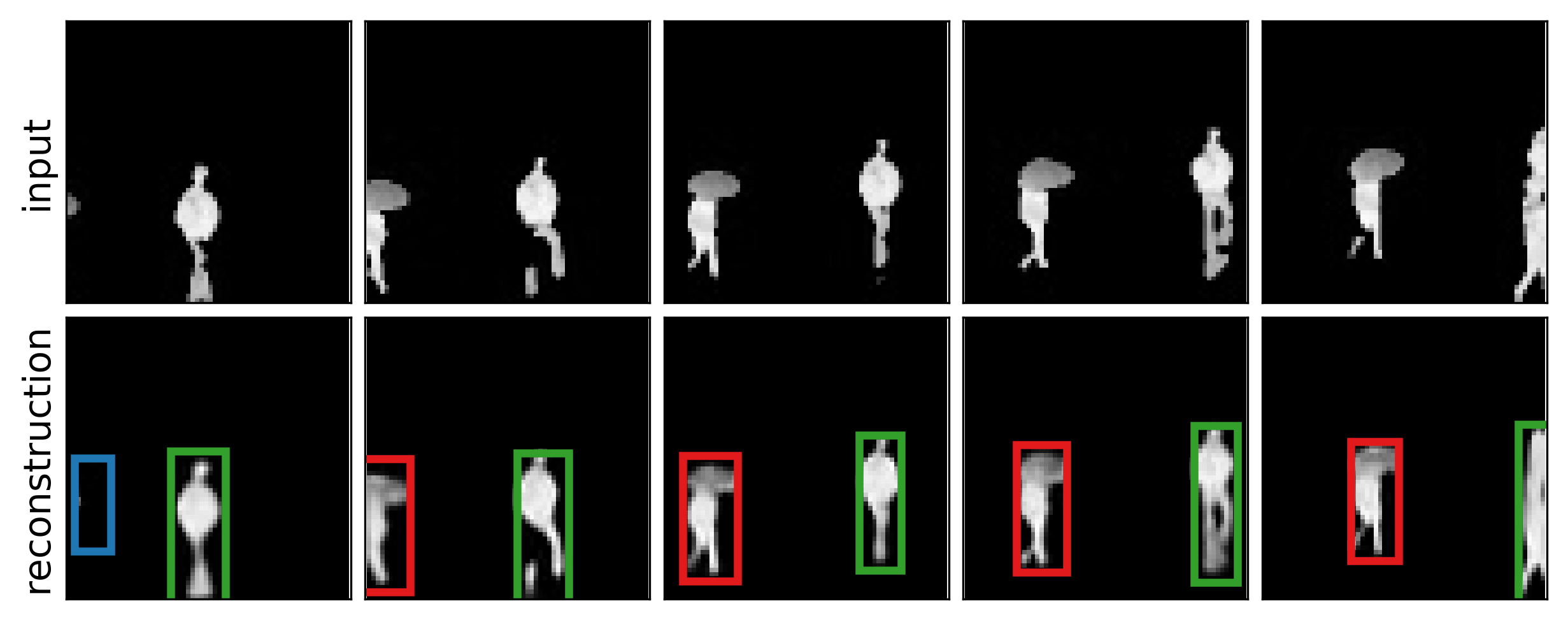}
        \includegraphics[width=\linewidth]{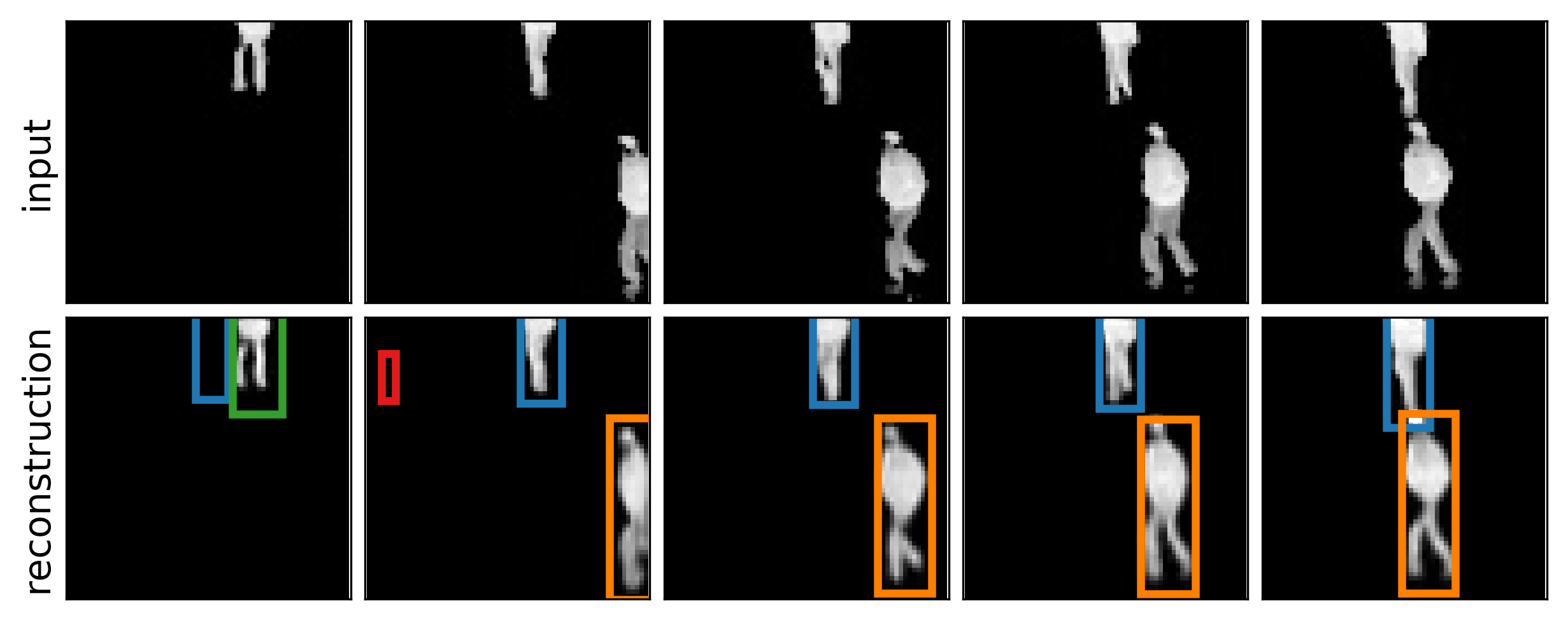}
        \includegraphics[width=\linewidth]{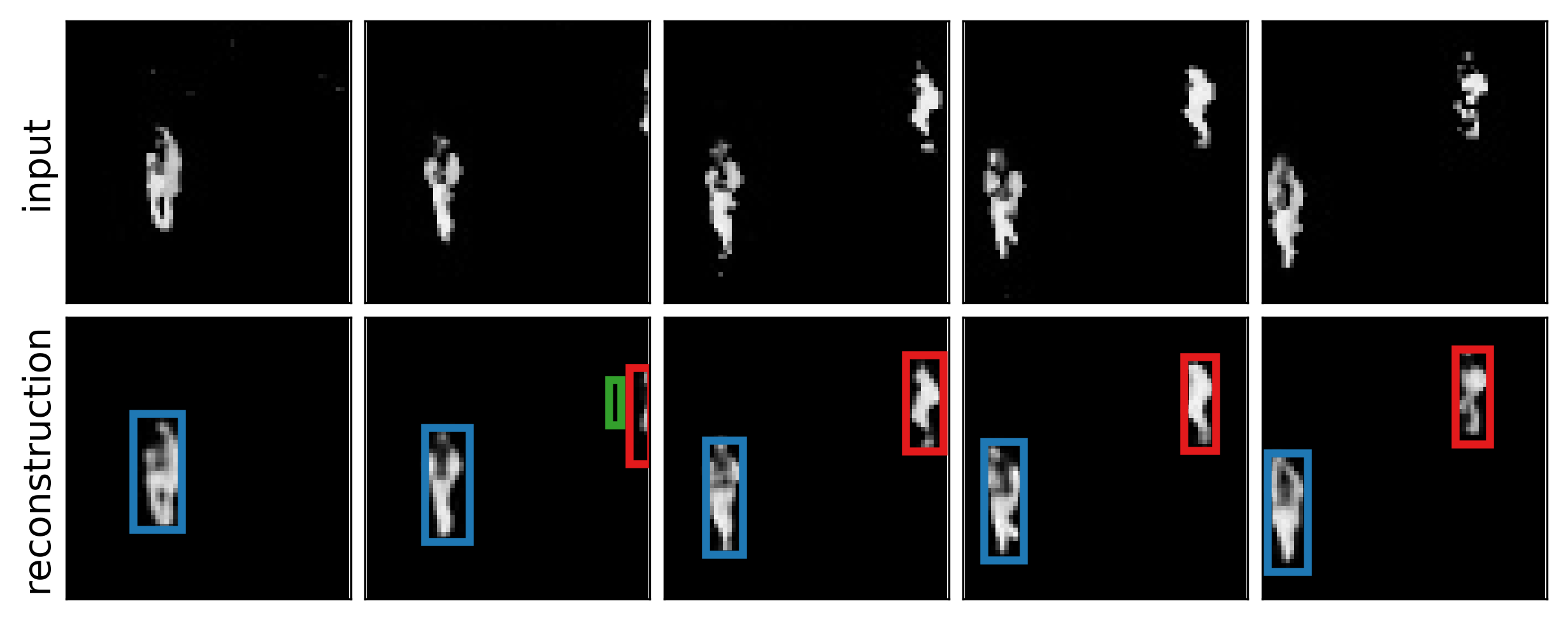}
        \includegraphics[width=\linewidth]{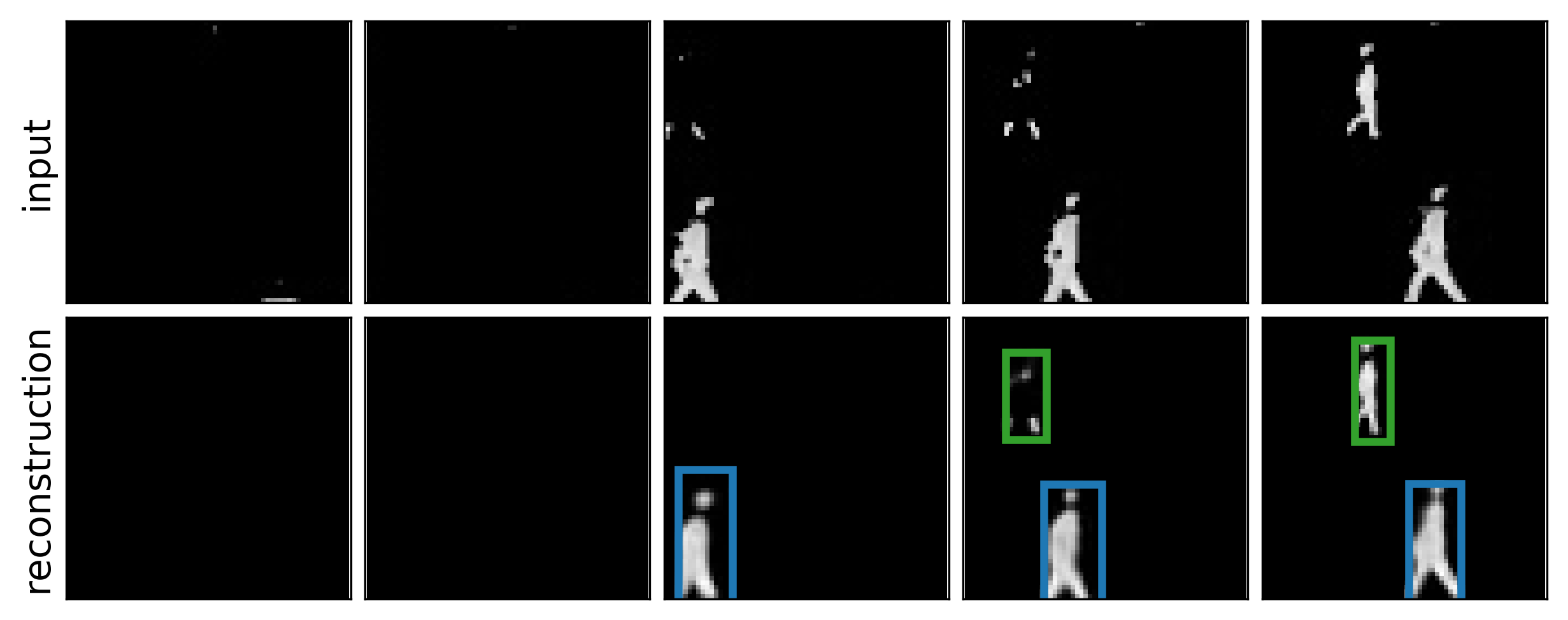}
    \end{minipage}
    \hfill
    \begin{minipage}[c]{0.49\linewidth}
        \centering
        \includegraphics[width=\linewidth]{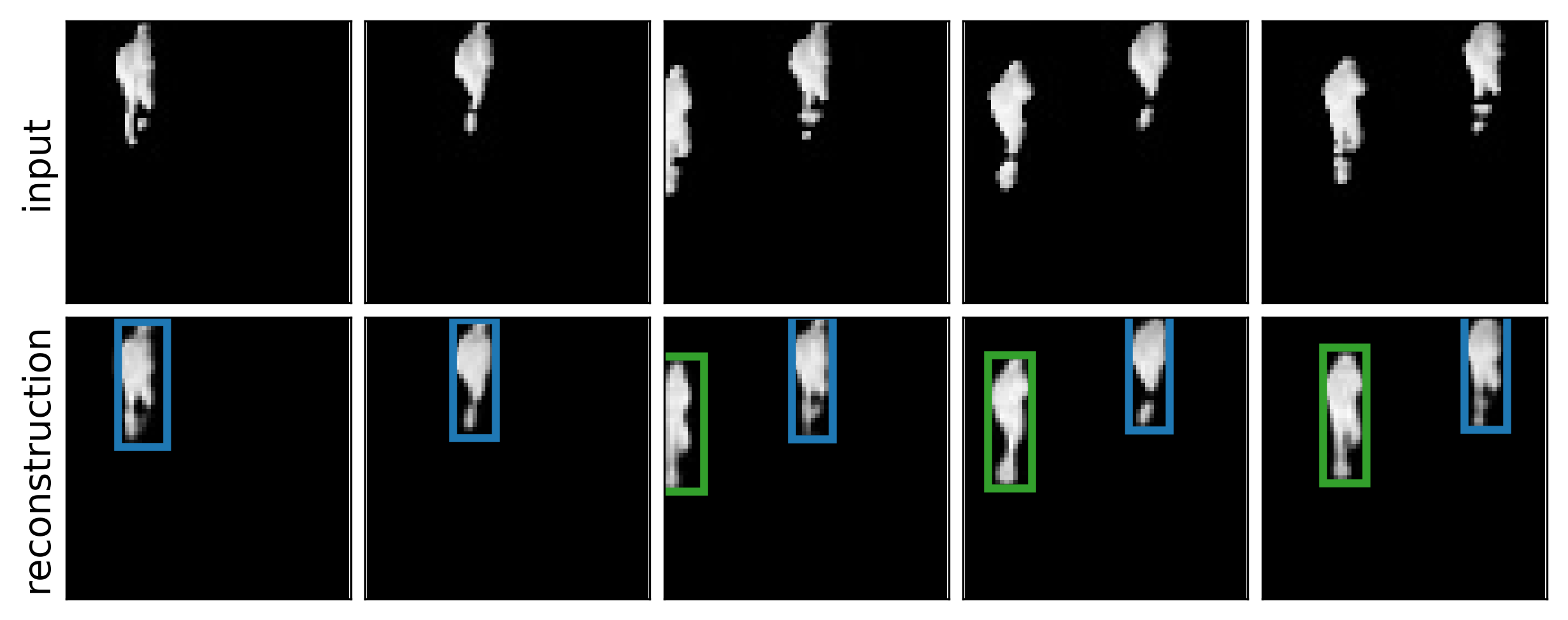}
        \includegraphics[width=\linewidth]{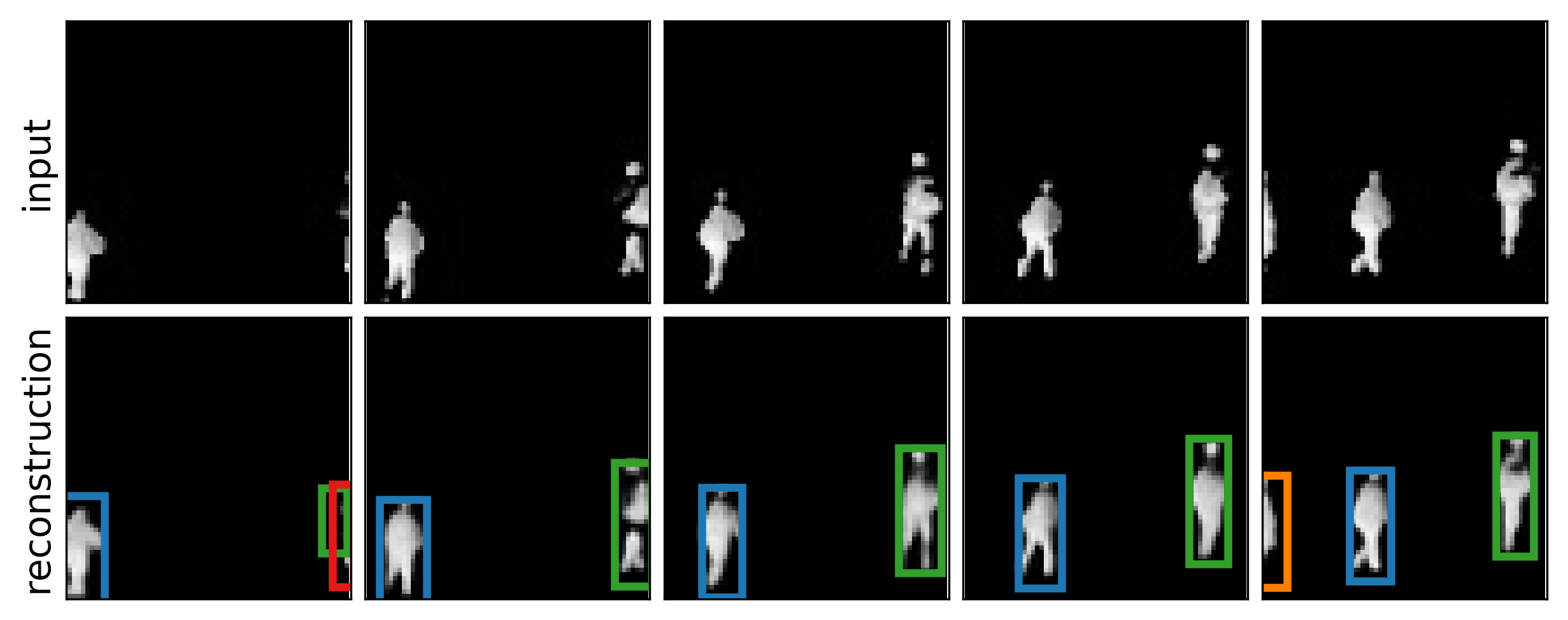}
        \includegraphics[width=\linewidth]{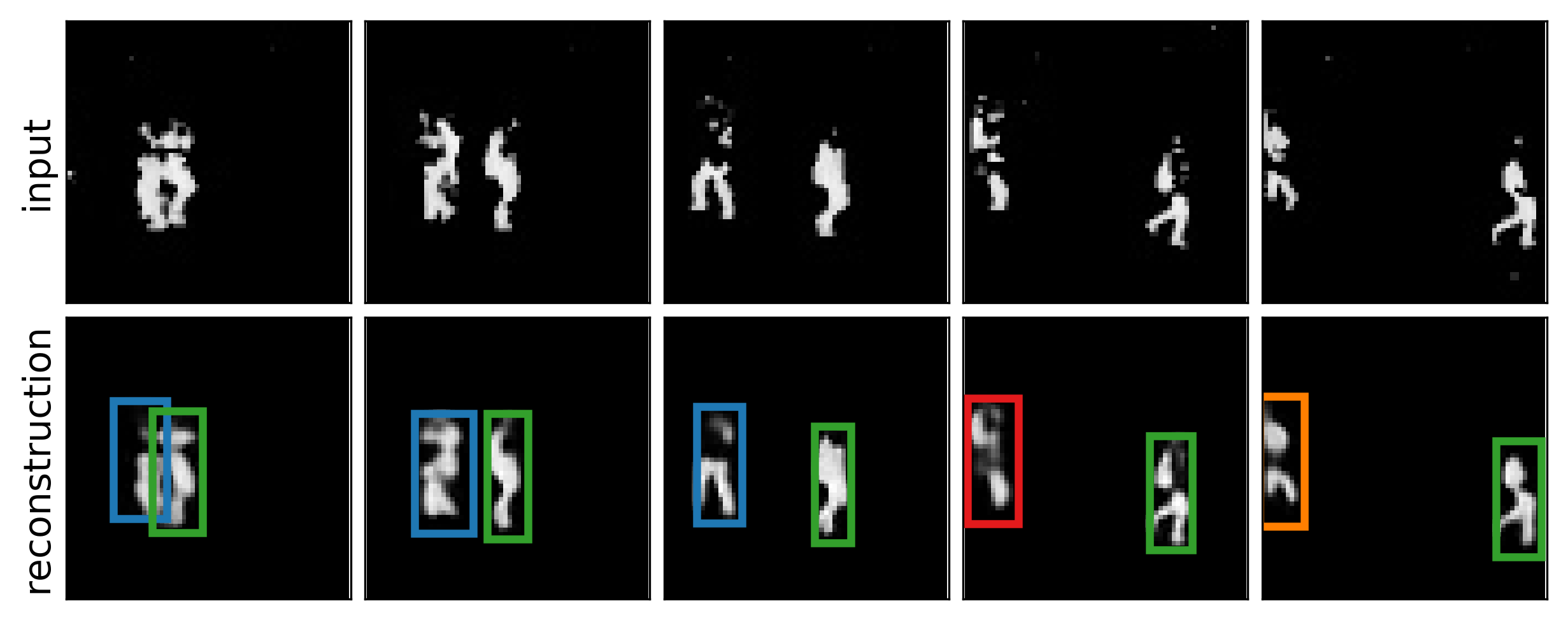}
        \includegraphics[width=\linewidth]{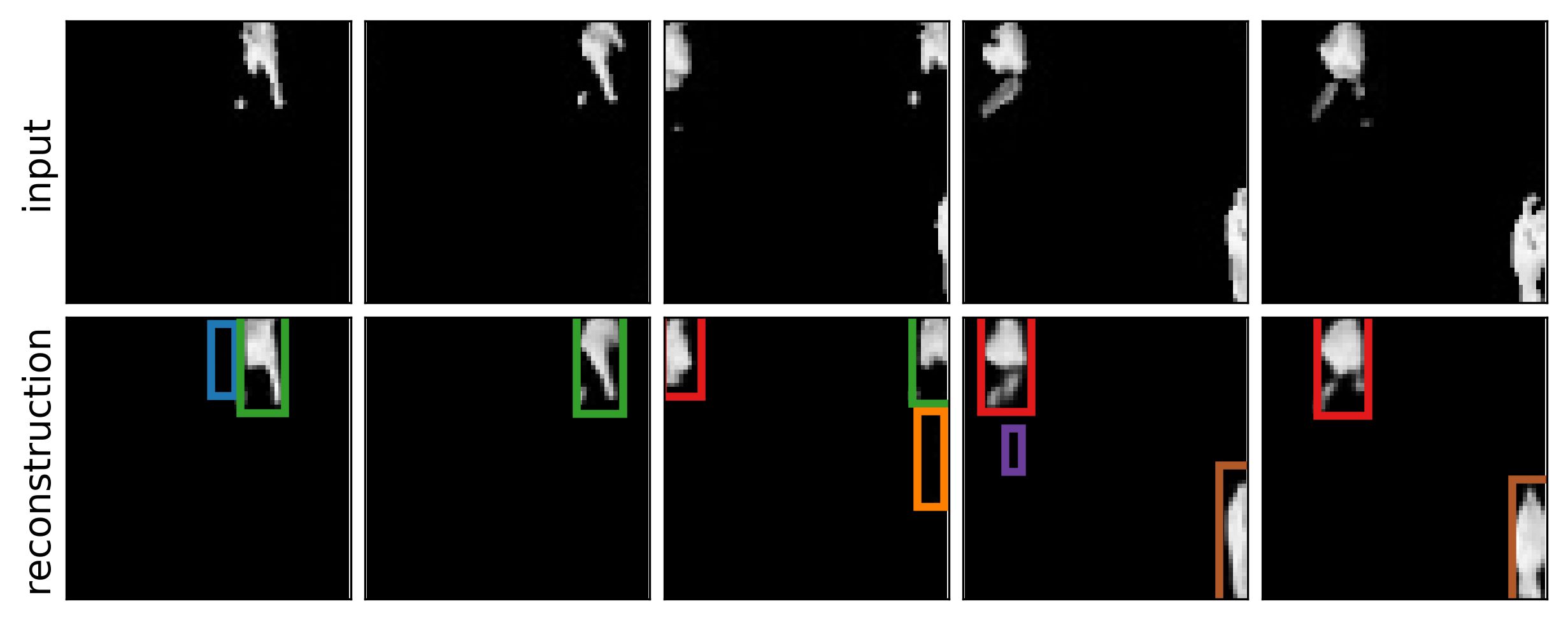}
        \includegraphics[width=\linewidth]{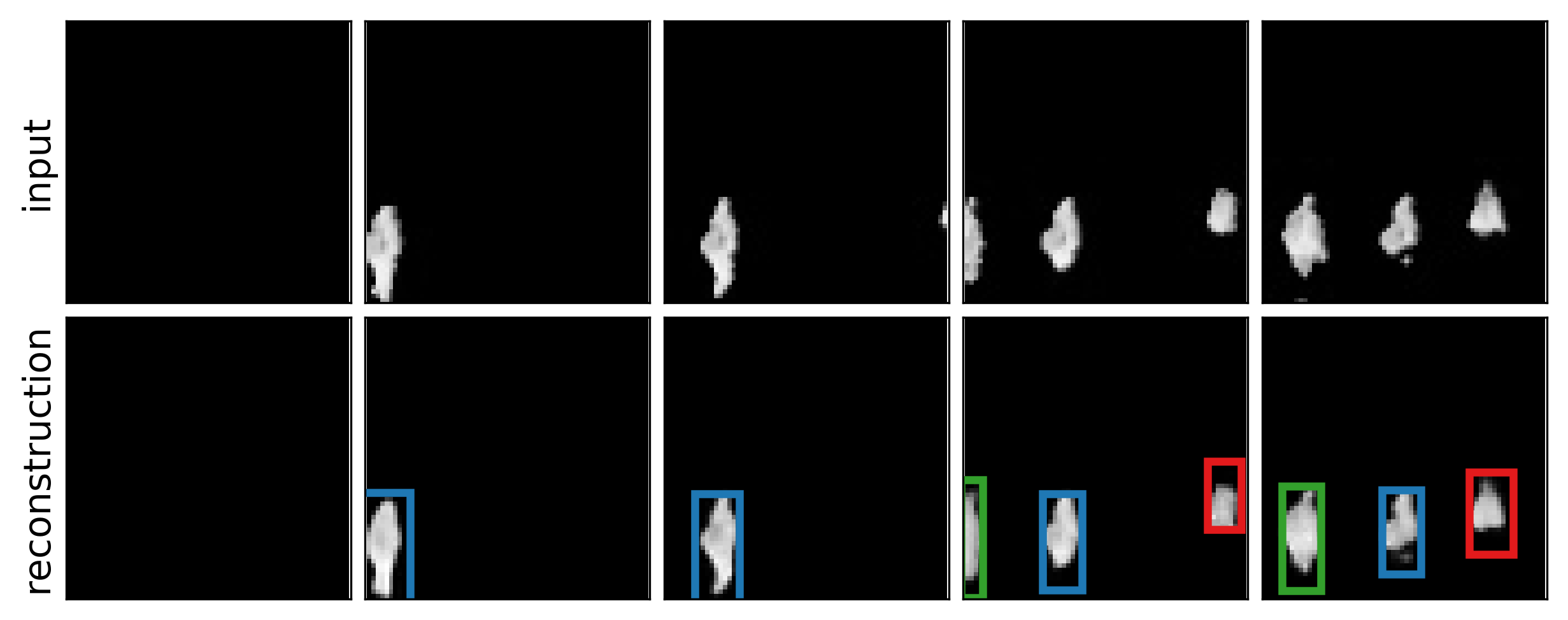}
    \end{minipage}
    \captionof{figure}{Sequences of input (first row) and \gls{SQAIR} reconstructions with marked glimpse locations. While not perfect (spurious detections, missed objects), they are temporally consistent and similar in appearance to the inputs.}
    \label{fig:duke_recs_additional}
\end{center}

\begin{center}
    \begin{minipage}[c]{0.49\linewidth}
        \centering
         \includegraphics[width=\linewidth]{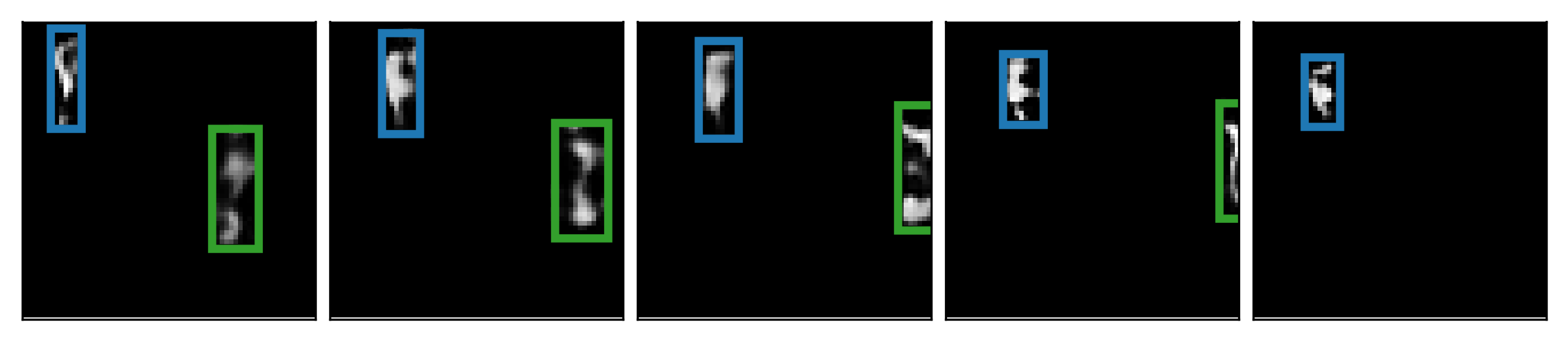}
        \includegraphics[width=\linewidth]{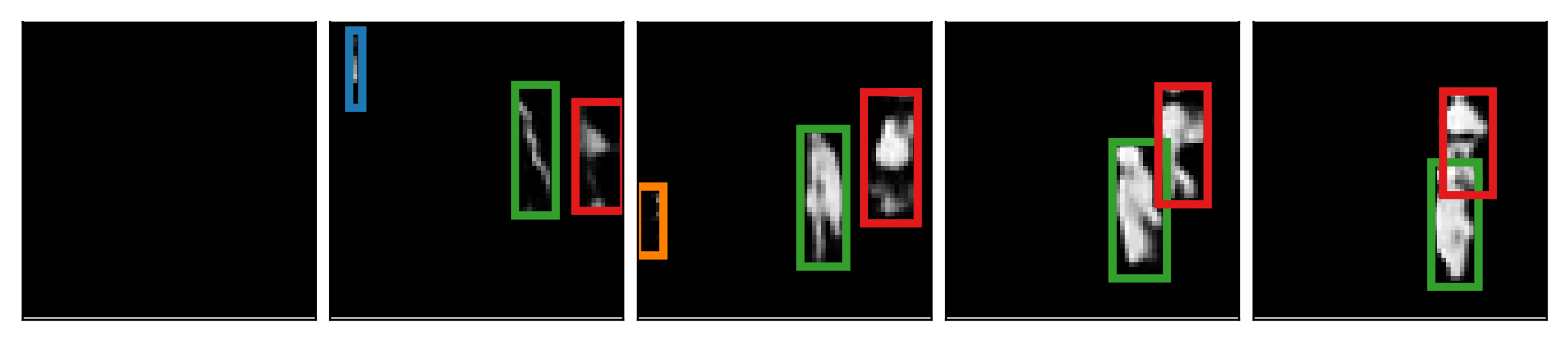}
        \includegraphics[width=\linewidth]{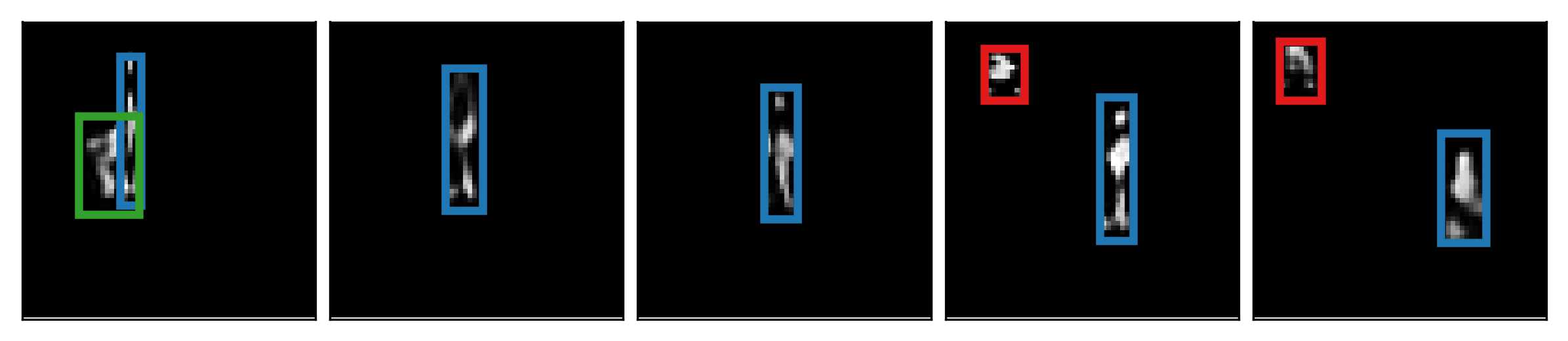}
        \includegraphics[width=\linewidth]{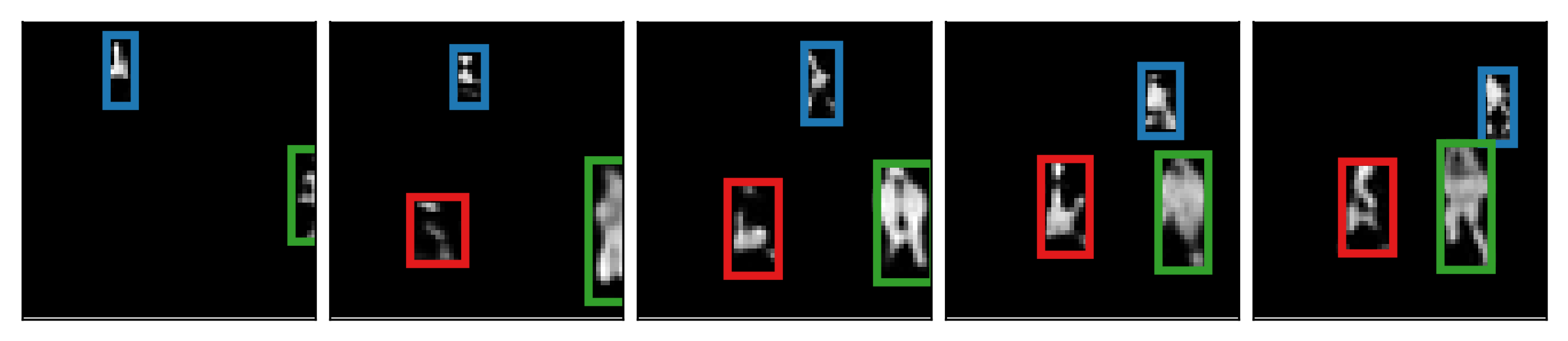}
        \includegraphics[width=\linewidth]{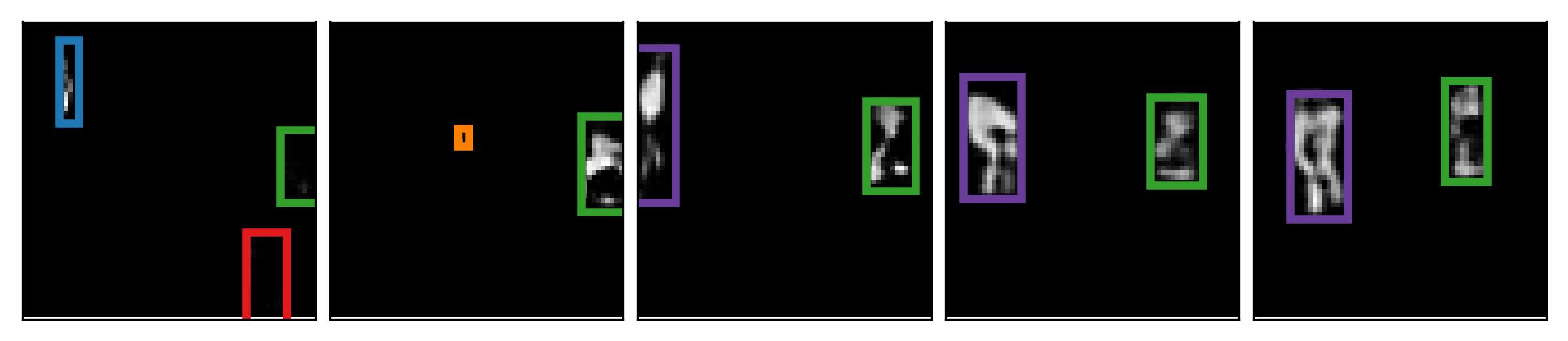}
        \includegraphics[width=\linewidth]{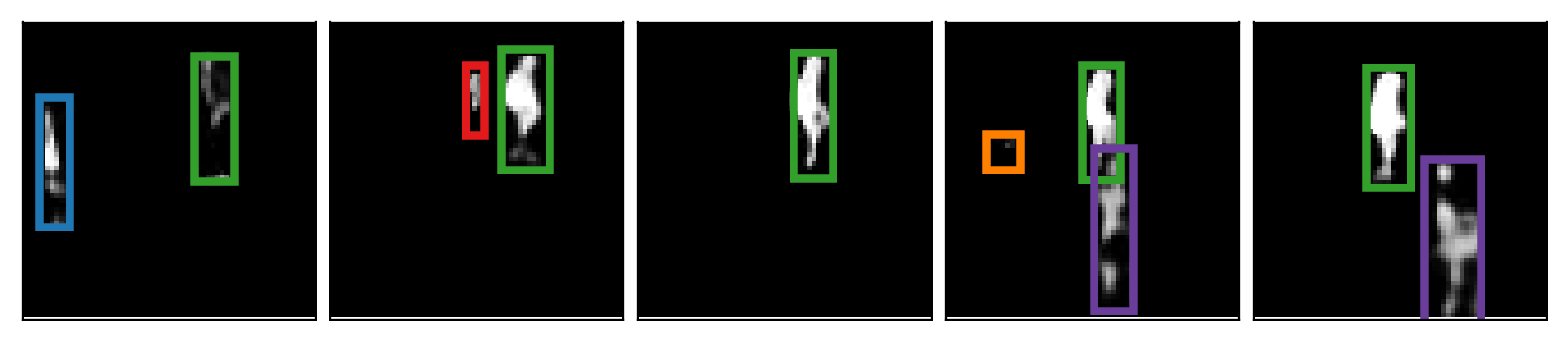}
        \includegraphics[width=\linewidth]{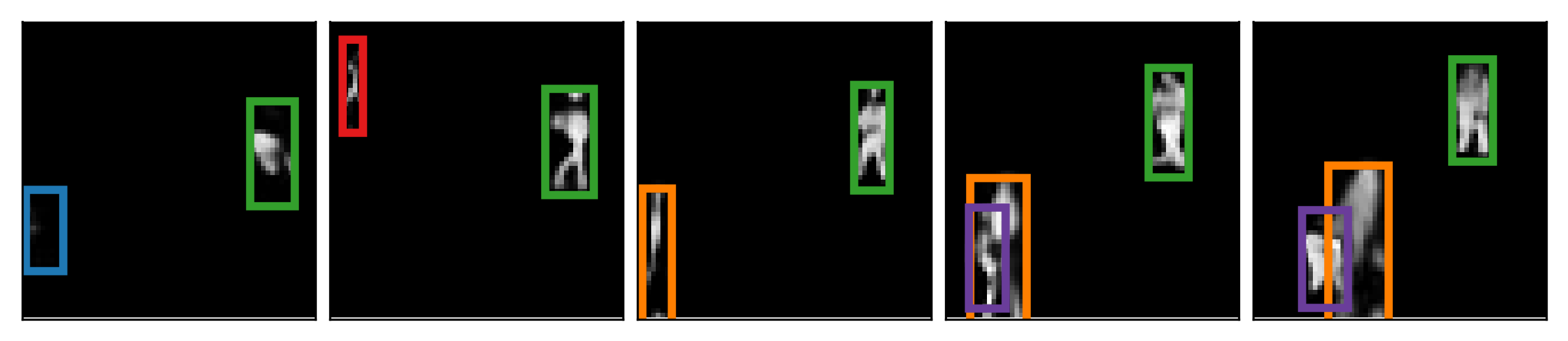}
        \includegraphics[width=\linewidth]{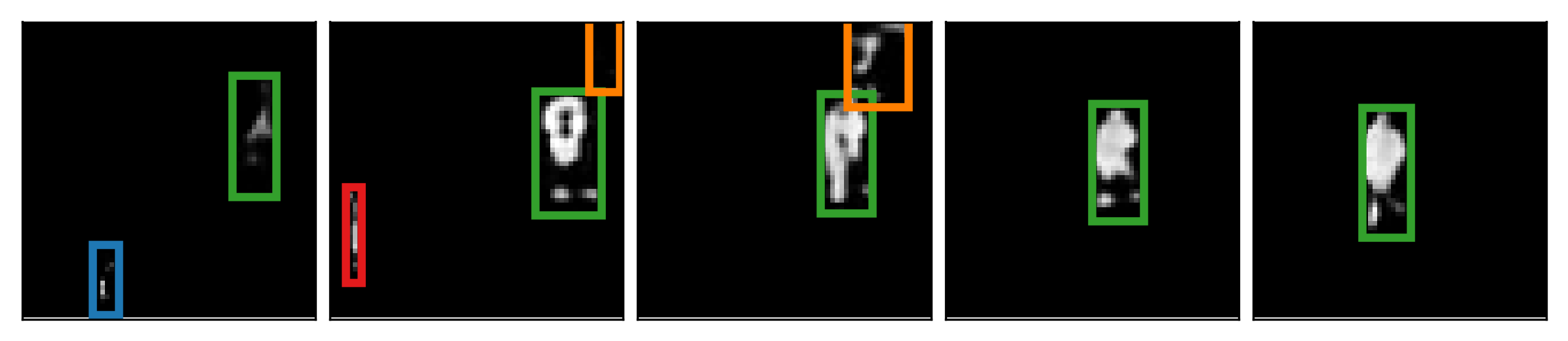}
    \end{minipage}
    \hfill
    \begin{minipage}[c]{0.49\linewidth}
        \centering
            \includegraphics[width=\linewidth]{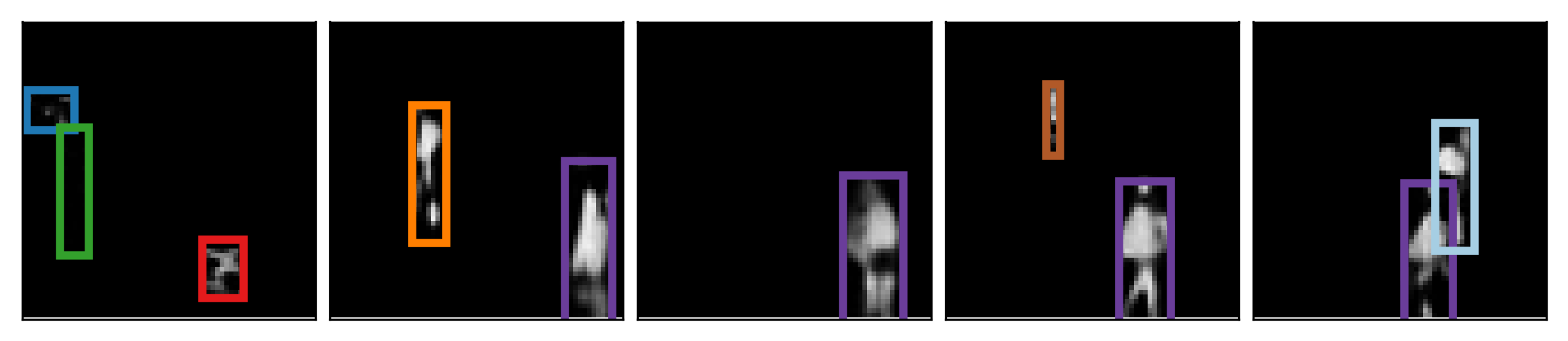}
        \includegraphics[width=\linewidth]{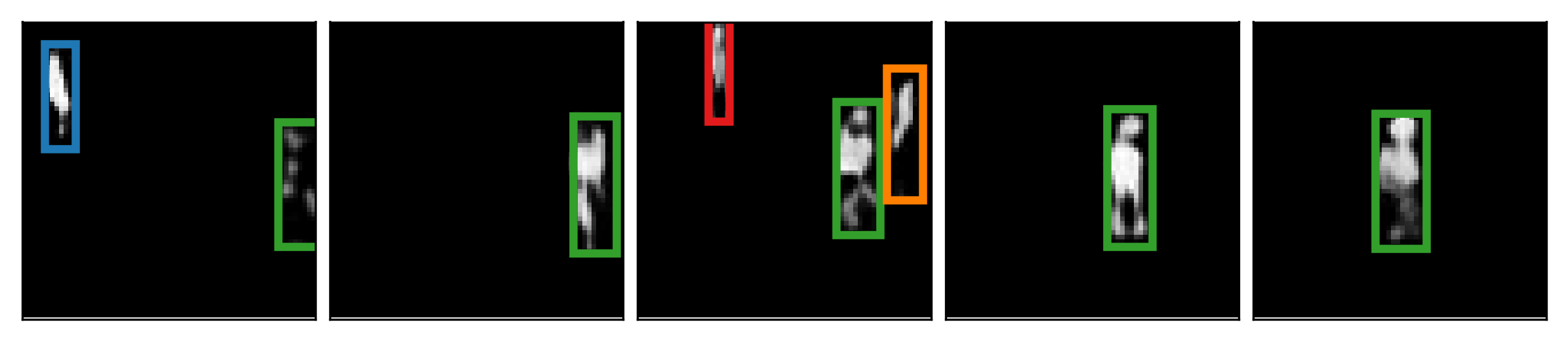}
        \includegraphics[width=\linewidth]{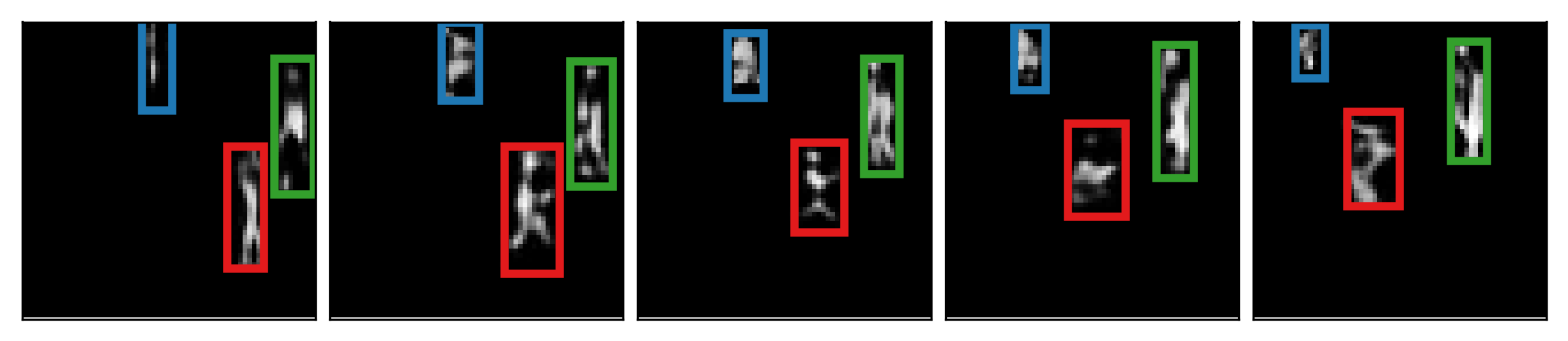}
        \includegraphics[width=\linewidth]{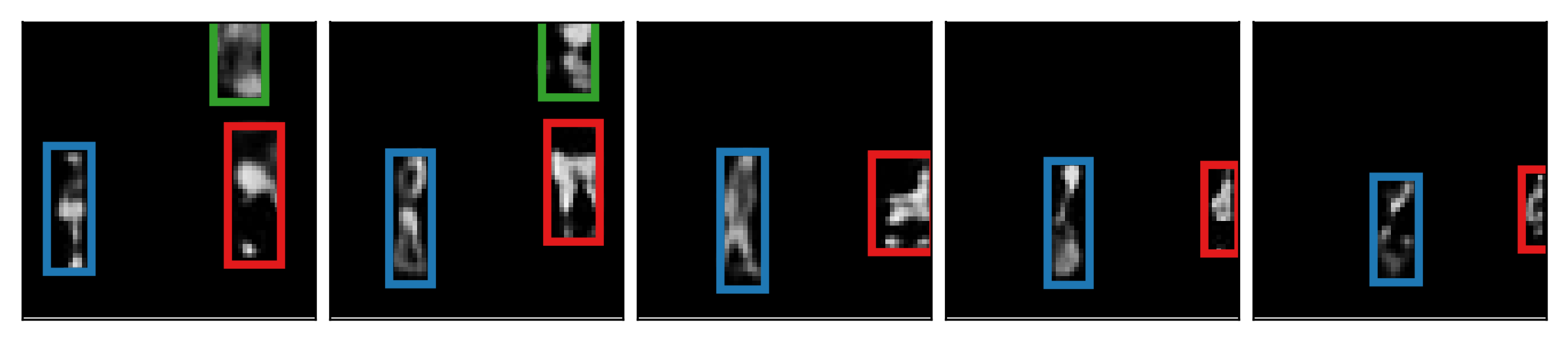}
        \includegraphics[width=\linewidth]{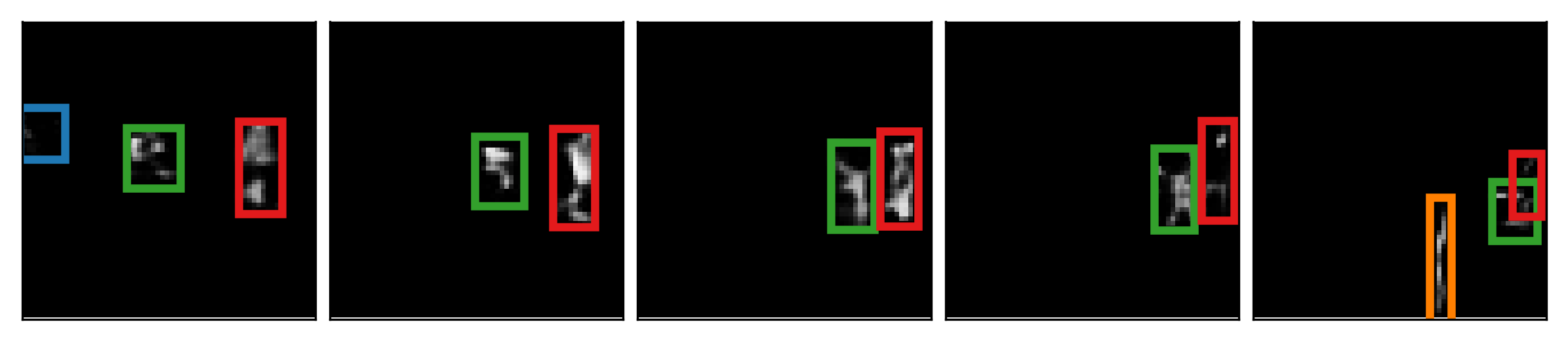}
        \includegraphics[width=\linewidth]{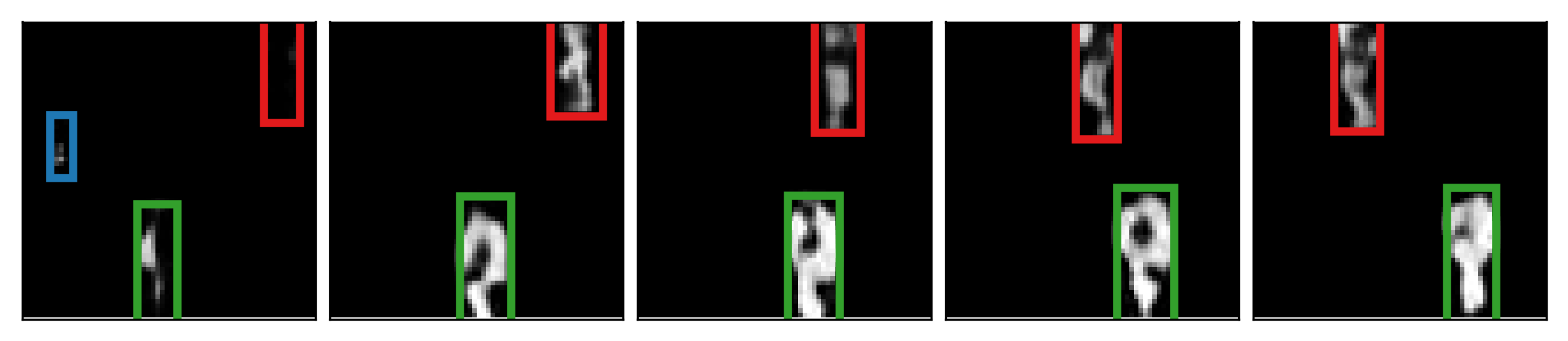}
        \includegraphics[width=\linewidth]{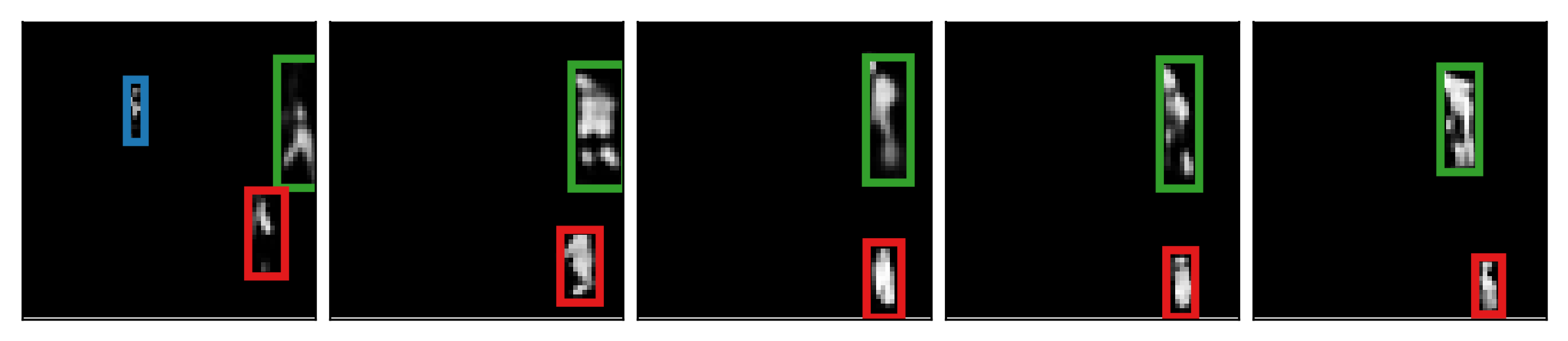}
    \end{minipage}
    \captionof{figure}{Samples with marked glimpse locations from \gls{SQAIR} trained on the DukeMTMC dataset. Both appearance and motion is spatially consistent. Generated objects are similar in appearance to pedestrians in the training data. Samples are noisy, but so is the dataset.}
    \label{fig:duke_samples_additional}
\end{center}

\begin{center}
    \includegraphics[width=\linewidth]{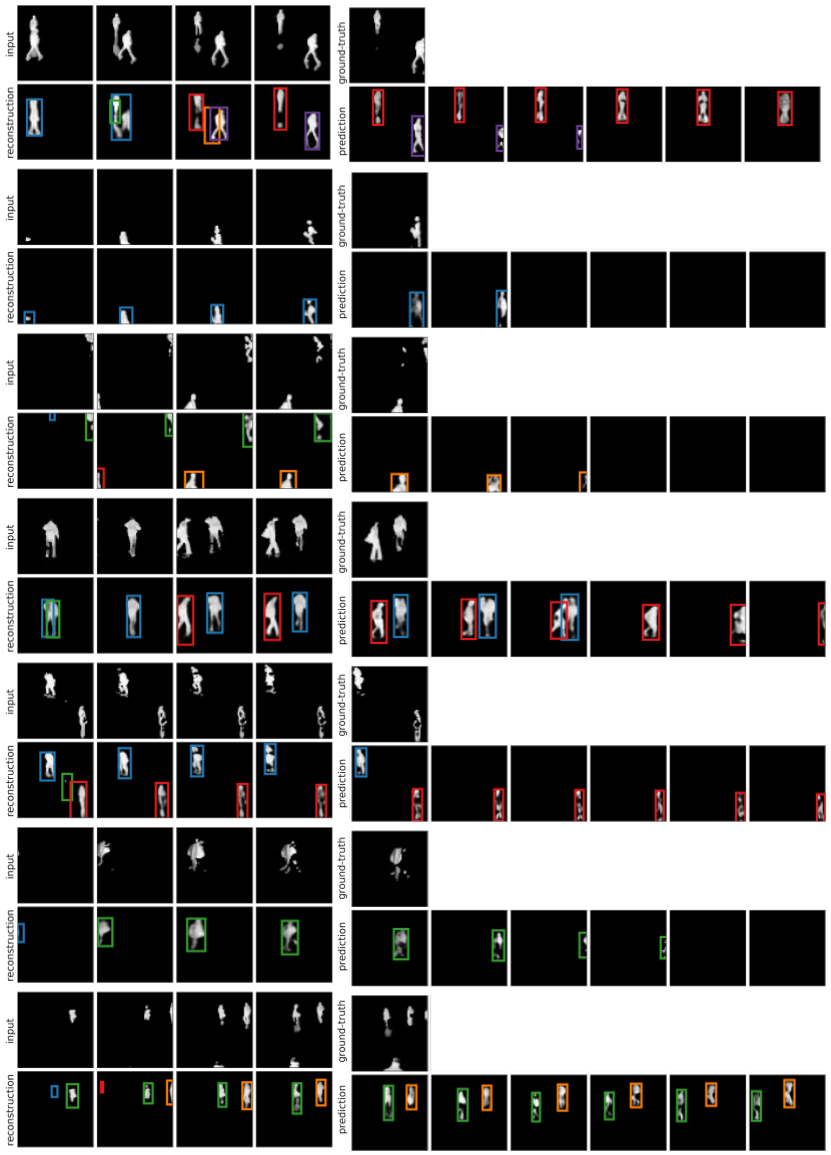}
    \captionof{figure}{Conditional generation from \gls{SQAIR}, which sees only the first four frames in every case. Top is the input sequence (and the remaining ground-truth), while bottom is reconstruction (first four time-steps) and then generation.}
    \label{fig:duke_cond_gen_sqair}
\end{center}

\end{document}